\documentclass[letterpaper]{article} 
\usepackage{aaai2026}  
\usepackage{times}  
\usepackage{helvet}  
\usepackage{courier}  
\usepackage[hyphens]{url}  
\usepackage{graphicx} 
\usepackage{natbib}  
\usepackage{caption} 
\usepackage{annotate-equations}
\usepackage{algorithm}
\usepackage{algorithmic}

\usepackage{amsthm}
\usepackage{pifont}
\usepackage{enumitem}
\usepackage{amsmath}
\usepackage{amssymb}
\usepackage{booktabs}
\usepackage{multirow}
\usepackage{makecell}
\usepackage{colortbl}
\usepackage{tikz}
\usepackage{subfigure}
\usepackage{enumitem}
\usepackage{lipsum}
\usepackage[most]{tcolorbox}
\newtcolorbox{mathbox}{
  colframe=black,
  colback=white,
  sharp corners,
  boxrule=0.5pt,
  breakable,
  left=1.5pt, right=1.5pt, top=1.5pt, bottom=1.5pt
}

\newcommand{\modelname}{SA\textsuperscript{2}GFM}

\newcommand{\eg}{\textit{e.g.}}

\theoremstyle{definition}

\newtheorem{Deriv}{Derivation}

\usepackage{newfloat}
\usepackage{listings}
\DeclareCaptionStyle{ruled}{labelfont=normalfont,labelsep=colon,strut=off} 
\floatstyle{ruled}
\newfloat{listing}{tb}{lst}{}
\floatname{listing}{Listing}
\title{SA\textsuperscript{2}GFM: Enhancing Robust Graph Foundation Models\\with Structure-Aware Semantic Augmentation}
\author{
    Junhua Shi\textsuperscript{\rm 1}, Qingyun Sun\textsuperscript{\rm 1}\thanks{Corresponding author.}, Haonan Yuan\textsuperscript{\rm 1}, Xingcheng Fu\textsuperscript{\rm 2}
}
\affiliations{
    \textsuperscript{\rm 1}SKLCCSE, School of Computer Science and Engineering, Beihang University, Beijing, China\\
    \textsuperscript{\rm 2}Key Lab of Education Blockchain and Intelligent Technology, Guangxi Normal University, Guilin, China\\
    \{shijunhua, sunqy, yuanhn\}@buaa.edu.cn, fuxc@gxnu.edu.cn
}

\usepackage{bibentry}

\begin{document}

\maketitle

\begin{abstract}
While Graph Foundation Models (GFMs) have achieved notable progress across diverse tasks recently, their robustness under domain noise, structural perturbations, and even adversarial attacks remains largely underexplored. A core limitation lies in the inadequate modeling of hierarchical structural semantics, which are intrinsic priors and critical for generalization.
In this work, we propose \textbf{\modelname}, a robust \underline{\textbf{GFM}} framework that enhances the domain-adaptable representations through \underline{\textbf{S}}tructure-\underline{\textbf{A}}ware \underline{\textbf{S}}emantic \underline{\textbf{A}}ugmentation.
First, to embed the hierarchical structural priors, we transform entropy-based encoding trees into structure-aware textual prompts for feature augmentation. The enriched inputs are processed by a novel self-supervised Information Bottleneck mechanism that distills the robust and transferable representations through structure-guided compression.
To mitigate the negative transfer in cross-domain adaptation, we develop an expert adaptive routing mechanism that integrates a mixture-of-experts architecture with a null expert design.
To enable efficient downstream adaptation, we propose a fine-tuning module that efficiently optimizes the hierarchical structures through the joint intra- and inter-community structure learning.
Extensive experiments validate the superiority of \modelname~over effectiveness and robustness against random noise and adversarial perturbations on node and graph classification compared with 9 state-of-the-art baselines.
\end{abstract}

\section{Introduction}

\begin{figure}[!t]
  \centering
  \includegraphics[width=1\linewidth]{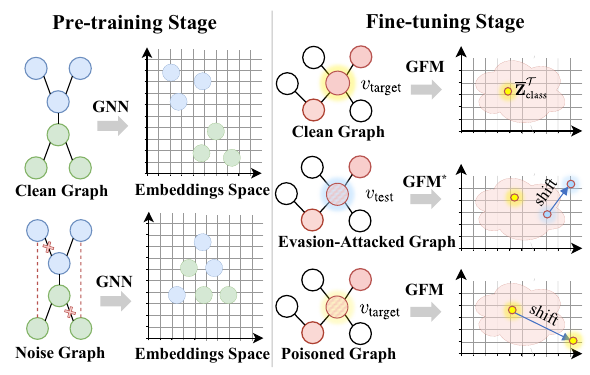}
  \caption{Effects of different attacks in pre-training and fine-tuning. The asterisk (\textasteriskcentered) denotes the trained module.}
  \label{fig:intro}
\end{figure}
Graph Neural Networks (GNNs) have achieved impressive success in learning graph-structured data~\cite{kipf2017semisupervised, veličković2018graph} across a wide range of domains. However, traditional GNNs are designed for specific tasks and datasets, limiting their ability to transfer learned knowledge to unseen domains~\cite{Hu2020Strategies}. To address this limitation, recent work has advanced towards Graph Foundation Models (GFMs), which aim to learn general-purpose graph representations via large-scale multi-domain pre-training and efficient downstream adaptation with limited supervision~\cite{yu2024text, wang2025multidomain, yuan2025how}. The paradigm of ``pretrain-then-finetune'' promises to build adaptable GFMs that generalize across a broad spectrum of graph-based tasks and domains~\cite{DBLP:journals/corr/abs-2310-11829}.

As shown in Figure \ref{fig:intro}, the robustness of Graph Foundation Models (GFMs) under noisy and adversarial perturbations remains largely underexplored \cite{wang2025multidomain}. Particularly, a fundamental limitation arises from the insufficient modeling of hierarchical structural semantics: the inductive priors that are crucial for ensuring both adaptation and robustness. Existing GFMs predominantly adopt shallow message-passing GNNs as their backbone architectures, which are theoretically limited by the 1-Weisfeiler-Lehman (1-WL) test~\cite{xu2018how, morris2019weisfeiler}, rendering them incapable of distinguishing structurally similar yet semantically distinct patterns. Consequently, their learned representations often neglect long-range dependencies and higher-order structure semantics, leaving them vulnerable to noise and misalignment under real-world deployment.
Moreover, most GFMs overlook the opportunity to embed structure-aware semantics~\cite{yu2024text, yuan2025how}. They tend to directly encode raw node attributes, which are often incomplete or noisy, while ignoring global community hierarchies that provide stable semantic anchors across domains. Without explicit modeling of such structure-induced semantics.

Beyond architectural expressiveness, GFMs often struggle in real-world deployment where robustness is essential. Existing domain adaptation GFMs (\eg, MDGPT~\cite{yu2024text}, BRIDGE)~\cite{yuan2025how} rely on over-idealized assumptions like dimension alignment or domain invariance, which frequently break across heterogeneous graphs~\cite{wang2024dissecting}. Meanwhile, some other GFMs claim to be perturbation robust by applying global structure learning~\cite{wang2025multidomain}, but they often incur high computational costs and remain fragile to localized perturbations or adversarial attacks~\cite{kataria2024ugc, zhang2024gder}. A more critical yet overlooked problem is negative transfer: when structural or semantic gaps between domains are large, the na\"ive source aggregation can severely degrade downstream performance, especially under dynamic or non-stationary settings~\cite{NEURIPS2024_59e73ff8, yuan2023environment}.

The aforementioned challenges expose three critical bottlenecks for constructing a robust GFM:
\textbf{(1) Pre-training: How to embed structural semantic priors while preserving information purity?}
Raw node features are insufficient for capturing transferable patterns across domains, yet na\"ive feature enhancement may introduce spurious correlations. A principled mechanism is needed to embed high-level structural priors while compressing irrelevant noise.
\textbf{(2) Knowledge fusion: How to mitigate negative transfer under domain discrepancy?}
The diverse nature of real-world graphs necessitates adaptive fusion strategies that can selectively route relevant knowledge and suppress misleading signals from unrelated sources.
(3) \textbf{Downstream fine-tuning: How to efficiently optimize structures under noise and adversarial perturbations?}
Topology is the most vulnerable aspect of graphs, yet precise and efficient structure refinement remains an open problem due to the high cost and fragility of global structure learning strategies.

To address the challenges, we propose a novel \textbf{\modelname}, a robust \underline{\textbf{G}}raph \underline{\textbf{F}}oundation \underline{\textbf{M}}odel empowered by \underline{\textbf{S}}tructure-\underline{\textbf{A}}ware \underline{\textbf{S}}emantic \underline{\textbf{A}}ugmentation. Our key insight is to embed hierarchical semantic priors using entropy-based encoding trees, which generate structure-aware prompts to enrich node features. The augmented inputs are processed by a self-supervised Information Bottleneck mechanism to learn robust, domain-adaptable representations by compressing redundancy and preserving label-relevant information. To mitigate domain mismatch, we adopt an expert routing module with a Mixture-of-Experts and a null expert to suppress irrelevant sources. Finally, a lightweight fine-tuning strategy refines graph structures through hierarchical optimization for improved robustness.
\textbf{Our contributions are:}
\begin{itemize}[leftmargin=*]
    \item We propose \modelname, a robust GFM framework by synergistically addressing key interrelated challenges in feature enhancement, principled and efficient structure optimization, and knowledge fusion.
    \item It integrates a structure-aware pre-training strategy utilizing encoding tree textual prompts and self-supervised Information Bottleneck, a domain-adaptive MoE with a null expert for transfer mitigation, and an efficient fine-tuning module for hierarchical structure refinement.
    \item Extensive experiments show superiority of \modelname~over effectiveness and robustness against noise and adversarial perturbations on node and graph classification compared with 9 state-of-the-art baselines.
\end{itemize}

\section{{Related Work}}
\label{sec:related_work}
Detailed related works are discussed in Appendix~\ref{sec:related_work_add}.
\subsection{Multi-domain Graph Foundation Models}
Multi-domain pre-training is fundamental for constructing graph foundation models capable of adapting across heterogeneous graphs~\cite{yuan2025dg, yuangraver}. Existing methods typically align either node features, using domain tokens~\cite{yu2024text} or domain-invariant aligners~\cite{yuan2025how}, or graph structures via Graph Structure Learning (GSL)~\cite{wang2025multidomain}. Some approaches also explore flexible transfer frameworks to reduce negative transfer~\cite{ju2025graphbridge}. However, most of these efforts address feature and structure adaptation in isolation, lacking a unified perspective that jointly considers both levels.

\subsection{Robust Graph Representation Learning}
The line of research aims to improve stability under feature and structure noise. For feature robustness, prior work leverages data augmentation~\cite{zheng2024intramix, huang2025learn} and adversarial training~\cite{lee2025selfsupervised}. For structural robustness, Graph Structure Learning (GSL) is widely used to counter topological perturbations~\cite{zugner2018adversarial, yuan2024dynamic}. However, most methods treat feature and structure separately, and many GSL techniques remain costly and coarse-grained.

\subsection{Graph Strcutural Entropy}
Graph Structural Entropy is an information-theoretic measure that captures the hierarchical organization of a graph through community encoding~\cite{li2016structural}. This hierarchy has shown promise in guiding representation learning~\cite{bai2024hcgae}, yet its use for enhancing raw node features and improving the robustness of graph foundation models remains largely unexplored.
\section{Preliminary}
\label{sec:preliminary}

\textbf{Notations.}
A graph is denoted as $G = \left(\mathcal{V}, \mathcal{E}\right)$, where $\mathcal{V}$ is the node set and $\mathcal{E}$ is the edge set. Let $\mathbf{A}\in\{0,1\}^{N \times N}$ be the adjacency matrix and $\mathbf{X}\in\mathbb{R}^{N \times d_i}$ be the node feature matrix, where $N = | \mathcal{V} |$ is the number of nodes and $d_i$ denotes input dimension. $\mathbf{Z}$ is hidden embeddings of $\mathbf{X}$.

\textbf{Problem Settings.}
Our proposed \modelname~operates under the multi-domain pre-training and few-shot fine-tuning paradigm. Given $n$ source graphs $\{G^\mathcal{S}\} \in \mathcal{G}^\mathcal{S}$ from multiple domains $\{D^\mathcal{S}\} \in \mathcal{D}^\mathcal{S}$ with their labels $\{Y^\mathcal{S}\} \in \mathcal{Y}^\mathcal{S}$, the pre-training goal is to train a graph learner $h=g(\mathcal{F}_{\boldsymbol{\Theta}}(\cdot))$ on multi-domain graph datasets, after which the pre-training parameter $\boldsymbol{\Theta}^\star$ is frozen.
During fine-tuning, given a set of target graphs $\{G^\mathcal{T}\} \in \mathcal{G}^\mathcal{T}$ (potentially corrupted by noise or adversarial attacks) from target domain $\{D^\mathcal{T}\}\in\mathcal{D}^\mathcal{T}$ (seen or unseen) with $m$ accessible labels $\{Y^\mathcal{T}\}\in\mathcal{Y}^\mathcal{T}$ under $m$-shot setting ($m \ll n$), the ultimate goal is to leverage the pre-trained $\boldsymbol{\Theta}^\star$ to achieve robust performance on the unlabeled nodes of the target graph.
\begin{figure*}[!t]
  \centering
\includegraphics[width=0.98\textwidth,height=6cm,keepaspectratio]{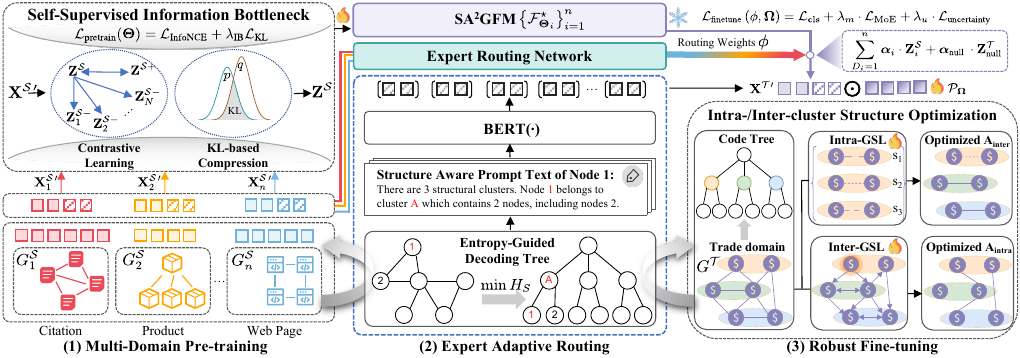}
  \caption{Overview of the \modelname~framework. The framework first pre-trains an ensemble of expert encoders on multiple source domains using Structure-Aware Semantic Augmentation and an Information Bottleneck objective. During fine-tuning, a novel Expert Adaptive Routing mechanism selects knowledge to mitigate negative transfer, while a Hierarchical Structure Optimization module refines the target graph's topology before final prediction.}
  \label{fig:framework}
\end{figure*}

\section{Proposed Framework: \modelname}

In this section, we introduce the proposed \modelname~with its framework illustrated in Figure~\ref{fig:framework}.

\subsection{Multi-domain Pre-training with Self-Supervised Information Bottleneck}
\label{seq:pretrain}

We begin by encoding structural priors into node features via textual prompts derived from entropy-based encoding trees.

\subsubsection{Structure-Aware Semantic Augmentation.}
\begin{figure}[!t]
  \centering
  \includegraphics[width=8cm,keepaspectratio]{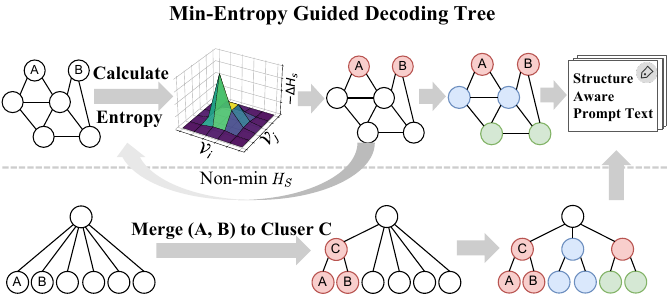}
  \caption{Construct code tree by minimizing structure entropy to generate structure aware prompt text.}
  \label{fig:prompt_tree}
\end{figure}

Real-world graphs exhibit rich hierarchical priors that are not explicitly encoded in node. To expose such a structure to encoders, we propose to convert the latent community hierarchy into textual prompts, which allows the pre-trained language models to serve as a bridge between structural regularities and semantic representations.
To achieve this, we leverage the theory of graph structural entropy~\cite{long2020graphstone}, which identifies an optimal recursive clustering by minimizing entropy over volume-weighted partitions. Given a partition $\mathcal{C}^\mathcal{S} = \{C_1, \cdots, C_K\}$ of graph $G^\mathcal{S}$, the entropy is:
\begin{equation}
\label{eq:1}
    H_S(G^\mathcal{S}) = - \sum\nolimits_{k=1}^{K} \frac{\operatorname{Vol}(C_k)}{\operatorname{Vol}(G^\mathcal{S})} \log \frac{\operatorname{Vol}(C_k)}{\operatorname{Vol}(G^\mathcal{S})},
\end{equation}
where $\operatorname{Vol}(C_k)$ represents the sum of degrees in cluster $C_k$. Based on this tree, we construct a structured textual prompt $\mathbf{t}_i$ for node $v_i$ that captures its structural role, such as: ``\textit{There are $K$ structural clusters. Node $v_i$ belongs to cluster $C_k$, which contains $N_K$ nodes, including $v_i$, \ldots, $v_j$.}”.
As shown in Figure~\ref{fig:prompt_tree}, the encoding tree guides prompt generation by selecting a node’s local cluster, identifying its peer nodes, and transforming this information into language models. This prompt is then embedded via BERT and fused:
\begin{equation}
    \mathbf{x}^{\mathcal{S}\prime}_i = \operatorname{SVD}(\mathbf{x}_i^\mathcal{S}) \oplus \operatorname{SVD}(\operatorname{BERT}(\mathbf{t}_i)),
\label{eq:2}
\end{equation}
where $\operatorname{SVD}(\cdot)$ is implemented by truncated SVD~\cite{stewart1993early} that aligns feature dimensions as $d_0$.

\subsubsection{Self-Supervised Information Bottleneck.}
Given aligned features $\mathbf{X}^{\mathcal{S}\prime}$, our goal is to learn the robust and compressed representations $\mathbf{Z}^{\mathcal{S}}\!=\!\mathcal{F}_{\boldsymbol{\Theta}}(\mathbf{X}^{\mathcal{S}\prime}, \mathbf{A}^{\mathcal{S}})\!\in\!\mathbb{R}^d$ that retain task-relevant semantics while suppressing noise. Following the Information Bottleneck (IB) principle~\cite{tishby1999information}, we formulate a self-supervised objective to maximize consistency between similar nodes while compressing redundant input information.
Formally, we define the self-supervised IB (SS-IB) objective as:
\begin{equation}
\label{eq:3}
    \mathcal{L}_{\text{SS-IB}} = -I(\mathbf{Z}^{\mathcal{S}}; \mathbf{Z}^{\mathcal{S}+}) + \beta\cdot I(\mathbf{Z}^{\mathcal{S}}; \mathbf{X}^{\mathcal{S}\prime}),
\end{equation}
where the prediction term $I(\mathbf{Z}^{\mathcal{S}}; \mathbf{Z}^{\mathcal{S}+})$ maximizes consistency between an anchor $\mathbf{Z}^{\mathcal{S}}$ and its positive sample $\mathbf{Z}^{\mathcal{S}+}$, which is sampled from its structural neighborhood based on the encoding tree. The compression term $I(\mathbf{Z}^{\mathcal{S}}; \mathbf{X}^{\mathcal{S}\prime})$ enforces compression by limiting the information retained from the input $\mathbf{X}^{\mathcal{S}\prime}$. $\beta$ is a balancing hyperparameter.

As directly optimizing Eq.~\eqref{eq:3} is generally intractable, we adopt variational bounds to implement $\mathcal{L}_{\text{SS-IB}}$.

\begin{Deriv}[Lower Bound of $I(\mathbf{Z}^{\mathcal{S}}; \mathbf{Z}^{\mathcal{S}+})$]
    Prediction term $I(\mathbf{Z}^{\mathcal{S}}; \mathbf{Z}^{\mathcal{S}+})$ is lower-bounded by InfoNCE~\cite{oord2018representation}. For each anchor $v_i$ and positive node $v_{i}^+$:
    \newpage
    \begin{align}
        \!\!\!\!\!\!I(\mathbf{Z}^{\mathcal{S}}; \mathbf{Z}^{\mathcal{S}+})&\geqslant  -\mathcal{L}_{\text{InfoNCE}}\notag\\
        &= \frac{1}{N^+} \sum\nolimits_{i=1}^{N^+}\log\! \frac{\exp(\langle\mathbf{Z}^{\mathcal{S}}_i, \mathbf{Z}^{\mathcal{S}+}_{i}\rangle / \tau)}{\sum_{j=0}^{N^-} \exp(\langle\mathbf{Z}^{\mathcal{S}}_i, \mathbf{Z}^{\mathcal{S}}_j\rangle / \tau)},\!\!\!\!
    \label{eq:deriv_1}
    \end{align}
    where $N^+$ and $N^-$ are the number of positive and negative samples, which are non-redundantly and consistently sampled from the source domain depending on whether there is a direct link~\cite{liu2023graphprompt}. $\tau$ is a temperature.
\label{deriv:deriv_1}
\end{Deriv}

\begin{Deriv}[Upper Bound of $I(\mathbf{Z}^{\mathcal{S}}; \mathbf{X}^{\mathcal{S}\prime})$]
    Compression term $I(\mathbf{Z}^{\mathcal{S}}; \mathbf{X}^{\mathcal{S}\prime})$ is upper-bounded by the KL divergence between posterior $q(\mathbf{Z}^{\mathcal{S}}|\mathbf{X}^{\mathcal{S}\prime})$ and prior $p(\mathbf{Z}^{\mathcal{S}})$:
    \begin{equation}
        I(\mathbf{Z}^{\mathcal{S}}; \mathbf{X}^{\mathcal{S}\prime})\leqslant \frac{1}{N} \sum\nolimits_{i=1}^{N}\operatorname{KL}[q(\mathbf{Z}_i^{\mathcal{S}}|\mathbf{X}_i^{\mathcal{S}\prime})~\|~p(\mathbf{Z}_i^{\mathcal{S}})],
    \label{eq:deriv_2}
    \end{equation}
    We assume a standard Gaussian prior $p(\mathbf{Z}^{\mathcal{S}}) = \mathcal{N}(\boldsymbol{0}, \mathbf{I})$, and apply the graph encoder to produce the mean $\boldsymbol{\mu}_i$ and the log-variance $\log\boldsymbol{\sigma}_i^2$ of $q(\mathbf{Z}^{\mathcal{S}}_i|\mathbf{X}_i^{\mathcal{S}\prime})$ for each node $v_i$, from which we sample $\mathbf{Z}^{\mathcal{S}}_i$ via reparameterization trick:
    \begin{equation}
        \mathbf{Z}^{\mathcal{S}}_i \sim q(\mathbf{Z}^{\mathcal{S}}_i|\mathbf{X}_j^{\mathcal{S}\prime}) = \mathcal{N}\big(\boldsymbol{\mu}_i, \operatorname{diag}(\boldsymbol{\sigma}_i^2)\big).
    \end{equation}
\label{deriv:deriv_2}
We provide detailed proofs in Appendix~\ref{sec:proof}.
\end{Deriv}

\subsubsection{Pre-training Objective.}
By integrating these bounds, we transform $\mathcal{L}_\text{SS-IB}$ in Eq.~\eqref{eq:3} into a tractable objective given $n$ source domain graphs with trade-off coefficient $\lambda_{\text{IB}}$:
\begin{equation}
\label{eq:7}
    \mathcal{L}_{\text{pretrain}}(\boldsymbol{\Theta}) = \frac{1}{n}\sum\nolimits_{D_i=1}^{n}\eqnmarkbox[black]{I1}{\mathcal{L}_{\text{InfoNCE}}} + \lambda_{\text{IB}} \eqnmarkbox[black]{I2}{\mathcal{L}_{\text{KL}}},
\end{equation}
\annotate[yshift=-0.1em]{below,right}{I1}{{\small{$\text{Eq.~\eqref{eq:deriv_1}}$}}}
\annotate[yshift=-0.1em]{below,right}{I2}{{\small{$\text{Eq.~\eqref{eq:deriv_2}}$}}}

\noindent after which the learned $\boldsymbol{\Theta}^\star$ is frozen.
Overall, pre-training phase leverages pre-trained language models as a bridge to transform structural priors into node features, enabling the capture of transferable knowledge that lays robust foundation for next expert routing and downstream fine-tuning.

\subsection{Expert Adaptive Routing with Negative Transfer Mitigation}
\label{sec:moe}
To prevent negative transfer, we introduce an expert adaptive routing mechanism, which not only ``selects'' helpful experts but also ``rejects'' negative ones via a learnable null expert.

\subsubsection{Gated Routing with Null-Expert.}
To enable fine-grained adaptation across multiple source domains while mitigating negative transfer, we introduce a gated routing network $\mathcal{R}_{\boldsymbol{\phi}}$ that dynamically modulates the contribution of each source expert from $\mathcal{F}_{\boldsymbol{\Theta}}^\star=\{f_{\boldsymbol{\Theta}_1}^\star, f_{\boldsymbol{\Theta}_2}^\star, \cdots, f_{\boldsymbol{\Theta}_n}^\star\}$ via a learned routing weights $\boldsymbol{\alpha} = \{\boldsymbol{\alpha}_1, \dots, \boldsymbol{\alpha}_n, \boldsymbol{\alpha}_{\text{null}}\}$, where the null expert explicitly captures and suppresses irrelevant knowledge.

Specifically, we first derive prototypes for the target and source domains. Let $\bar{\mathbf{Z}}^\mathcal{T}$ denote the mean-pooled embedding of $m$ ($m \ll n$) few-shot supporting samples (node or graph) in the target graph(s).
Similaryly, let $\{\bar{\mathbf{Z}}_i^\mathcal{S}\}_{i=1}^n$ represent the corresponding prototypes from each of the $n$ source domains. We define the cosine similarity vector $\mathbf{S} \in \mathbb{R}^n$:
\begin{equation}
\label{eq:8}
    \mathbf{S}_i = \langle \bar{\mathbf{Z}}^\mathcal{T}\!, \bar{\mathbf{Z}}_i^\mathcal{S} \rangle / \|\bar{\mathbf{Z}}^\mathcal{T}\!\| \!\cdot\! \|\bar{\mathbf{Z}}_i^\mathcal{S}\|,
\end{equation}
where $\mathbf{S}$ is passed through a learnable router $\mathcal{R}_{\boldsymbol{\phi}}$, yielding the unnormalized logits for the gating distribution:
\begin{equation}
\label{eq:9}
    \boldsymbol{\alpha} = \operatorname{Softmax}(\mathcal{R}_{\boldsymbol{\phi}}(\mathbf{S})) \in \Delta^{n+1},
\end{equation}
where $\Delta^{n+1}$ is the ($n\!+\!1$)-dimensional probability simplex.
Let $\mathbf{Z}_i^{\mathcal{S}} \in \mathbb{R}^{N_i \times d}$ be the node-level embeddings from the $i$-th expert, and let $\mathbf{Z}_{\text{null}}^{\mathcal{T}}$ denote the embedding generated by a shallow GCN trained solely on the target graph. The final task-specific representation is a convex combination:
\begin{equation}
\label{eq:routing_softmax}
    \mathbf{Z}_{\text{final}}^{\mathcal{T}} = \sum\nolimits_{D_i=1}^{n} \boldsymbol{\alpha}_i \cdot \mathbf{Z}_i^{\mathcal{S}} + \boldsymbol{\alpha}_{\text{null}} \cdot \mathbf{Z}_{\text{null}}^{\mathcal{T}},
\end{equation}
which enables the model to selectively integrate transferable domain knowledge, and further assign high mass to null experts when there is no source expert that aligns semantically or structurally, thus mitigating negative transfer.

\subsubsection{Uncertainty-Aware Routing Regularization.}
To inspire confident expert selection and avoid the overly diffuse gating, we incorporate an entropy-based regularization over the mixture-of-experts (MoE) architecture:
\begin{equation}
\label{eq:fuse_embeddings}
    \mathcal{L}_{\text{MoE}}(\boldsymbol{\phi}) = \mathcal{H}(\boldsymbol{\alpha}) = -\sum\nolimits_{j=1}^{k+1} \boldsymbol{\alpha}_j \log \boldsymbol{\alpha}_j,
\end{equation}
where $\mathcal{H}(\cdot)$ is the entropy. It penalizes uncertain mixtures and promotes sparse, decisive routing behaviors. In particular, when all source experts are irrelevant, this regularization amplifies the model’s preference for the null expert.

\subsection{Fine-tuning with Efficient Hierarchical Structure Optimization}
\label{sec:finetune}
While expert routing has aligned the source knowledge, performance still hinges on target graphs. Potential structural noise and adversarial perturbations can greatly hinder message passing. Though structure learning improves robustness~\cite{zugner2018adversarial, zhu2021deep}, it is often inefficient and coarse-grained. We thus propose a lightweight structure optimization strategy.

To ensure a seamless transition from pre-training to fine-tuning, we preserve structural consistency by reusing the entropy encoding tree to partition the target graph ${G}^\mathcal{T}$ into a set of clusters $\mathcal{C}^\mathcal{T}$. This partition forms the basis for the hierarchical structure optimization during fine-tuning, enabling targeted refinement at both the intra- and inter-cluster levels. While intra-cluster optimization enhances local structural fidelity, inter-cluster optimization focuses on global signal propagation and robustness.

\subsubsection{Intra-cluster Structure Learning.}
For each of cluster $c \!\in\! \mathcal{C}^\mathcal{T}$, we refine its internal topology by refining edges with multi-head attention. Let $\mathbf{Z}^\mathcal{T}_c$ denote the node embedding in cluster $c$, which are projected to $\mathbf{H}_c\!=\!\operatorname{ReLU}(\mathbf{Z}^\mathcal{T}_c \mathbf{W}_p)$. Attention matrix $\mathbf{S}_{\text{attn},c}$ is computed via:
\begin{equation}
\label{eq:12}
    \mathbf{S}_{\text{attn},c}  = \frac{1}{H} \sum\nolimits_{h=1}^{H} \operatorname{Softmax}\Big(\frac{\mathbf{Q}_h \mathbf{K}_h^\top}{\sqrt{d_k}}\Big), 
\end{equation}
where $\mathbf{Q}_h =\mathbf{Z}_c \mathbf{W}_h^Q$, and $\mathbf{K}_h = \mathbf{H}_c \mathbf{W}_h^K$.
To enforce the cross-head consistency, we introduce an uncertainty loss that penalizes variance among heads:
\begin{equation}
\label{eq:13}
    \mathcal{L}_{\text{uncertainty}} = \frac{\sum_{c \in \mathcal{C}^\mathcal{T}} \sum_{i,j \in c, i \neq j} \operatorname{Var}_{h=1}^{H}\big(\mathbf{S}_{ij}^{(h)}\big)}{|\{(i,j) | i,j \in c, c \in \mathcal{C}^\mathcal{T}, i \neq j\}|},
\end{equation}
where $\operatorname{Var}(\cdot)$ denotes variation.
Refined intra-cluster $\mathbf{A}^{\mathcal{T}'}_{\text{intra},c}$ fuses the original $\mathbf{A}^{\mathcal{T}}_c$ and attention-derived scores:
\begin{equation}
\label{eq:14}
    \mathbf{A}^{\mathcal{T}'}_{\text{intra},c} = (1-\mathbf{W}_c)\cdot\mathbf{A}_c^\mathcal{T} + \mathbf{W}_c  \cdot\mathbf{S}_{\text{attn}, c},
\end{equation}
where $\mathbf{W}_c$ is a learnable fusion weight per cluster.

\subsubsection{Inter-cluster Structure Learning.}
As intra-cluster learning focuses on refining dense local structures, inter-cluster learning aims to regulate potentially noisy global connections, which are critical for propagation but often include irrelevant edges.
To selectively retain informative inter-cluster dependencies, we approximate personalized propagation of neural predictions~\cite{gasteiger2018combining} to compute a soft influence matrix:
\begin{equation}
    \mathbf{S}^\mathcal{T} = (1-\alpha)\cdot\sum\nolimits_{t=0}^{T} (\alpha\cdot \tilde{\mathbf{A}}^\mathcal{T})^t,
\end{equation}
where $\tilde{\mathbf{A}}^\mathcal{T}$ is normalized adjacency.
We then derive a probabilistic pruning mask via thresholded activation to refine the original inter-cluster structure $\mathbf{A}_{\text{inter}}^\mathcal{T}$:
\begin{equation}
    \mathbf{A}^{\mathcal{T}'}_{\text{inter}} = \mathbf{A}_{\text{inter}}^\mathcal{T} \odot \mathbf{P}^\mathcal{T},~\mathbf{P}^\mathcal{T} = \sigma(\mathbf{S}^\mathcal{T} - \boldsymbol{\theta}_{\text{thres}}),
\end{equation}
where $\sigma$ is the Sigmoid and $\boldsymbol{\theta}_{\text{thres}}$ is a learnable threshold.
Let $\mathbf{A}^{\mathcal{T}'}_{\text{intra, full}}$ be the block-diagonal matrix formed by assembling the refined intra-cluster matrices $\{\mathbf{A}^{\mathcal{T}'}_{\text{intra},c}\}_{c \in \mathcal{C}^\mathcal{T}}$. Optimized $\mathbf{A}^{\mathcal{T}\prime}$ for target graph is then an adaptive combination:
\begin{equation}
\label{eq:17}
    \mathbf{A}^{\mathcal{T}'} = \mathbf{W}_s \cdot \mathbf{A}^{\mathcal{T}'}_{\text{intra, full}} + (1-\mathbf{W}_s) \cdot \mathbf{A}^{\mathcal{T}'}_{\text{inter}},
\end{equation}
where $\mathbf{W}_s$ is a learnable trade-off weights.

\subsubsection{Fine-tuning with Prompted Structure.}
Given $\mathbf{A}^{\mathcal{T}'}$,  we embed the learnable prompts to guide downstream adaptation. For each few-shot sample, the prompted embedding is:
\begin{equation}
\label{eq:get_embeddings}
    \mathbf{Z}_i^\mathcal{T} = \mathcal{F}_{\boldsymbol{\Theta}}^\star\big(\boldsymbol{\mathcal{P}}_{\boldsymbol{\Omega}} \odot \mathbf{X}_i^{\mathcal{T}'},\ \mathbf{A}^{\mathcal{T}'}\big)\in \mathbb{R}^d,
\end{equation}
where $\boldsymbol{\mathcal{P}}_{\boldsymbol{\Omega}}$ is the learnable prompt and $\mathbf{X}_i^{\mathcal{T}'}$ is the feature aligned similarly as Eq.~\eqref{eq:1}.
To maintain consistency with the pre-training objective, we leverage a universal task template based on contrastive loss that aligns query embeddings with class prototypes under the updated structure. Given $m$ few-shot support samples ${(G_i^\mathcal{T}, Y_i^\mathcal{T})}$, the objective is:
\begin{equation}
\label{eq:cls_loss}
    \mathcal{L}_{\text{cls}}(\boldsymbol{\Omega}) = -\sum\nolimits_{i=1}^m \log \frac{\exp\big(\big\langle\mathbf{Z}_i^\mathcal{T}, \bar{\mathbf{Z}}_{Y_i}^\mathcal{T}\big\rangle/\tau\big)}{\sum_{Y_j} \exp\big(\big\langle\mathbf{Z}_i^\mathcal{T},\bar{\mathbf{Z}}_{Y_j}^\mathcal{T}\big\rangle/\tau\big)},
\end{equation}
where $\bar{\mathbf{Z}}_{Y_j}^\mathcal{T}$ is prototype of $Y_j$ and $\langle\cdot,\!\cdot\rangle$ is the inner product.
Thus, the overall fine-tuning objective is:
\begin{equation}
\label{eq:finetune_loss}
    \mathcal{L}_{\text{finetune}}(\boldsymbol{\phi},\boldsymbol{\Omega})=\mathcal{L}_{\text{cls}}+\lambda_m \cdot \mathcal{L}_{\text{MoE}} + \lambda_u \cdot \mathcal{L}_{\text{uncertainty}},
\end{equation}
where $\lambda_1$, $\lambda_2$ are hyperparameters.

\subsection{Complexity Analysis}
Suppose $f$ is an $L$-layer GNN. Then, the per-epoch computational complexity of the pre-training stage is $\mathcal{O}(nL|\mathcal{E^S}|d)$, mainly arising from the SS-IB pre-training objective over $n$ source graphs. The per-epoch complexity of the fine-tuning stage is $\mathcal{O}(nL|\mathcal{E^T}|d + \sum_{c}|\mathcal{V}^\mathcal{T}_c|^2)$, where the main cost comes from the encoding process and intra-cluster attention (with $|\mathcal{V}^\mathcal{T}_c|$ denoting the number of nodes in cluster $c$). The overall complexity is comparable to that of baselines such as MDGPT, while the structure optimization is more efficient than MDGFM. More details can be found in Appendix~\ref{sec:alg}.

\section{Experiment}
\label{sec:exp}
We evaluate the proposed \modelname~with the following research questions:
\begin{itemize}[leftmargin=*]
    \item \textit{\textbf{RQ1:}} How robust is \modelname~against random noise and adversarial attacks on features and structures? 
    \item \textit{\textbf{RQ2:}} How does the performance change as the intensity of noise or attack severity increases?
    \item \textit{\textbf{RQ3:}} Which section contributes most to the performance?
    \item \textit{\textbf{RQ4:}} How sensitive to hyperparameters fluctuation?
    \item \textit{\textbf{RQ5:}} How well does pre-training support transferable and task-adaptive representations?
\end{itemize}
Additional settings and results are in Appendix~\ref{sec:exp_setting} and~\ref{sec:additional_results}.
\subsection{Experimental Settings}
\subsubsection{Datasets.}
To highlight multi-domain pre-training capability, we adopt \textbf{seven} benchmark datasets from \textbf{three} domains:
\begin{itemize}[leftmargin=*]
    \item \textbf{Citation:} \texttt{Cora} \cite{mccallum2000automating},  \texttt{CiteSeer} \cite{giles1998citeseer},  \texttt{PubMed} \cite{sen2008collective}, \texttt{ogbn-arXiv} \cite{hu2020open}.
    \item \textbf{Products:} \texttt{ogbn-Tech}, \texttt{ogbn-Home} \cite{hu2020open}.
    \item \textbf{Web Page:} \texttt{Wiki-CS} \cite{mernyei2020wikics}.
\end{itemize}
Dataset statistics are summarized in Table~\ref{tab:datasets}.
\subsubsection{Baselines.}
Include \textbf{nine} baselines from \textbf{four} categories:
\begin{itemize}[leftmargin=*]
    \item \texttt{GCN} (Kipf et al. 2017), \texttt{GAT} \cite{veličković2018graph}.
    \item \textbf{Single Domain:} \texttt{DGI} \cite{velickovic2019deep}, \texttt{GraphCL} \cite{you2020graphcl}, \texttt{GraphPrompt} \cite{liu2023graphprompt}.
    \item \textbf{Multi Domain:} \texttt{MDGPT} \cite{yu2024text}, \texttt{GCOPE} \cite{zhao2024all}, \texttt{GraphBridge} \cite{ju2025graphbridge}.
    \item \textbf{Robust GFM:} \texttt{MDGFM} \cite{wang2025multidomain}.
\end{itemize}
\subsubsection{Pre-training and Fine-tuning.}
We focus on \textbf{two} settings:
\begin{itemize}[leftmargin=*]
    \item \textbf{Cross-Datasets:} Pre-training and fine-tuning on \textbf{different} datasets from the \textbf{same} domain. While one dataset is selected as the target, the rest are sources.
    \item \textbf{Cross-Domain:} Pre-training and fine-tuning on \textbf{different} datasets from the \textbf{different} domain.
\end{itemize}
\subsubsection{Few-shot Fine-tuning.}
We adopt a $C$-way 5-shot setting for each class. For graph classification, ego-graphs centered on target nodes are used, inheriting center label. Accuracy with standard deviation is used for evaluation over 20 runs.

\subsubsection{Noise and Attack.} 
We assess robustness in \textbf{two} scenarios:
\begin{itemize}[leftmargin=*]
    \item \textbf{Non-targeted Attack (Noise):} We inject random noise by perturbing either the structure or node features, with a perturbation rate $\lambda \in~$\{0.4, 0.8\}. \textbf{(1)} For node feature attacks, we randomly inject Gaussian noise. \textbf{(2)} For structure attacks, we randomly remove edges from the graph. 
        
    \item \textbf{Targeted Attack:} We employ the powerful NETTACK \cite{zugner2018adversarial} platform to perform attacks on specific nodes with the default number of perturbations $p \in~$\{1, 2, 3\}. \textbf{(1)} For evasion attacks, the model is finetuned on a clean support set and attacked during testing on the query set. \textbf{(2)} For poisoning attacks, the entire support set is perturbed before fine-tuning. 
\end{itemize}
\begin{table*}[!t]
\centering
\resizebox{\linewidth}{!}{
\setlength{\tabcolsep}{1pt}
{\renewcommand{\arraystretch}{0.95}
    \begin{tabular}{lcccccccccccccccc}
    \toprule
    \multirow{2}[6]{*}{\textbf{Source}} & \multicolumn{4}{c}{\textbf{Cross-Dataset}} & \multicolumn{12}{c}{\textbf{Cross-Domain}} \\
\cmidrule{2-5} \cmidrule(l){6-17}           & \multicolumn{4}{c}{\makecell{\texttt{Cora} \quad \texttt{CiteSeer}\\ \texttt{ogbn-Home}\quad \texttt{Wiki-CS}}}       & \multicolumn{4}{c}{\makecell{\texttt{Cora} \quad \texttt{CiteSeer}\\ \texttt{PubMed}\quad\texttt{Wiki-CS}}}       & \multicolumn{4}{c}{\makecell{\texttt{Cora} \quad \texttt{CiteSeer}\\ \texttt{PubMed}\quad\texttt{ogbn-Home}}}       & \multicolumn{4}{c}{\makecell{\texttt{ogbn-Home}\quad\texttt{ogbn-Tech}\\ \texttt{Wiki-CS}}} \\
    \cmidrule{1-5} \cmidrule(l){6-17} 
    \textbf{Target} & \multicolumn{4}{c}{\texttt{PubMed}}   & \multicolumn{4}{c}{\texttt{ogbn-Home}}      & \multicolumn{4}{c}{\texttt{Wiki-CS}}      & \multicolumn{4}{c}{\texttt{ogbn-arxiv}} \\
    \midrule
    \textbf{Noise \& Attacks} & feat. & struct. & evas. & pois. & feat. & struct. & evas. & pois. & feat. & struct. & evas. & pois. & feat. & struct. & evas. & pois. \\
    \midrule
    \texttt{GCN} (backbone)   & 44.6\scalebox{0.6}{±8.7\textcolor{white}{0}}  & 40.9\scalebox{0.6}{±9.8\textcolor{white}{0}}  & 36.7\scalebox{0.6}{±11.0}  & 32.4\scalebox{0.6}{±9.9\textcolor{white}{0}}  & 42.9\scalebox{0.6}{±11.9}  & 51.0\scalebox{0.6}{±12.9}  & 47.8\scalebox{0.6}{±13.6}  & 46.4\scalebox{0.6}{±13.4}  & 38.4\scalebox{0.6}{±6.9\textcolor{white}{0}}  & 42.5\scalebox{0.6}{±9.9\textcolor{white}{0}}  & 33.8\scalebox{0.6}{±7.6\textcolor{white}{0}}  & 38.5\scalebox{0.6}{±8.8\textcolor{white}{0}}  & 39.3\scalebox{0.6}{±5.2\textcolor{white}{0}}  & 38.8\scalebox{0.6}{±4.7\textcolor{white}{0}}  & 36.5\scalebox{0.6}{±5.3\textcolor{white}{0}}  & 38.3\scalebox{0.6}{±6.1\textcolor{white}{0}}  \\
    \texttt{GAT}   & 43.9\scalebox{0.6}{±7.8\textcolor{white}{0}}  & 44.2\scalebox{0.6}{±9.1\textcolor{white}{0}}  & 37.5\scalebox{0.6}{±8.6\textcolor{white}{0}}  & 38.3\scalebox{0.6}{±8.3\textcolor{white}{0}}  & 43.9\scalebox{0.6}{±9.3\textcolor{white}{0}}  & 53.5\scalebox{0.6}{±6.7\textcolor{white}{0}}  & 46.5\scalebox{0.6}{±6.3\textcolor{white}{0}}  & 56.0\scalebox{0.6}{±7.7\textcolor{white}{0}}  & 37.1\scalebox{0.6}{±6.2\textcolor{white}{0}}  & 46.6\scalebox{0.6}{±4.9\textcolor{white}{0}}  & 34.4\scalebox{0.6}{±5.6\textcolor{white}{0}}  & 37.6\scalebox{0.6}{±6.7\textcolor{white}{0}}  & 40.9\scalebox{0.6}{±5.7\textcolor{white}{0}}  & 41.4\scalebox{0.6}{±5.6\textcolor{white}{0}}  & 34.5\scalebox{0.6}{±6.7\textcolor{white}{0}}  & 38.5\scalebox{0.6}{±5.4\textcolor{white}{0}}  \\
    \midrule
    \texttt{DGI}   & 46.7\scalebox{0.6}{±7.9\textcolor{white}{0}}  & 41.2\scalebox{0.6}{±7.5\textcolor{white}{0}}  & 42.1\scalebox{0.6}{±7.6\textcolor{white}{0}}  & 35.9\scalebox{0.6}{±8.9\textcolor{white}{0}}  & 48.7\scalebox{0.6}{±7.0\textcolor{white}{0}}  & 61.0\scalebox{0.6}{±6.2\textcolor{white}{0}}  & 54.0\scalebox{0.6}{±4.9\textcolor{white}{0}}  & 51.0\scalebox{0.6}{±10.6}  & 45.6\scalebox{0.6}{±5.6\textcolor{white}{0}}  & 44.1\scalebox{0.6}{±7.7\textcolor{white}{0}}  & 41.1\scalebox{0.6}{±6.5\textcolor{white}{0}}  & 45.9\scalebox{0.6}{±5.1\textcolor{white}{0}}  & 39.3\scalebox{0.6}{±4.2\textcolor{white}{0}}  & 46.6\scalebox{0.6}{±4.9\textcolor{white}{0}}  & 36.7\scalebox{0.6}{±5.4\textcolor{white}{0}}  & 44.8\scalebox{0.6}{±4.5\textcolor{white}{0}}  \\
    \texttt{GraphCL} & 54.9\scalebox{0.6}{±9.5\textcolor{white}{0}}  & 48.9\scalebox{0.6}{±8.8\textcolor{white}{0}}  & 42.7\scalebox{0.6}{±7.8\textcolor{white}{0}}  & 37.5\scalebox{0.6}{±7.6\textcolor{white}{0}}  & 47.7\scalebox{0.6}{±9.1\textcolor{white}{0}}  & 48.9\scalebox{0.6}{±7.8\textcolor{white}{0}}  & 55.9\scalebox{0.6}{±6.6\textcolor{white}{0}}  & 53.0\scalebox{0.6}{±9.1\textcolor{white}{0}}  & 42.8\scalebox{0.6}{±6.2\textcolor{white}{0}}  & 49.3\scalebox{0.6}{±5.6\textcolor{white}{0}}  & 42.4\scalebox{0.6}{±6.1\textcolor{white}{0}}  & 41.3\scalebox{0.6}{±4.7\textcolor{white}{0}}  & 38.8\scalebox{0.6}{±4.6\textcolor{white}{0}}  & 43.0\scalebox{0.6}{±4.9\textcolor{white}{0}}  & 37.8\scalebox{0.6}{±4.5\textcolor{white}{0}}  & 30.0\scalebox{0.6}{±5.5\textcolor{white}{0}}  \\
    \texttt{GraphPrompt} & 47.8\scalebox{0.6}{±8.8\textcolor{white}{0}}  & 45.2\scalebox{0.6}{±8.1\textcolor{white}{0}}  & 45.3\scalebox{0.6}{±7.5\textcolor{white}{0}}  & 39.6\scalebox{0.6}{±6.1\textcolor{white}{0}}  & 56.0\scalebox{0.6}{±7.3\textcolor{white}{0}}  & 58.0\scalebox{0.6}{±8.4\textcolor{white}{0}}  & 55.3\scalebox{0.6}{±8.3\textcolor{white}{0}}  & 53.5\scalebox{0.6}{±6.6\textcolor{white}{0}}  & 51.9\scalebox{0.6}{±5.9\textcolor{white}{0}}  & 41.5\scalebox{0.6}{±7.1\textcolor{white}{0}}  & 39.4\scalebox{0.6}{±8.1\textcolor{white}{0}}  & 32.5\scalebox{0.6}{±3.7\textcolor{white}{0}}  & 41.4\scalebox{0.6}{±4.6\textcolor{white}{0}}  & 43.4\scalebox{0.6}{±5.2\textcolor{white}{0}}  & 39.0\scalebox{0.6}{±4.7\textcolor{white}{0}}  & 42.0\scalebox{0.6}{±4.0\textcolor{white}{0}}  \\
    \midrule
    \texttt{MDGPT} & 48.6\scalebox{0.6}{±4.2\textcolor{white}{0}}  & 56.5\scalebox{0.6}{±5.5\textcolor{white}{0}}  & 52.0\scalebox{0.6}{±6.4\textcolor{white}{0}}  & 42.1\scalebox{0.6}{±5.8\textcolor{white}{0}}  & 59.3\scalebox{0.6}{±25.1}  & 54.8\scalebox{0.6}{±25.1}  & 54.0\scalebox{0.6}{±24.1}  & 57.4\scalebox{0.6}{±16.4}  & 42.1\scalebox{0.6}{±7.0\textcolor{white}{0}}  & 50.4\scalebox{0.6}{±5.0\textcolor{white}{0}}  & 49.5\scalebox{0.6}{±8.9\textcolor{white}{0}}  & 40.4\scalebox{0.6}{±7.2\textcolor{white}{0}}  & 42.5\scalebox{0.6}{±7.2\textcolor{white}{0}}  & 46.3\scalebox{0.6}{±6.7\textcolor{white}{0}}  & 45.6\scalebox{0.6}{±5.3\textcolor{white}{0}}  & 48.3\scalebox{0.6}{±4.7\textcolor{white}{0}}  \\
    \texttt{GCOPE} & 53.2\scalebox{0.6}{±13.3}  & 55.6\scalebox{0.6}{±11.4}  & 48.7\scalebox{0.6}{±10.9}  & 44.7\scalebox{0.6}{±9.4\textcolor{white}{0}}  & 57.0\scalebox{0.6}{±23.5}  & 56.0\scalebox{0.6}{±24.4}  & 55.6\scalebox{0.6}{±24.3}  & 50.0\scalebox{0.6}{±23.9}  & 46.6\scalebox{0.6}{±9.3\textcolor{white}{0}}  & 46.8\scalebox{0.6}{±11.8}  & 42.0\scalebox{0.6}{±10.6}  & 43.6\scalebox{0.6}{±11.4}  & 48.7\scalebox{0.6}{±5.9\textcolor{white}{0}}  & 50.0\scalebox{0.6}{±7.2\textcolor{white}{0}}  & 39.1\scalebox{0.6}{±5.9\textcolor{white}{0}}  & 49.8\scalebox{0.6}{±7.6\textcolor{white}{0}}  \\
    \texttt{GraphBridge} & 51.1\scalebox{0.6}{±6.8\textcolor{white}{0}}  & 52.4\scalebox{0.6}{±4.3\textcolor{white}{0}}  & 44.0\scalebox{0.6}{±10.1}  & 46.9\scalebox{0.6}{±7.4\textcolor{white}{0}}  & 63.0\scalebox{0.6}{±4.5\textcolor{white}{0}}  & 62.2\scalebox{0.6}{±5.2\textcolor{white}{0}}  & 57.1\scalebox{0.6}{±3.7\textcolor{white}{0}}  & 51.3\scalebox{0.6}{±6.6\textcolor{white}{0}}  & 50.1\scalebox{0.6}{±6.6\textcolor{white}{0}}  & 47.3\scalebox{0.6}{±5.7\textcolor{white}{0}}  & 43.2\scalebox{0.6}{±10.1}  & 42.4\scalebox{0.6}{±8.3\textcolor{white}{0}}  & 46.5\scalebox{0.6}{±5.7\textcolor{white}{0}}  & 47.3\scalebox{0.6}{±6.6\textcolor{white}{0}}  & 47.5\scalebox{0.6}{±6.2\textcolor{white}{0}}  & 44.5\scalebox{0.6}{±6.0\textcolor{white}{0}}  \\
    \midrule
    \texttt{MDGFM} & \underline{57.3\scalebox{0.6}{±6.7\textcolor{white}{0}}}  & \underline{58.4\scalebox{0.6}{±7.3\textcolor{white}{0}}}  & \underline{53.4\scalebox{0.6}{±7.3\textcolor{white}{0}}}  & \underline{50.8\scalebox{0.6}{±5.2\textcolor{white}{0}}}  & \underline{65.0\scalebox{0.6}{±15.9}}  & \underline{65.3\scalebox{0.6}{±17.0}}  & \underline{62.9\scalebox{0.6}{±15.3}}  & \underline{62.1\scalebox{0.6}{±16.2}}  & \underline{53.2\scalebox{0.6}{±6.9\textcolor{white}{0}}}  & \underline{52.0\scalebox{0.6}{±5.2\textcolor{white}{0}}}  & \underline{50.1\scalebox{0.6}{±6.2\textcolor{white}{0}}}  & \underline{46.4\scalebox{0.6}{±5.7\textcolor{white}{0}}}  & \underline{55.9\scalebox{0.6}{±4.1\textcolor{white}{0}}}  & \underline{55.7\scalebox{0.6}{±4.9\textcolor{white}{0}}}  & \underline{50.8\scalebox{0.6}{±5.0\textcolor{white}{0}}}  & \underline{50.4\scalebox{0.6}{±4.7\textcolor{white}{0}}}  \\
    \midrule
    \textbf{\modelname~(ours)} & \textbf{60.0\scalebox{0.6}{±5.2\textcolor{white}{0}}}  & \textbf{60.1\scalebox{0.6}{±1.4\textcolor{white}{0}}}  & \textbf{56.9\scalebox{0.6}{±9.8\textcolor{white}{0}}}  & \textbf{54.5\scalebox{0.6}{±1.2\textcolor{white}{0}}}  & \textbf{68.9\scalebox{0.6}{±6.2\textcolor{white}{0}}}  & \textbf{69.0\scalebox{0.6}{±5.2\textcolor{white}{0}}}  & \textbf{65.9\scalebox{0.6}{±2.7\textcolor{white}{0}}}  & \textbf{64.0\scalebox{0.6}{±1.3\textcolor{white}{0}}}  & \textbf{55.9\scalebox{0.6}{±5.3\textcolor{white}{0}}}  & \textbf{55.9\scalebox{0.6}{±1.4\textcolor{white}{0}}}  & \textbf{53.3\scalebox{0.6}{±4.9\textcolor{white}{0}}}  & \textbf{50.6\scalebox{0.6}{±1.2\textcolor{white}{0}}}  & \textbf{57.9\scalebox{0.6}{±7.3\textcolor{white}{0}}}  & \textbf{57.8\scalebox{0.6}{±1.4\textcolor{white}{0}}}  & \textbf{55.0\scalebox{0.6}{±5.8\textcolor{white}{0}}}  & \textbf{53.0\scalebox{0.6}{±1.2\textcolor{white}{0}}}  \\
    \bottomrule
    \end{tabular}%
    }
}
\vspace{-0.2cm}
\caption{Accuracy (\% ± std. for 20 runs) of \textbf{5-shot node classification}. Best scores are in \textbf{bold}, runner-ups are \underline{underlined}. ``feat.'' and ``struct.'' denote non-targeted attacks ($\lambda=$~0.4), while ``evas.'' and ``pois.'' denote targeted attacks ($p=$~3).}
\label{tab:res_main_node}
\end{table*}

\begin{table*}[!t]
\vspace{-0.1cm}
\centering
\resizebox{\linewidth}{!}{
\setlength{\tabcolsep}{1pt}
{\renewcommand{\arraystretch}{0.95}
    \begin{tabular}{lcccccccccccccccc}
    \toprule
    \multirow{2}[6]{*}{\textbf{Source}} & \multicolumn{8}{c}{\textbf{Cross-Dataset}} & \multicolumn{8}{c}{\textbf{Cross-Domain}} \\
\cmidrule{2-9} \cmidrule(l){10-17}           & \multicolumn{4}{c}{\makecell{\texttt{CiteSeer} \quad \texttt{PubMed}\\ \texttt{ogbn-Home}\quad \texttt{Wiki-CS}}}       & \multicolumn{4}{c}{\makecell{\texttt{Cora}~\texttt{CiteSeer}~\texttt{PubMed}\\ \texttt{ogbn-Home}\quad\texttt{Wiki-CS}}}       & \multicolumn{4}{c}{\makecell{\texttt{Cora} \quad \texttt{CiteSeer}\\ \texttt{PubMed}\quad\texttt{ogbn-Home}}}       & \multicolumn{4}{c}{\makecell{\texttt{ogbn-Home}\quad\texttt{ogbn-Tech}\\ \texttt{Wiki-CS}}} \\
    \cmidrule{1-9} \cmidrule(l){10-17} 
    \textbf{Target} & \multicolumn{4}{c}{\texttt{Cora}}   & \multicolumn{4}{c}{\texttt{ogbn-Tech}}      & \multicolumn{4}{c}{\texttt{Wiki-CS}}      & \multicolumn{4}{c}{\texttt{ogbn-arxiv}} \\
    \midrule
    \textbf{Noise \& Attacks} & feat. & struct. & evas. & pois. & feat. & struct. & evas. & pois. & feat. & struct. & evas. & pois. & feat. & struct. & evas. & pois. \\
    \midrule
    \texttt{GCN} (backbone)   & 44.9\scalebox{0.6}{±8.0\textcolor{white}{0}}  & 51.2\scalebox{0.6}{±5.8\textcolor{white}{0}}  & 42.9\scalebox{0.6}{±5.8\textcolor{white}{0}}  & 42.3\scalebox{0.6}{±7.4\textcolor{white}{0}}  & 68.7\scalebox{0.6}{±12.9}  & 67.4\scalebox{0.6}{±11.6}  & 60.7\scalebox{0.6}{±10.6}  & 59.2\scalebox{0.6}{±9.8\textcolor{white}{0}}  & 52.6\scalebox{0.6}{±10.9}  & 43.8\scalebox{0.6}{±11.7}  & 40.8\scalebox{0.6}{±6.7\textcolor{white}{0}}  & 41.4\scalebox{0.6}{±9.6\textcolor{white}{0}}  & 38.4\scalebox{0.6}{±9.0\textcolor{white}{0}}  & 41.9\scalebox{0.6}{±10.1}  & 32.8\scalebox{0.6}{±7.5\textcolor{white}{0}}  & 35.7\scalebox{0.6}{±5.8\textcolor{white}{0}}  \\
    \texttt{GAT}   & 47.1\scalebox{0.6}{±5.3\textcolor{white}{0}}  & 48.2\scalebox{0.6}{±5.8\textcolor{white}{0}}  & 45.4\scalebox{0.6}{±5.4\textcolor{white}{0}}  & 48.2\scalebox{0.6}{±6.8\textcolor{white}{0}}  & 71.1\scalebox{0.6}{±8.4\textcolor{white}{0}}  & 69.0\scalebox{0.6}{±10.0}  & 64.6\scalebox{0.6}{±10.3}  & 59.1\scalebox{0.6}{±9.7\textcolor{white}{0}}  & 46.6\scalebox{0.6}{±8.2\textcolor{white}{0}}  & 45.2\scalebox{0.6}{±4.0\textcolor{white}{0}}  & 45.7\scalebox{0.6}{±5.4\textcolor{white}{0}}  & 48.8\scalebox{0.6}{±5.7\textcolor{white}{0}}  & 44.5\scalebox{0.6}{±9.2\textcolor{white}{0}}  & 43.8\scalebox{0.6}{±8.5\textcolor{white}{0}}  & 40.1\scalebox{0.6}{±8.7\textcolor{white}{0}}  & 42.2\scalebox{0.6}{±7.7\textcolor{white}{0}}  \\
    \midrule
    \texttt{DGI}   & 49.6\scalebox{0.6}{±5.4\textcolor{white}{0}}  & 53.0\scalebox{0.6}{±5.2\textcolor{white}{0}}  & 54.6\scalebox{0.6}{±6.5\textcolor{white}{0}}  & 50.7\scalebox{0.6}{±5.1\textcolor{white}{0}}  & 71.7\scalebox{0.6}{±7.0\textcolor{white}{0}}  & 71.9\scalebox{0.6}{±8.3\textcolor{white}{0}}  & 70.2\scalebox{0.6}{±7.7\textcolor{white}{0}}  & 62.9\scalebox{0.6}{±7.4\textcolor{white}{0}}  & 52.0\scalebox{0.6}{±5.7\textcolor{white}{0}}  & 49.2\scalebox{0.6}{±4.8\textcolor{white}{0}}  & 45.1\scalebox{0.6}{±5.1\textcolor{white}{0}}  & 49.0\scalebox{0.6}{±4.5\textcolor{white}{0}}  & 45.5\scalebox{0.6}{±7.4\textcolor{white}{0}}  & 47.4\scalebox{0.6}{±6.8\textcolor{white}{0}}  & 43.4\scalebox{0.6}{±5.9\textcolor{white}{0}}  & 42.4\scalebox{0.6}{±5.2\textcolor{white}{0}}  \\
    \texttt{GraphCL} & 54.5\scalebox{0.6}{±4.6\textcolor{white}{0}}  & 58.5\scalebox{0.6}{±5.5\textcolor{white}{0}}  & 52.2\scalebox{0.6}{±6.6\textcolor{white}{0}}  & 43.9\scalebox{0.6}{±5.5\textcolor{white}{0}}  & 77.2\scalebox{0.6}{±5.2\textcolor{white}{0}}  & 76.9\scalebox{0.6}{±6.3\textcolor{white}{0}}  & 73.3\scalebox{0.6}{±6.6\textcolor{white}{0}}  & 54.0\scalebox{0.6}{±6.8\textcolor{white}{0}}  & 61.6\scalebox{0.6}{±6.3\textcolor{white}{0}}  & 50.5\scalebox{0.6}{±3.5\textcolor{white}{0}}  & 48.9\scalebox{0.6}{±4.2\textcolor{white}{0}}  & 43.1\scalebox{0.6}{±3.7\textcolor{white}{0}}  & 51.7\scalebox{0.6}{±6.6\textcolor{white}{0}}  & 45.3\scalebox{0.6}{±6.1\textcolor{white}{0}}  & 42.8\scalebox{0.6}{±5.9\textcolor{white}{0}}  & 49.5\scalebox{0.6}{±4.9\textcolor{white}{0}}  \\
    \texttt{GraphPrompt} & \underline{69.0\scalebox{0.6}{±5.8\textcolor{white}{0}}}  & 57.1\scalebox{0.6}{±5.3\textcolor{white}{0}}  & 50.3\scalebox{0.6}{±4.8\textcolor{white}{0}}  & 50.8\scalebox{0.6}{±4.8\textcolor{white}{0}}  & 83.0\scalebox{0.6}{±9.6\textcolor{white}{0}}  & 73.8\scalebox{0.6}{±10.2}  & 75.3\scalebox{0.6}{±9.4\textcolor{white}{0}}  & 61.1\scalebox{0.6}{±9.5\textcolor{white}{0}}  & 52.3\scalebox{0.6}{±6.3\textcolor{white}{0}}  & 60.1\scalebox{0.6}{±4.2\textcolor{white}{0}}  & 47.8\scalebox{0.6}{±5.0\textcolor{white}{0}}  & 47.2\scalebox{0.6}{±5.0\textcolor{white}{0}}  & 50.5\scalebox{0.6}{±6.6\textcolor{white}{0}}  & 48.6\scalebox{0.6}{±5.0\textcolor{white}{0}}  & 43.2\scalebox{0.6}{±5.6\textcolor{white}{0}}  & 41.2\scalebox{0.6}{±4.9\textcolor{white}{0}}  \\
    \midrule
    \texttt{MDGPT} & 63.5\scalebox{0.6}{±6.2\textcolor{white}{0}}  & 64.2\scalebox{0.6}{±6.2\textcolor{white}{0}}  & 60.0\scalebox{0.6}{±8.1\textcolor{white}{0}}  & 53.9\scalebox{0.6}{±6.1\textcolor{white}{0}}  & 73.3\scalebox{0.6}{±8.7\textcolor{white}{0}}  & 79.1\scalebox{0.6}{±6.4\textcolor{white}{0}}  & 73.6\scalebox{0.6}{±7.3\textcolor{white}{0}}  & 68.8\scalebox{0.6}{±8.9\textcolor{white}{0}}  & 56.0\scalebox{0.6}{±8.3\textcolor{white}{0}}  & 53.6\scalebox{0.6}{±7.4\textcolor{white}{0}}  & 55.2\scalebox{0.6}{±7.5\textcolor{white}{0}}  & 49.0\scalebox{0.6}{±8.3\textcolor{white}{0}}  & 47.1\scalebox{0.6}{±6.8\textcolor{white}{0}}  & 52.9\scalebox{0.6}{±8.5\textcolor{white}{0}}  & 50.1\scalebox{0.6}{±6.1\textcolor{white}{0}}  & 43.8\scalebox{0.6}{±6.3\textcolor{white}{0}}  \\
    \texttt{GCOPE} & 59.2\scalebox{0.6}{±10.1}  & 63.5\scalebox{0.6}{±7.1\textcolor{white}{0}}  & 58.7\scalebox{0.6}{±11.0}  & 58.5\scalebox{0.6}{±8.3\textcolor{white}{0}}  & 76.3\scalebox{0.6}{±11.2}  & 81.5\scalebox{0.6}{±8.5\textcolor{white}{0}}  & 72.1\scalebox{0.6}{±6.5\textcolor{white}{0}}  & 79.6\scalebox{0.6}{±6.6\textcolor{white}{0}}  & 50.1\scalebox{0.6}{±12.1}  & 54.5\scalebox{0.6}{±17.6}  & 52.4\scalebox{0.6}{±14.7}  & 40.1\scalebox{0.6}{±10.6}  & 57.5\scalebox{0.6}{±12.9}  & 56.8\scalebox{0.6}{±12.6}  & 52.8\scalebox{0.6}{±12.1}  & 46.5\scalebox{0.6}{±9.7\textcolor{white}{0}}  \\
    \texttt{GraphBridge} & 62.4\scalebox{0.6}{±5.0\textcolor{white}{0}}  & 61.0\scalebox{0.6}{±4.8\textcolor{white}{0}}  & 62.1\scalebox{0.6}{±5.7\textcolor{white}{0}}  & 56.6\scalebox{0.6}{±5.0\textcolor{white}{0}}  & 81.0\scalebox{0.6}{±7.5\textcolor{white}{0}}  & 80.8\scalebox{0.6}{±10.6}  & 76.9\scalebox{0.6}{±7.7\textcolor{white}{0}}  & 73.9\scalebox{0.6}{±7.0\textcolor{white}{0}}  & 61.5\scalebox{0.6}{±3.6\textcolor{white}{0}}  & 56.5\scalebox{0.6}{±3.6\textcolor{white}{0}}  & 58.6\scalebox{0.6}{±3.6\textcolor{white}{0}}  & 48.6\scalebox{0.6}{±3.6\textcolor{white}{0}}  & 54.7\scalebox{0.6}{±9.2\textcolor{white}{0}}  & 57.9\scalebox{0.6}{±9.4\textcolor{white}{0}}  & 55.0\scalebox{0.6}{±6.5\textcolor{white}{0}}  & 42.9\scalebox{0.6}{±7.7\textcolor{white}{0}}  \\
    \midrule
    \texttt{MDGFM} & 65.9\scalebox{0.6}{±4.7\textcolor{white}{0}}  & \underline{67.8\scalebox{0.6}{±4.9\textcolor{white}{0}}}  & \underline{65.8\scalebox{0.6}{±6.3\textcolor{white}{0}}}  & \underline{62.6\scalebox{0.6}{±5.8\textcolor{white}{0}}}  & \underline{85.3\scalebox{0.6}{±9.7\textcolor{white}{0}}}  & \underline{85.9\scalebox{0.6}{±7.6\textcolor{white}{0}}}  & \underline{83.1\scalebox{0.6}{±8.8\textcolor{white}{0}}}  & \textbf{83.5\scalebox{0.6}{±9.6\textcolor{white}{0}}}  & \textbf{66.4\scalebox{0.6}{±5.2\textcolor{white}{0}}}  & \underline{62.9\scalebox{0.6}{±4.6\textcolor{white}{0}}}  & \underline{61.5\scalebox{0.6}{±6.7\textcolor{white}{0}}}  & \underline{56.5\scalebox{0.6}{±8.0\textcolor{white}{0}}}  & \underline{62.0\scalebox{0.6}{±6.0\textcolor{white}{0}}}  & \underline{63.0\scalebox{0.6}{±5.0\textcolor{white}{0}}}  & \underline{59.2\scalebox{0.6}{±5.5\textcolor{white}{0}}}  & \underline{56.7\scalebox{0.6}{±6.1\textcolor{white}{0}}}  \\
    \midrule
    \textbf{\modelname~(ours)} & \textbf{69.5\scalebox{0.6}{±6.5\textcolor{white}{0}}}  & \textbf{69.7\scalebox{0.6}{±5.5\textcolor{white}{0}}}  & \textbf{68.4\scalebox{0.6}{±1.5\textcolor{white}{0}}}  & \textbf{64.6\scalebox{0.6}{±1.4\textcolor{white}{0}}}  & \textbf{87.7\scalebox{0.6}{±9.5\textcolor{white}{0}}}  & \textbf{87.8\scalebox{0.6}{±9.4\textcolor{white}{0}}}  & \textbf{86.7\scalebox{0.6}{±6.5\textcolor{white}{0}}}  & \underline{82.6\scalebox{0.6}{±1.4\textcolor{white}{0}}}  & \underline{64.4\scalebox{0.6}{±8.5\textcolor{white}{0}}}  & \textbf{64.2\scalebox{0.6}{±6.4\textcolor{white}{0}}}  & \textbf{63.5\scalebox{0.6}{±3.5\textcolor{white}{0}}}  & \textbf{59.0\scalebox{0.6}{±1.2\textcolor{white}{0}}}  & \textbf{63.9\scalebox{0.6}{±6.4\textcolor{white}{0}}}  & \textbf{63.9\scalebox{0.6}{±5.4\textcolor{white}{0}}}  & \textbf{62.8\scalebox{0.6}{±9.4\textcolor{white}{0}}}  & \textbf{58.1\scalebox{0.6}{±1.2\textcolor{white}{0}}}  \\
    \bottomrule
    \end{tabular}%
    }
}
\vspace{-0.17cm}
\caption{Accuracy (\% ± std. over 20 runs) of \textbf{5-shot graph classification}. Notations are consistent with Table~\ref{tab:res_main_node}.}
\label{tab:res_main_graph}
\end{table*}

\begin{figure*}[!t]
\captionsetup{skip=2pt}
\centering
\begin{minipage}[t]{0.49\linewidth}
    \centering
    \includegraphics[width=\linewidth]{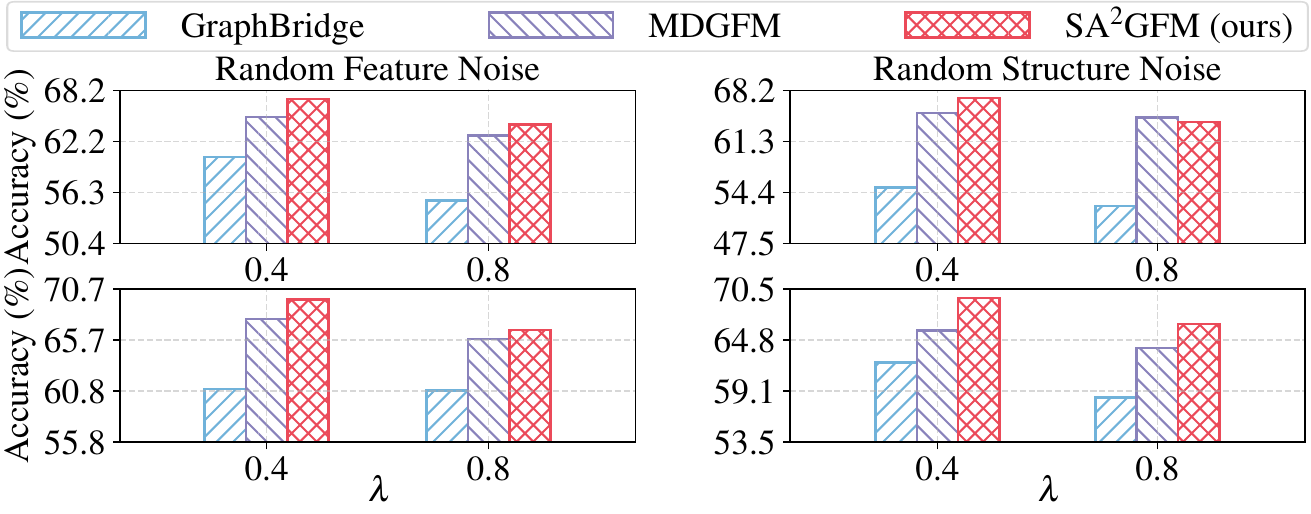}
    \caption{Performance on \texttt{Cora} with varying random noise perturbations ($\lambda$): node (top) and graph (bottom) classification.}
    \label{fig:rq2_a}
\end{minipage}%
\hfill
\begin{minipage}[t]{0.49\linewidth}
    \centering
    \includegraphics[width=\linewidth]{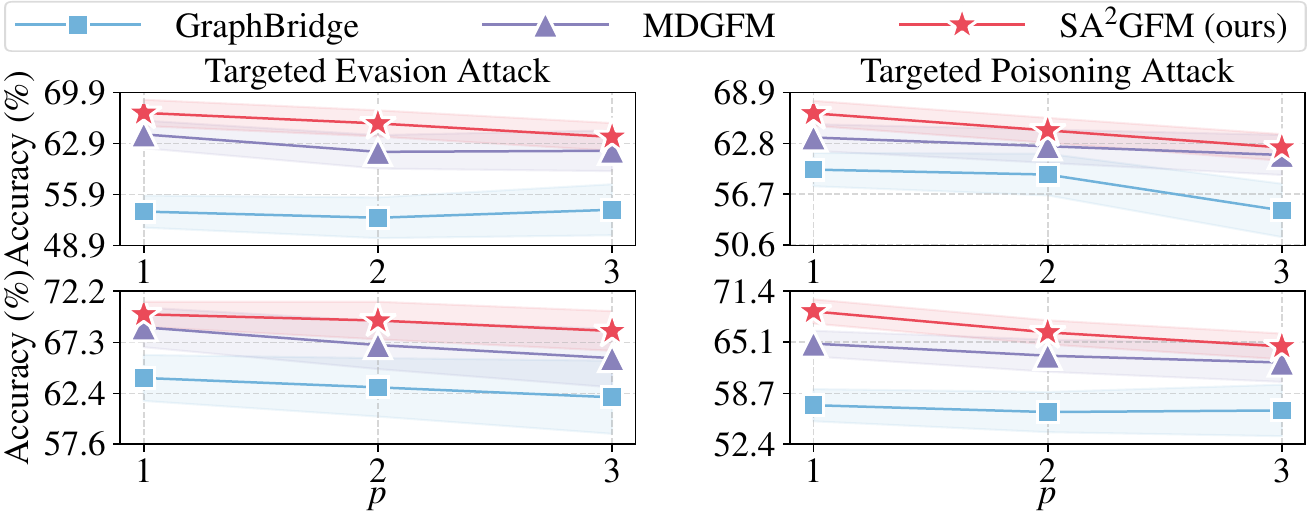}
    \caption{Performance on \texttt{Cora} with varying adversarial attack degree ($p$): node (top) and graph (bottom) classification.}
    \label{fig:rq2_b}
\end{minipage}
\end{figure*}

\subsection{\textit{RQ1:} Against Noise and Adversarial Attacks}
\label{sec:exp_main}
Results (Tables~\ref{tab:res_main_node}, Table~\ref{tab:res_main_graph}) demonstrate that: 
\ding{182} Under both the node and graph classification, \modelname~consistently outperforms all attacks. On average, it achieves +5.9\% (node) and +2.4\% (graph) accuracy improvements over the runner-up, indicating the strong robustness of \modelname~to both non-targeted and target attacks. 
\ding{183} Compared to \texttt{MDGFM}, the strongest baseline with a robust domain adaptation module, \modelname~achieves +5.1\% average gain in the challenging cross-domain settings, confirming its superiority in mitigating negative transfer under the large domain shifts. 
\ding{184} Compared to other baselines without or weaker robustness designs, \modelname~achieves an average improvement of +12.5\%, verifying the impact of our structure-aware pretraining and null-expert routing, which together suppress spurious correlations and prevent negative transfer, contributing to more stable and robust adaptation under various perturbation scenarios.
\ding{185} Additional results in Appendix~\ref{sec:additional_results}. 

\begin{figure*}[!t]
\vspace{-0.15cm}
\captionsetup{skip=2pt}
\centering
\begin{minipage}[t]{0.49\linewidth}
    \centering
    \includegraphics[width=\linewidth]{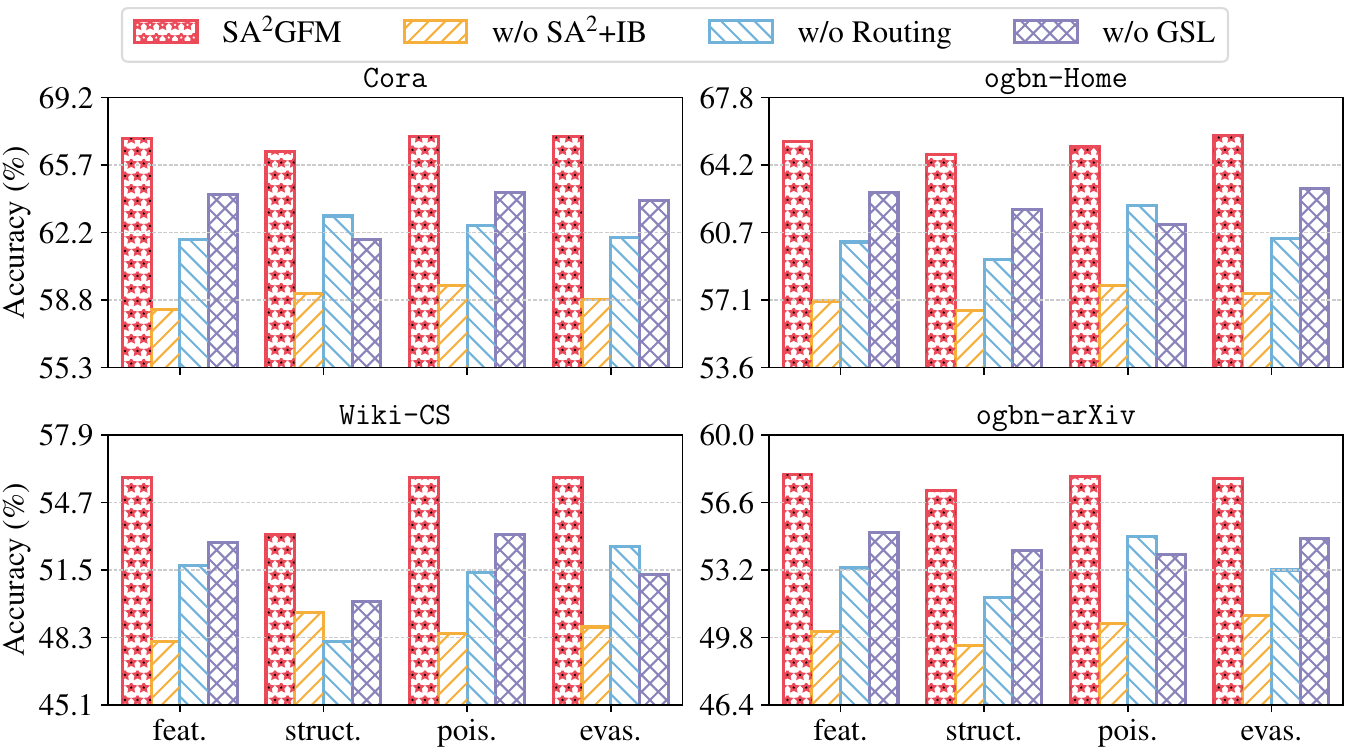}
    \caption{Ablation study on node classification under the non-targeted (random, $\lambda=$~0.4) and targeted ($p = $~1) attacks.}
    \label{fig:rq3}
\end{minipage}%
\hfill
\begin{minipage}[t]{0.49\linewidth}
    \centering
    \includegraphics[width=\linewidth]{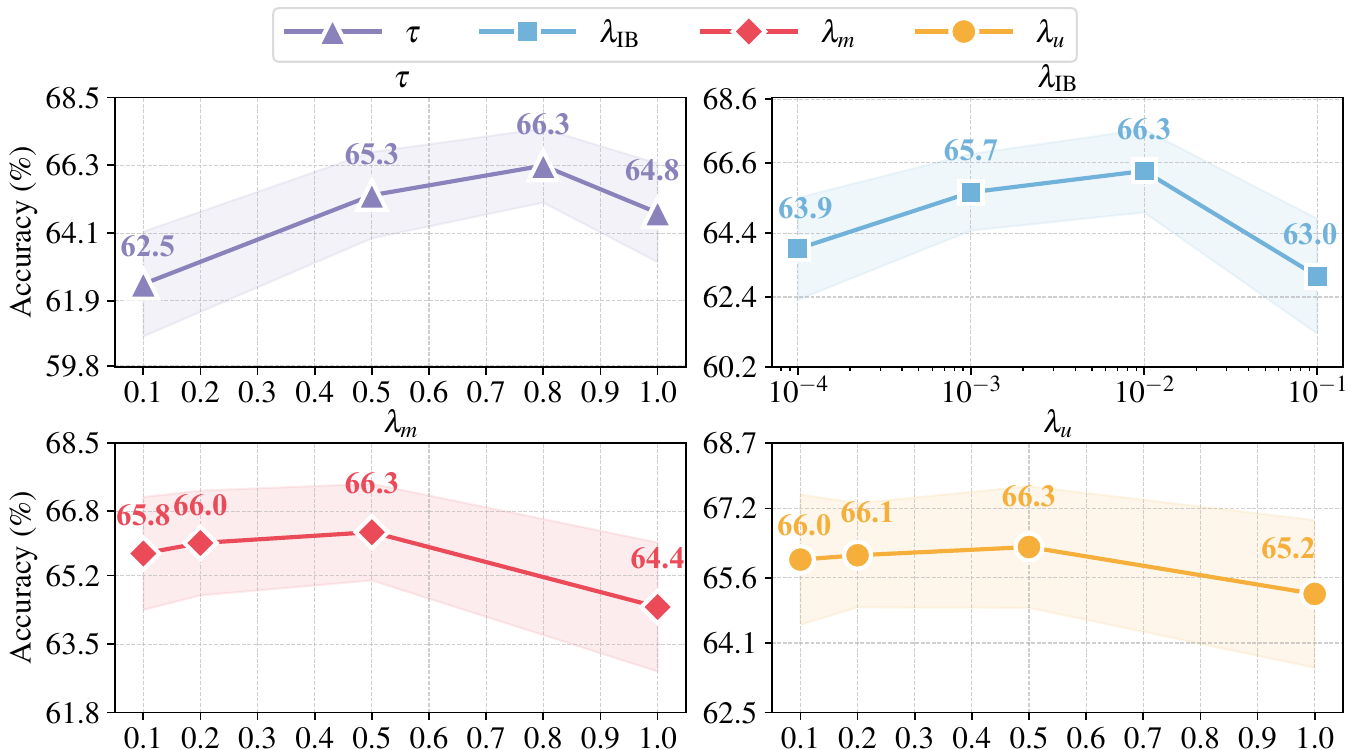}
    \caption{Hyperparameter sensitivity study on node classification under the targeted poisoning attack ($p=$~1).}
    \label{fig:rq4}
\end{minipage}
\vspace{-0.15cm}
\end{figure*}

\subsection{\textit{RQ2:} Analysis of Perturbation Degree}
\label{sec:exp_degree}
We next systematically evaluate the performance degradation of under progressively intensified perturbations.
Results (Figure~\ref{fig:rq2_a} and Figure~\ref{fig:rq2_b}) show that:
\ding{182} Though the performance of all methods degrades as the noise degree or attack severity increases, the drop of \modelname~is significantly slower and more stable throughout.
\ding{183} This observed trend suggests that \modelname~not only achieves stronger base-level robustness but also maintains prediction stability under escalating perturbation levels.
\ding{184} Notably, in the most severe settings (\eg, $\lambda = $~0.8 or $p =$~3), \modelname~consistently outperforms the selected strong baselines MDGFM and GraphBridge by large margins in both node and graph classification, demonstrating its robustness and scalability.

\subsection{\textit{RQ3:} Ablation Study}
\label{sec:exp_component_analysis}
We construct three variants to assess the key components.
\begin{itemize}[leftmargin=*]
    \item \textbf{{\modelname} (\textit{w/o} SA\textsuperscript{2}~+~IB):} removes both the Structure-Aware Semantic-Augmentation and the self-supervised IB module described in Section~\ref{seq:pretrain}.
    \item \textbf{{\modelname} (\textit{w/o} Routing):} removes the selective expert routing mechanism described in Section~\ref{sec:moe}.
    \item \textbf{{\modelname} (\textit{w/o} GSL):} removes the hierarchical structure optimization module described in Section~\ref{sec:finetune}.
\end{itemize}
Results (Figure~\ref{fig:rq3}) reveal all three components contribute to its robustness with varying impact.
\ding{182} Removing ``SA\textsuperscript{2}+IB'' causes the largest drop in accuracy, confirming its key role in learning the transferable, noise-resistant representations.
\ding{183} Removing ``Routing'' degrades performance, especially under targeted attacks.
\ding{184} Removing ``GSL'' leads to reduced accuracy, particularly under structural perturbations.
\ding{185} Overall, the full \modelname~achieves superior robustness through the coordinated contributions of each modules.

\subsection{\textit{RQ4:} Analysis of Hyperparameter Sensitivity}
\label{sec:exp_hyperparam_analysis}
We analyze the sensitivity of \modelname~to four key hyperparameters: $\tau$, $\lambda_{\text{IB}}$, $\lambda_m$, and $\lambda_u$.
Results (Figure~\ref{fig:rq4}) demonstrate that \modelname~is generally stable across a wide range of hyperparameters.
\ding{182} Each of the parameters exhibits a smooth performance curve, with optimal values around the default setting.
\ding{183} Notably, $\lambda_{\text{IB}}$ = 10$^{-\text{2}}$ and $\lambda_m =$~0.5 yield the highest robustness under targeted poisoning attacks, while overly small or large values lead to moderate drops.
\ding{184} The trends suggest that while tuning improves robustness, the model is not overly sensitive to fluctuations, demonstrating reliable performance without extensive parameter searching.

\begin{figure}[!t]
\captionsetup{skip=2pt}
\centering
\begin{minipage}[t]{0.329\linewidth}
    \centering
    \includegraphics[width=\linewidth]{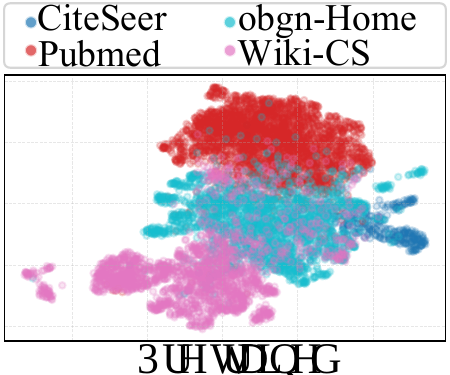}
\end{minipage}
\hspace{-0.12cm}
\hfill
\begin{minipage}[t]{0.329\linewidth}
    \centering
    \includegraphics[width=\linewidth]{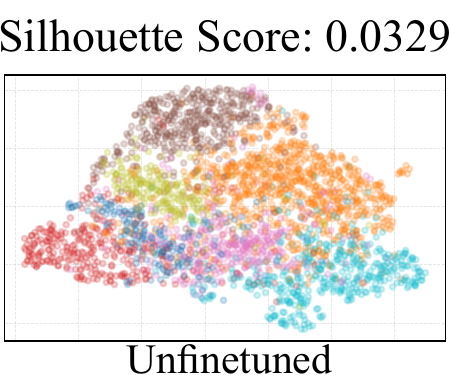}
\end{minipage}
\hspace{-0.12cm}
\hfill
\begin{minipage}[t]{0.329\linewidth}
    \centering
    \includegraphics[width=\linewidth]{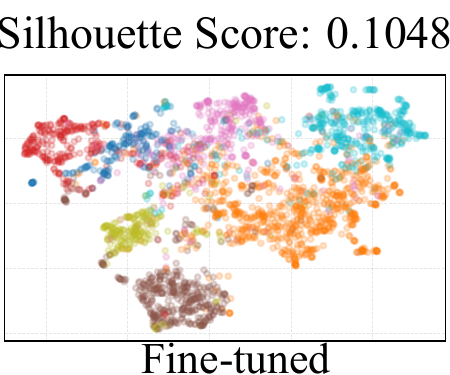}
\end{minipage}
\caption{Node embeddings on \texttt{Cora} (cross-dataset).}
\label{fig:rq5_vis_dataset}
\vspace{-0.3cm}
\end{figure}

\begin{figure}[!t]
\captionsetup{skip=2pt}
\centering
\begin{minipage}[t]{0.329\linewidth}
    \centering
    \includegraphics[width=\linewidth]{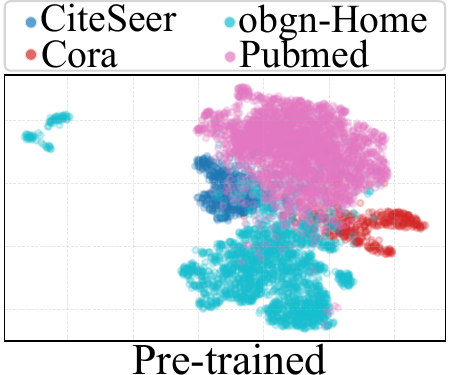}
\end{minipage}
\hspace{-0.12cm}
\hfill
\begin{minipage}[t]{0.329\linewidth}
    \centering
    \includegraphics[width=\linewidth]{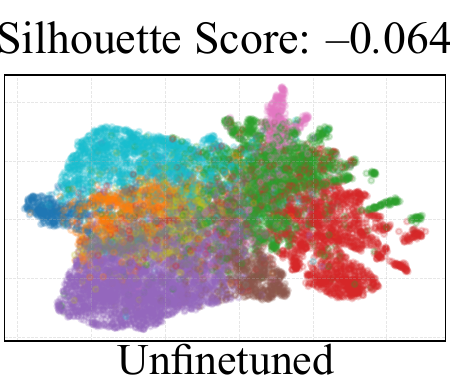}
\end{minipage}
\hspace{-0.12cm}
\hfill
\begin{minipage}[t]{0.329\linewidth}
    \centering
    \includegraphics[width=\linewidth]{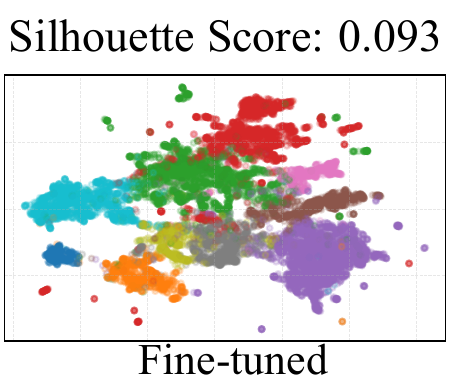}
\end{minipage}
\caption{Node embeddings on \texttt{Wiki-CS} (cross-domain).}
\label{fig:rq5_vis_domain}
\vspace{-0.35cm}
\end{figure}

\subsection{\textit{RQ5:} Embedding Visualizations}
\label{sec:exp_vis}
We utilize the UMAP~\cite{mcinnes2018umap} to visualize the node embeddings and employ the Silhouette Score to evaluate the quality of class-wise clustering (the larger, the better). As shown in Figure~\ref{fig:rq5_vis_dataset} and Figure~\ref{fig:rq5_vis_domain}, the pre-trained embeddings have captured transferable structural and semantic patterns, evidenced by the domain-level separation. However, class-wise separability remains weak without fine-tuning. After fine-tuning, the embeddings become more clustered along class boundaries, with increased Silhouette Scores, confirming effective task-specific alignment.

\section{Conclusion}
\label{sec:conclusion}
In this work, we propose a robust graph foundation model \modelname. Through unified pre-training and fine-tuning, our method effectively bridges upstream transferable knowledge with downstream structural adaptation. Extensive experiments demonstrate that \modelname~achieves superior robustness against both the random noise and adversarial perturbations, as well as strong generalization across domains, and consistently yields gains from its key modules, offering a promising foundation for reliable and adaptable GFM.

\clearpage
\newpage
\section*{Acknowledgments}
The corresponding author is Qingyun Sun. This work is supported in part by the National Natural Science Foundation of China (NSFC) under grants No. 62427808 and No. 623B2010, by the Fundamental Research Funds for the Central Universities, by the Basic Ability Enhancement Program for Young and Middle-aged Teachers of Guangxi (No. 2024KY0073), and by the Academic Excellence Foundation of BUAA for PhD Students. We extend our sincere thanks to all authors for their valuable contributions.
\bibliography{main}

@inproceedings{
  kipf2017semisupervised,
  title={Semi-supervised classification with graph convolutional networks},
  author={Thomas N. Kipf and Max Welling},
  booktitle={ICLR},
  year={2017},
}

@inproceedings{
  veličković2018graph,
  title={Graph attention networks},
  author={Petar Veličković and Guillem Cucurull and Arantxa Casanova and Adriana Romero and Pietro Liò and Yoshua Bengio},
  booktitle={ICLR},
  year={2018}
}

@inproceedings{
  Hu2020Strategies,
  title={Strategies for pre-training graph neural networks},
  author={Weihua Hu and Bowen Liu and Joseph Gomes and Marinka Zitnik and Percy Liang and Vijay Pande and Jure Leskovec},
  booktitle={ICLR},
  year={2020}
}

@article{yu2024text,
  title={Text-free multi-domain graph pre-training: Toward graph foundation models},
  author={Yu, Xingtong and Zhou, Chang and Fang, Yuan and Zhang, Xinming},
  journal={arXiv preprint arXiv:2405.13934},
  year={2024}
}

@inproceedings{
  wang2025multidomain,
  title={Multi-domain graph foundation models: Robust knowledge transfer via topology alignment},
  author={Shuo Wang and Bokui Wang and Zhixiang Shen and Boyan Deng and Zhao Kang},
  booktitle={ICML},
  year={2025}
}

@inproceedings{
  yuan2025how,
  title={How much can transfer? {BRIDGE}: Bounded multi-domain graph foundation model with generalization guarantees},
  author={Haonan Yuan and Qingyun Sun and Junhua Shi and Xingcheng Fu and Bryan Hooi and Jianxin Li and Philip S. Yu},
  booktitle={ICML},
  year={2025}
}

@article{DBLP:journals/corr/abs-2310-11829,
  title={Towards graph foundation models: A survey and beyond},
  author={Liu, Jiawei and Yang, Cheng and Lu, Zhiyuan and Chen, Junze and Li, Yibo and Zhang, Mengmei and Bai, Ting and Fang, Yuan and Sun, Lichao and Yu, Philip S and others},
  journal={CoRR},
  year={2023}
}

@inproceedings{
  xu2018how,
  title={How powerful are graph neural networks?},
  author={Keyulu Xu and Weihua Hu and Jure Leskovec and Stefanie Jegelka},
  booktitle={ICLR},
  year={2019}
}

@inproceedings{
  morris2019weisfeiler,
  title={Weisfeiler and leman go neural: Higher-order graph neural networks},
  author={Christopher Morris and Martin Ritzert and Matthias Fey and William L. Hamilton and Jan Eric Lenssen and Gaurav Rattan and Martin Grohe},
  booktitle={AAAI},
  volume={33},
  number={01},
  pages={4602--4609},
  year={2019}
}

@inproceedings{
  wang2024dissecting,
  title={Dissecting the failure of invariant learning on graphs},
  author={Qixun Wang and Yifei Wang and Yisen Wang and Xianghua Ying},
  booktitle={NeurIPS},
  year={2024}
}

@inproceedings{
  tishby1999information,
  author={Naftali Tishby and Fernando C. Pereira and William Bialek},
  title={The information bottleneck method},
  booktitle={Proceedings of the 37th Annual Allerton Conference on Communication, Control and Computing},
  year={1999},
  pages={368--377}
}

@inproceedings{
  lee2025selfsupervised,
  title={Self-supervised adversarial purification for graph neural networks},
  author={Woohyun Lee and Hogun Park},
  booktitle={ICML},
  year={2025}
}

@inproceedings{
  zugner2018adversarial,
  author={Z{\"u}gner, Daniel and Akbarnejad, Amir and G{\"u}nnemann, Stephan},
  title={Adversarial attacks on neural networks for graph data},
  booktitle={KDD},
  pages={2847--2856},
  year={2018}
}

@inproceedings{
  zhang2024gder,
  title={{GDeR}: Safeguarding efficiency, balancing, and Robustness via Prototypical Graph Pruning},
  author={Guibin Zhang and Haonan Dong and Yuchen Zhang and Zhixun Li and Dingshuo Chen and Kai Wang and Tianlong Chen and Yuxuan Liang and Dawei Cheng and Kun Wang},
  booktitle={NeurIPS},
  year={2024}
}

@inproceedings{
  kataria2024ugc,
  author={Kataria, Mohit and Kumar, Sandeep and Jayadeva},
  title={{UGC}: Universal graph coarsening},
  booktitle={NeurIPS},
  pages={63057--63081},
  year={2024}
}

@inproceedings{
  ju2025graphbridge,
  title={{GraphBridge}: Towards arbitrary transfer learning in {GNN}s},
  author={Li Ju and Xingyi Yang and Qi Li and Xinchao Wang},
  booktitle={ICLR},
  year={2025}
}

@inproceedings{
  zheng2024intramix,
  author={Zheng, Shenghe and Wang, Hongzhi and Liu, Xianglong},
  title={{IntraMix}: Intra-class mixup generation for accurate labels and neighbors},
  booktitle={NeurIPS},
  pages={8951--8980},
  year={2024}
}

@inproceedings{
  huang2025learn,
  title={Learn beneficial noise as graph augmentation},
  author={Siqi Huang and Yanchen Xu and Hongyuan Zhang and Xuelong Li},
  booktitle={ICML},
  year={2025}
}

@inproceedings{
  li2025disentangle,
  author={Li, Haoyang and Wang, Xin and Zhu, Xueling and Wen, Weigao and Zhu, Wenwu},
  title={Disentangling invariant subgraph via variance contrastive estimation under distribution shifts},
  booktitle={ICML},
  year={2025},
  note={Poster}
}

@inproceedings{
  long2020graphstone,
  author={Long, Qingqing and Jin, Yilun and Song, Guojie and Li, Yi and Lin, Wei},
  title={Graph structural-topic neural network},
  booktitle={KDD},
  pages={1065--1073},
  year={2020}
}

@inproceedings{
 bai2024hcgae,
  author={Bai, Lu and Xu, Zhuo and Cui, Lixin and Li, Ming and Wang, Yue and Hancock, Edwin R.},
  title={{HC-GAE}: The hierarchical cluster-based graph auto-encoder for graph representation learning},
  booktitle={NeurIPS},
  year={2024}
}

@inproceedings{
  NEURIPS2024_59e73ff8,
   author = {Dan, Jun and Liu, Weiming and Xie, Chunfeng and Yu, Hua and Dong, Shunjie and Tan, Yanchao},
   booktitle = {NeurIPS},
   pages = {50230--50255},
   title = {{TFGDA}: Exploring topology and feature alignment in semi-supervised graph domain adaptation through robust clustering},
   volume = {37},
   year = {2024}
}

@inproceedings{
  gasteiger2018combining,
  title={Combining neural networks with personalized pagerank for classification on graphs},
  author={Johannes Gasteiger and Aleksandar Bojchevski and Stephan Günnemann},
  booktitle={ICLR},
  year={2019}
}

@inproceedings{
  alemi2017deep,
  title={Deep variational information bottleneck},
  author={Alexander A. Alemi and Ian Fischer and Joshua V. Dillon and Kevin Murphy},
  booktitle={ICLR},
  year={2017}
}

@article{zhu2021deep,
  title={Deep graph structure learning for robust representations: A survey},
  author={Zhu, Yanqiao and Xu, Weizhi and Zhang, Jinghao and Liu, Qiang and Wu, Shu and Wang, Liang},
  journal={arXiv preprint arXiv:2103.03036},
  volume={14},
  pages={1--1},
  year={2021}
}

@article{stewart1993early,
  title={On the early history of the singular value decomposition},
  author={Stewart, Gilbert W},
  journal={SIAM Review},
  volume={35},
  number={4},
  pages={551--566},
  year={1993}
}

@inproceedings{poole2019variational,
  title={On variational bounds of mutual information},
  author={Poole, Ben and Ozair, Sherjil and Van Den Oord, Aaron and Alemi, Alex and Tucker, George},
  booktitle={ICML},
  pages={5171--5180},
  year={2019},
  organization={PMLR}
}

@article{oord2018representation,
  title={Representation learning with contrastive predictive coding},
  author={Oord, Aaron van den and Li, Yazhe and Vinyals, Oriol},
  journal={arXiv preprint arXiv:1807.03748},
  year={2018}
}

@article{mccallum2000automating,
  title={Automating the construction of internet portals with machine learning},
  author={McCallum, Andrew Kachites and Nigam, Kamal and Rennie, Jason and Seymore, Kristie},
  journal={Information Retrieval},
  volume={3},
  pages={127--163},
  year={2000}
}

@inproceedings{giles1998citeseer,
  title={{CiteSeer}: An automatic citation indexing system},
  author={Giles, C Lee and Bollacker, Kurt D and Lawrence, Steve},
  booktitle={Proceedings of the Third ACM Conference on Digital libraries},
  pages={89--98},
  year={1998}
}

@article{sen2008collective,
  title={Collective classification in network data},
  author={Sen, Prithviraj and Namata, Galileo and Bilgic, Mustafa and Getoor, Lise and Galligher, Brian and Eliassi-Rad, Tina},
  journal={AI magazine},
  volume={29},
  number={3},
  pages={93--93},
  year={2008}
}

@inproceedings{hu2020open,
  title={Open graph benchmark: Datasets for machine learning on graphs},
  author={Hu, Weihua and Fey, Matthias and Zitnik, Marinka and Dong, Yuxiao and Ren, Hongyu and Liu, Bowen and Catasta, Michele and Leskovec, Jure},
  booktitle={NeurIPS},
  volume={33},
  pages={22118--22133},
  year={2020}
}

@article{mernyei2020wikics,
  title={{Wiki-CS}: A Wikipedia-based benchmark for graph neural networks},
  author={Mernyei, P{\'e}ter and Cangea, C{\u{a}}t{\u{a}}lina},
  journal={arXiv preprint arXiv:2007.02901},
  year={2020}
}

@inproceedings{liu2023graphprompt,
  author={Zemin Liu and Xingtong Yu and Yuan Fang and Xinming Zhang},
  title={{GraphPrompt}: Unifying pre-training and downstream tasks for graph neural networks},
  booktitle={WWW},
  pages={417--428},
  year={2023}
}

@inproceedings{you2020graphcl,
  author={Yuning You and Tianlong Chen and Yongduo Sui and Ting Chen and Zhangyang Wang and Yang Shen},
  title={Graph contrastive learning with augmentations},
  booktitle={NeurIPS},
  articleno={488},
  pages={1--12},
  year={2020}
}

@inproceedings{velickovic2019deep,
  author={Petar Veličković and William Fedus and William L. Hamilton and Pietro Liò and Yoshua Bengio and R. Devon Hjelm},
  title={Deep graph infomax},
  booktitle={ICLR},
  year={2019}
}

@article{li2016structural,
  title={Structural Information and Dynamical Complexity of Networks},
  author={Li, Angsheng and Pan, Yicheng},
  journal={IEEE Transactions on Information Theory},
  volume={62},
  number={6},
  year={2016}
}

@inproceedings{zhao2024all,
  title={All in one and one for all: A simple yet effective method towards cross-domain graph pretraining},
  author={Zhao, Haihong and Chen, Aochuan and Sun, Xiangguo and Cheng, Hong and Li, Jia},
  booktitle={KDD},
  pages={4443--4454},
  year={2024}
}

@article{mcinnes2018umap,
  title={{UMAP}: Uniform manifold approximation and projection},
  author={McInnes, Leland and Healy, John and Saul, Nathaniel and Gro{\ss}berger, Lukas},
  journal={Journal of Open Source Software},
  volume={3},
  number={29},
  pages={861},
  year={2018}
}

@inproceedings{tang2024higpt,
  title={HiGPT: Heterogeneous Graph Language Model},
  author={Tang, Jiabin and Yang, Yuhao and Wei, Wei and Shi, Lei and Xia, Long and Yin, Dawei and Huang, Chao},
  booktitle={KDD},
  pages={2842--2853},
  year={2024}
}

@article{yuan2023environment,
  title={Environment-aware dynamic graph learning for out-of-distribution generalization},
  author={Yuan, Haonan and Sun, Qingyun and Fu, Xingcheng and Zhang, Ziwei and Ji, Cheng and Peng, Hao and Li, Jianxin},
  journal={NeurIPS},
  volume={36},
  year={2023}
}

@inproceedings{yuan2024dynamic,
  title={Dynamic Graph Information Bottleneck},
  author={Yuan, Haonan and Sun, Qingyun and Fu, Xingcheng and Ji, Cheng and Li, Jianxin},
  booktitle={WWW},
  pages={469--480},
  year={2024}
}

@inproceedings{yuan2025dg,
  title={{DG-Mamba}: Robust and efficient dynamic graph structure learning with selective state space models},
  author={Yuan, Haonan and Sun, Qingyun and Wang, Zhaonan and Fu, Xingcheng and Ji, Cheng and Wang, Yongjian and Jin, Bo and Li, Jianxin},
  booktitle={AAAI},
  volume={39},
  number={21},
  pages={22272--22280},
  year={2025}
}

@inproceedings{yuangraver,
  title={{GRAVER}: Generative Graph Vocabularies for Robust Graph Foundation Models Fine-tuning},
  author={Haonan Yuan and Qingyun Sun and Junhua Shi and Xingcheng Fu and Bryan Hooi and Jianxin Li and Philip S. Yu},
  booktitle={NeurIPS},
  year={2025}
}

\clearpage
\newpage
\appendix
\counterwithin{table}{section}
\counterwithin{figure}{section}
\counterwithin{equation}{section}

\section{Algorithms and Complexity Analysis}
\label{sec:alg}
\modelname~includes two stages: pre-training and fine-tuning. We illustrate the pipeline for each stage in Algorithm~\ref{alg:alg_pretrain} and Algorithm~\ref{alg:alg_finetune}. The complexity is analyzed as follows.

\begin{algorithm}[H]
    \caption{The pre-training pipeline of~\modelname.}
    \label{alg:alg_pretrain}
    \textbf{Input:} Source graphs $\{G_i^\mathcal{S} = (\mathbf{A}_i^\mathcal{S}, \mathbf{X}_i^\mathcal{S})\}_{i=1}^{n}$; Pre-training epochs $E_p$; Learning rate $\eta_p$; Hyperparameter $\lambda_{\mathrm{IB}}$. \\
    \textbf{Output:} Pre-trained encoders $\{\mathcal{F}_{\boldsymbol{\Theta}_i}^{\star}\}_{i=1}^{n}$.
    
    \begin{algorithmic}[1]
        \FOR{each $G_i^\mathcal{S}$ in $\{G_i^\mathcal{S}\}_{i=1}^{n}$}
            \STATE Partition $G_i^\mathcal{S}$ into $K$ clusters using structural entropy;
            \FOR{each node $v_j$ in $G_i^\mathcal{S}$}
                \STATE Generate text $\mathbf{t}_j$ for $v_j$ based on cluster context;
                \STATE Compute augmented feature: $\mathbf{x}_j^{\mathcal{S}\prime} \gets$ Eq.~\eqref{eq:1};
            \ENDFOR
            \STATE Obtain the augmented feature matrix $\mathbf{X}_i^{\mathcal{S}\prime}$;
            \STATE Initialize encoder parameters $\boldsymbol{\Theta}_i$;
            \FOR{epoch $e = 1, \cdots, E_p$}
                \STATE Sample a batch of positive and negative pairs;
                \STATE $(\boldsymbol{\mu}_i, \boldsymbol{\sigma}_i) \gets \mathcal{F}_{\boldsymbol{\Theta}_i}(\mathbf{A}_i^\mathcal{S}, \mathbf{X}_i^{\mathcal{S}\prime})$;
                \STATE $\mathbf{Z}_i^\mathcal{S} \gets \boldsymbol{\mu}_i + \boldsymbol{\epsilon} \odot \boldsymbol{\sigma}_i,\ \boldsymbol{\epsilon} \sim \mathcal{N}(0, \mathbf{I})$;
                \STATE Compute contrastive loss: $\mathcal{L}_{\text{InfoNCE}} \gets$ Eq.~\eqref{eq:deriv_1};
                \STATE Compute compression loss: $\mathcal{L}_{\text{KL}} \gets$ Eq.~\eqref{eq:deriv_2};
                \STATE Compute total loss: $\mathcal{L}_{\text{pretrain}} \gets$ Eq.~\eqref{eq:7};
                \STATE Update $\boldsymbol{\Theta}_i$ by back-propagation.
            \ENDFOR
        \ENDFOR
    \end{algorithmic}
\end{algorithm}

\subsection{Computational Complexity Analysis of the Pre-training Stage (Algorithm~\ref{alg:alg_pretrain})}
\label{sec:appendix_complexity_pretrain}

This section aims to provide a detailed breakdown of the computational complexity of the \modelname~during the multi-domain pre-training stage, as outlined in Algorithm~\ref{alg:alg_pretrain}. The overall cost of the pre-training stage is primarily composed of two parts: a one-time Structure-Aware Semantic Augmentation for each source graph, followed by the iterative Self-Supervised Information Bottleneck training.

We analyze the computational cost for a single iteration of the outer loop in the algorithm (Line 1), which processes a single source graph $G_i^\mathcal{S} = (\mathbf{A}_i^\mathcal{S}, \mathbf{X}_i^\mathcal{S})$. This graph contains $|\mathcal{V}_i|$ nodes and $|\mathcal{E}_i|$ edges.

\subsubsection{Structure-Aware Semantic Augmentation (Lines 2-7).}
This step generates the augmented node feature matrix $\mathbf{X}_i^{\mathcal{S}\prime}$ for each source graph and is a pre-processing procedure.
\begin{itemize}[leftmargin=*]
    \item \textbf{Structural Entropy Partitioning (Line 2):} Computing the optimal hierarchical partition (encoding tree) for the graph $G_i^\mathcal{S}$ has a complexity of $\mathcal{O}(|\mathcal{E}_i| \log |\mathcal{V}_i|)$.

    \item \textbf{Feature Augmentation (Lines 3-6):} The algorithm generates a text prompt $\mathbf{t}_j$ and computes the augmented feature $\mathbf{x}_j^{\mathcal{S}\prime}$ for each node $v_j$ in the graph. This process involves using BERT and SVD. The total complexity for performing this operation on all $|\mathcal{V}_i|$ nodes is approximately $\mathcal{O}(|\mathcal{V}_i| \cdot (C_{\text{BERT}} + (d_i + d_{\text{BERT}}) \cdot d_0))$, where $C_{\text{BERT}}$ represents the BERT inference cost, $d_i$ and $d_{\text{BERT}}$ are the dimensions of the original features and BERT embeddings respectively, and $d_0$ is the aligned dimension.
\end{itemize}
The cost of this step is amortized over the entire training process and is typically much smaller than the subsequent iterative training cost.

\subsubsection{Self-Supervised Information Bottleneck (SS-IB) Training (Lines 8-16).}
SS-IB is the computational core of the pre-training, where it trains an expert encoder $\mathcal{F}_{\boldsymbol{\Theta}_i}$ independently for each source graph $G_i^\mathcal{S}$ over $E_p$ epochs.
\begin{itemize}[leftmargin=*]
    \item \textbf{Encoder Forward Pass (Line 11):} In each epoch, the main computation is the forward pass of the GNN encoder $\mathcal{F}_{\boldsymbol{\Theta}_i}$. It takes the adjacency matrix $\mathbf{A}_i^\mathcal{S}$ and the augmented feature matrix $\mathbf{X}_i^{\mathcal{S}\prime}$ as input. For an $L$-layer GNN, a single forward pass on a sparse graph has a complexity of $\mathcal{O}(L \cdot |\mathcal{E}_i| \cdot d)$, where $d$ is the hidden feature dimension.

    \item \textbf{Loss Computation and Backpropagation (Lines 13-16):} The complexity of computing $\mathcal{L}_{\text{InfoNCE}}$ and $\mathcal{L}_{\text{KL}}$ is linear with respect to the number of nodes $|\mathcal{V}_i|$, i.e., $\mathcal{O}(|\mathcal{V}_i| \cdot d)$. The update of parameters $\boldsymbol{\Theta}_i$ depends on the backpropagation through the entire computation graph, which has a complexity comparable to the forward pass. Therefore, the main cost of a single iteration is dominated by the GNN computation.

    \item \textbf{Total Training Cost:} Repeating this process for $E_p$ epochs, the total complexity for training a single expert encoder $\mathcal{F}_{\boldsymbol{\Theta}_i}$ is $\mathcal{O}(E_p \cdot L \cdot |\mathcal{E}_i| \cdot d)$.
\end{itemize}

\subsubsection{Overall Pre-training Complexity.}
The outer loop of Algorithm~\ref{alg:alg_pretrain} (Line 1) indicates that the above process is performed independently for all $n$ source domains. Therefore, the total computational complexity of pre-training is the sum of the training costs for all $n$ experts. The final complexity is determined by the number of pre-training epochs ($E_p$), the number of layers in the GNN encoder ($L$), and its hidden feature dimension ($d$). Assuming an average number of edges $|\mathcal{E}|$ across all graphs, the overall complexity is:
\begin{equation}
    \mathcal{O}(n \cdot E_p \cdot L \cdot |\mathcal{E}| \cdot d)
\end{equation}
The analysis shows the pre-training complexity scales linearly with the number of source domains and the number of edges in the graphs, enabling it to efficiently scale to multi-domain and large-scale graph scenarios.

\subsection{Computational Complexity Analysis of the Fine-tuning Stage (Algorithm~\ref{alg:alg_finetune})}
\label{sec:appendix_complexity_finetune}
\begin{algorithm}[!t]
    \caption{The fine-tuning pipeline of~\modelname.}
    \label{alg:alg_finetune}
    \textbf{Input:} 5 nodes class-balanced sampled from a target graphs $\{G^\mathcal{T}\}\in\mathcal{G}^\mathcal{T}$ (node classification) or 5 target graphs $\{G^\mathcal{T}\}\in\mathcal{G}^\mathcal{T}$ from target domain $\{D^\mathcal{T}\}\in\mathcal{D}^\mathcal{T}$ (graph classification); Pre-trained experts $\{\mathcal{F}_{\boldsymbol{\Theta}_i}^{\star}\}_{i=1}^{n}$; Epochs $E_f$; Expert routing network $\mathcal{R}_{\boldsymbol{\phi}}$; Learnable prompt $\mathcal{P}_{\boldsymbol{\Omega}}$; Learning rate $\eta_f$; Hyperparameters $\lambda_{m}, \lambda_{u}$.
    
    \textbf{Output:} Fine-tuned \modelname~with $\{\boldsymbol{\Theta}^\star, \boldsymbol{\phi}^\star, \boldsymbol{\Omega}^\star\}$.
    
    \begin{algorithmic}[1]
        \STATE Initialize the routing network: $\boldsymbol{\alpha} \gets$ Eq.~\eqref{eq:8} and Eq.~\eqref{eq:9}; and randomly initialize other parameters $\boldsymbol{\Omega}$;
        \STATE Partition $G^{\mathcal{T}}$ into clusters $\mathcal{C}^\mathcal{T}$ using structural entropy;
        \FOR{each node $v_i$ in $G^{\mathcal{T}}$}
            \STATE Compute augmented feature: $\mathbf{x}_i^{\mathcal{T}\prime} \gets$ Eq.~\eqref{eq:1};
        \ENDFOR
        \STATE Obtain augmented feature matrix $\mathbf{X}^{\mathcal{T}\prime}$;
        \FOR{$e=$ 1, $\cdots$, $E_f$}
        \STATE Compute MoE regularization loss: $\mathcal{L}_{\text{MoE}} \gets$ Eq.~\eqref{eq:12};
        \STATE Compute uncertainty loss: $\mathcal{L}_{\text{uncertain}} \gets$ Eq.~\eqref{eq:14};
        \STATE Refine target graph structure: $\mathbf{A}^{\mathcal{T}'} \gets$ Eq.~\eqref{eq:17};
        \STATE Generate final embeddings: $\mathbf{Z}^\mathcal{T} \gets$ Eq.~\eqref{eq:get_embeddings};
        \STATE Compute classification loss: $\mathcal{L}_{\text{cls}} \gets$ Eq.~\eqref{eq:cls_loss};
        \STATE Compute fine-tuning loss: $\mathcal{L}_{\text{finetune}} \gets$ Eq.~\eqref{eq:finetune_loss};
        \STATE Update $\{\boldsymbol{\phi}, \boldsymbol{\Omega}\}$ by back-propagation.
        \ENDFOR
    \end{algorithmic}
\end{algorithm}

The fine-tuning stage, outlined in Algorithm~\ref{alg:alg_finetune}, adapts the pre-trained model to a target graph $G^\mathcal{T} = (\mathbf{A}^\mathcal{T}, \mathbf{X}^\mathcal{T})$ with $|\mathcal{V_T}|$ nodes and $|\mathcal{E_T}|$ edges, where the number of fine-tuning epochs, $E_f$, is small. The complexity analysis is broken down into a one-time preprocessing step and the per-epoch iterative training cost.

\subsubsection{Preprocessing (Lines 1-6).}
Before the main training loop, the model performs a series of one-time setup operations.
\begin{itemize}[leftmargin=*]
    \item \textbf{Parameter Initialization (Line 1):} This involves initializing the learnable parameters, including the routing network $\mathcal{R}_{\boldsymbol{\phi}}$ and the prompts $\mathcal{P}_{\boldsymbol{\Omega}}$. The calculation of initial routing weights $\boldsymbol{\alpha}$ requires computing prototypes from the small, few-shot support set, making its cost negligible.

    \item \textbf{Structure-Aware Semantic Augmentation (Lines 2-6):} The calculation process for this step is identical to that of the pre-training stage and is performed only once. It consists of two main parts: Structural Entropy Partitioning (Line 2) with a complexity of $\mathcal{O}(|\mathcal{E_T}| \log |\mathcal{V_T}|)$, and Feature Augmentation (Lines 3-6) with a complexity of $\mathcal{O}(|\mathcal{V_T}| \cdot (C_{\text{BERT}} + (d_i + d_{\text{BERT}}) \cdot d_0))$.
\end{itemize}

\subsubsection{Per-Epoch Training Complexity (Lines 7-14).}
The main computational load resides within the fine-tuning loop, which runs for $E_f$ epochs.
\begin{itemize}[leftmargin=*]
    \item \textbf{Loss Computations for Regularization (Lines 8-9):} The calculation of the MoE loss $\mathcal{L}_{\text{MoE}}$ and the uncertainty loss $\mathcal{L}_{\text{uncertain}}$ are computationally inexpensive. $\mathcal{L}_{\text{MoE}}$ is an entropy calculation over $n+1$ weights, costing $\mathcal{O}(n)$. The cost of $\mathcal{L}_{\text{uncertain}}$ is tied to the attention score computation within the structure optimization step.

    \item \textbf{Hierarchical Structure Optimization (Line 10):} This is a significant part of the per-epoch computation. It refines the graph topology $\mathbf{A}^\mathcal{T}$ into $\mathbf{A}^{\mathcal{T}'}$ by combining two components. First, \textit{intra-cluster learning} uses multi-head attention within each cluster $c$, which has a complexity quadratic to cluster size, totaling $\mathcal{O}(\sum_{c \in \mathcal{C}^\mathcal{T}} |\mathcal{V}_c|^2 \cdot d_k)$, where $|\mathcal{V}_c|$ is the number of nodes in cluster $c$. Secondly, \textit{inter-cluster learning} uses an APPNP-like propagation for $T$ steps, with a complexity of $\mathcal{O}(T \cdot |\mathcal{E_T}|)$.

    \item \textbf{Final Embedding Generation and Classification (Lines 11-12):} To compute the final classification loss $\mathcal{L}_{\text{cls}}$, the model first generates node embeddings $\mathbf{Z}^\mathcal{T}$ using the frozen pre-trained encoders $\{\mathcal{F}_{\boldsymbol{\Theta}_i}^{\star}\}_{i=1}^{n}$. This involves a full forward pass on the newly refined graph structure $\mathbf{A}^{\mathcal{T}'}$. The complexity of this step is $\mathcal{O}(n \cdot L \cdot |\mathcal{E_T}'| \cdot d)$, where $|\mathcal{E_T}'|$ is the number of edges in the optimized graph, which we can approximate as $\mathcal{O}(n \cdot L \cdot |\mathcal{E_T}| \cdot d)$.
\end{itemize}

\subsubsection{Overall Fine-tuning Complexity.}
The one-time preprocessing costs are amortized and considered low-order terms. The dominant cost is within the training loop. Summing the complexities of the most expensive steps per epoch, the overall complexity for the fine-tuning stage is:
\begin{equation}
    \mathcal{O}\left(E_f \cdot \left( \sum_{c}|\mathcal{V}_c|^2 d_k + T|\mathcal{E_T}| + nL|\mathcal{E_T}|d \right) \right)
\end{equation}
This formulation highlights that the per-epoch cost is independent of the number of experts $n$. The primary computational burden comes from the intra-cluster attention, making the process highly efficient for graphs that can be partitioned into reasonably small, balanced clusters. This efficiency is critical for the model's applicability in few-shot learning contexts.

\section{Proofs}
\label{sec:proof}
This section proves the approximation error bound of our practical objective with respect to the theoretical Self-Supervised Information Bottleneck (SS-IB) objective.

\subsection{Proof of Derivation~\ref{deriv:deriv_1}}
\label{sec:proof_1}
We first restate Derivation~\ref{deriv:deriv_1} for reference.
\begin{mathbox}
    \textbf{Derivation 1} (Lower Bound of $I(\mathbf{Z}^{\mathcal{S}}; \mathbf{Z}^{\mathcal{S}+})$)\textbf{.}
    Prediction term $I(\mathbf{Z}^{\mathcal{S}}; \mathbf{Z}^{\mathcal{S}+})$ is lower-bounded by InfoNCE~\cite{oord2018representation}. For each anchor $v_i$ and positive node $v_{i}^+$:
    \begin{align}
        \!\!\!\!\!\!I(\mathbf{Z}^{\mathcal{S}}; \mathbf{Z}^{\mathcal{S}+})&\geqslant  -\mathcal{L}_{\text{InfoNCE}}\notag\\
        &= \frac{1}{N^+} \sum\nolimits_{i=1}^{N^+}\log \frac{\exp(\langle\mathbf{Z}^{\mathcal{S}}_i, \mathbf{Z}^{\mathcal{S}+}_{i}\rangle / \tau)}{\sum_{j=0}^{N^-} \exp(\langle\mathbf{Z}^{\mathcal{S}}_i, \mathbf{Z}^{\mathcal{S}}_j\rangle / \tau)},\!\!\!\!
    \notag
    \end{align}
    where $N^+$ and $N^-$ are the number of positive and negative samples, which are non-redundantly and consistently sampled from the source domain depending on whether there is a direct link~\cite{liu2023graphprompt}. $\tau$ is a temperature.
\end{mathbox}

\begin{proof}
    We consider the standard mutual information:
    \begin{equation}
        I(\mathbf{Z}^{\mathcal{S}}; \mathbf{Z}^{\mathcal{S}+}) = \mathbb{E}_{p(\mathbf{Z}^{\mathcal{S}}, \mathbf{Z}^{\mathcal{S}+})} \left[ \log \frac{p(\mathbf{Z}^{\mathcal{S}}, \mathbf{Z}^{\mathcal{S}+})}{p(\mathbf{Z}^{\mathcal{S}})p(\mathbf{Z}^{\mathcal{S}+})} \right].
    \end{equation}
    However, $p(\mathbf{Z}^{\mathcal{S}})$ and $p(\mathbf{Z}^{\mathcal{S}+})$ is intractable. Following~\citet{oord2018representation, poole2019variational}, we approximate by a classification-based surrogate using Noise-Contrastive Estimation (NCE), where it distinguishes $N^+$ positive sample $\mathbf{Z}^{\mathcal{S}+}_{i}$ from $N^-$ negative samples $\{\mathbf{Z}^{\mathcal{S}}_j\}_{j=0}^{N^-}$.
    Let $q(i | \mathbf{Z}^{\mathcal{S}})$ be the probability of correctly identifying the positive sample among a set of ($N^- + 1$) candidates, defined as:
    \begin{equation}
        q(i^+|\mathbf{Z}^{\mathcal{S}}) = \frac{\exp(\operatorname{sim}(\mathbf{Z}^{\mathcal{S}}, \mathbf{Z}^{\mathcal{S}+}_{i}) / \tau)}{\sum\nolimits_{j=0}^{N^-} \exp(\operatorname{sim}(\mathbf{Z}^{\mathcal{S}}, \mathbf{Z}^{\mathcal{S}}_j) / \tau)}.
    \end{equation}
    Then, InfoNCE loss is given as the negative log-likelihood of identifying the positive:
    \begin{equation}
        \mathcal{L}_{\text{InfoNCE}} = -\mathbb{E}_{p} \left[ \log q(i^+|\mathbf{Z}^{\mathcal{S}}) \right].
    \end{equation}
    According to Theorem 1 from~\citet{poole2019variational}, the mutual information can be lower-bounded as:
    \begin{equation}
        I(\mathbf{Z}^{\mathcal{S}}; \mathbf{Z}^{\mathcal{S}+}) \geqslant  - \mathcal{L}_{\text{InfoNCE}}.
    \end{equation}
    This bound becomes tighter as $N^-$ increases. In practice, $N^-$ is the number of negative samples used per anchor.
    Let $\operatorname{sim}(\mathbf{Z}^{\mathcal{S}}_i, \mathbf{Z}^{\mathcal{S}}_j) = \langle \mathbf{Z}^{\mathcal{S}}_i, \mathbf{Z}^{\mathcal{S}}_j \rangle$ denote dot-product similarity, the explicit form becomes:
    \begin{equation}
        \mathcal{L}_{\text{InfoNCE}} = \log \Bigg( \sum\nolimits_{j=0}^{N^-} \exp\frac{\langle \mathbf{Z}^{\mathcal{S}}_i, \mathbf{Z}^{\mathcal{S}}_j \rangle}{\tau}  \Bigg) - \frac{\langle \mathbf{Z}^{\mathcal{S}}_i, \mathbf{Z}^{\mathcal{S}+}_{i} \rangle}{\tau}.
    \end{equation}
    Thus, combining the above gives the desired lower bound:
    \begin{equation}
        I(\mathbf{Z}^{\mathcal{S}}; \mathbf{Z}^{\mathcal{S}+}) \geqslant - \mathcal{L}_{\text{InfoNCE}}.   
    \end{equation}
    This concludes the proof.
\end{proof}

\subsection{Proof of Derivation~\ref{deriv:deriv_2}}
\label{sec:proof_2}
We first restate Derivation~\ref{deriv:deriv_2} for reference.
\begin{mathbox}
    \textbf{Derivation 2} (Upper Bound of $I(\mathbf{Z}^{\mathcal{S}}; \mathbf{X}^{\mathcal{S}\prime})$)\textbf{.}
    Compression term $I(\mathbf{Z}^{\mathcal{S}}; \mathbf{X}^{\mathcal{S}\prime})$ is upper-bounded by the KL divergence between posterior $q(\mathbf{Z}^{\mathcal{S}}|\mathbf{X}^{\mathcal{S}\prime})$ and prior $p(\mathbf{Z}^{\mathcal{S}})$:
    \begin{equation}
        I(\mathbf{Z}^{\mathcal{S}}; \mathbf{X}^{\mathcal{S}\prime})\leqslant \frac{1}{N} \sum\nolimits_{i=1}^{N}\operatorname{KL}[q(\mathbf{Z}_i^{\mathcal{S}}|\mathbf{X}_i^{\mathcal{S}\prime})~\|~p(\mathbf{Z}_i^{\mathcal{S}})],
    \notag
    \end{equation}
    We assume a standard Gaussian prior $p(\mathbf{Z}^{\mathcal{S}}) = \mathcal{N}(\boldsymbol{0}, \mathbf{I})$, and apply the graph encoder to produce the mean $\boldsymbol{\mu}_i$ and the log-variance $\log\boldsymbol{\sigma}_i^2$ of $q(\mathbf{Z}^{\mathcal{S}}_i|\mathbf{X}_i^{\mathcal{S}\prime})$ for each node $v_i$, from which we sample $\mathbf{Z}^{\mathcal{S}}_i$ via reparameterization trick:
    \begin{equation}
        \mathbf{Z}^{\mathcal{S}}_i \sim q(\mathbf{Z}^{\mathcal{S}}_i|\mathbf{X}_j^{\mathcal{S}\prime}) = \mathcal{N}\big(\boldsymbol{\mu}_i, \operatorname{diag}(\boldsymbol{\sigma}_i^2)\big).\notag
    \end{equation}
\end{mathbox}

\begin{proof}
    The mutual information between the representation $\mathbf{Z}^{\mathcal{S}}$ and input $\mathbf{X}^{\mathcal{S}\prime}$ is given by:
    \begin{equation}
        I(\mathbf{Z}^{\mathcal{S}}; \mathbf{X}^{\mathcal{S}\prime}) = \mathbb{E}_{p(\mathbf{X}^{\mathcal{S}\prime})} \left[ \operatorname{KL}(q(\mathbf{Z}^{\mathcal{S}}|\mathbf{X}^{\mathcal{S}\prime}) ~\|~ q(\mathbf{Z}^{\mathcal{S}})) \right]\!.\!\!\!\!
    \end{equation}
    However, the marginal distribution $q(\mathbf{Z}^{\mathcal{S}})$ is generally intractable. To obtain a tractable upper bound, we follow the Deep Variational Information Bottleneck (VIB)~\cite{alemi2017deep}, which replaces the intractable marginal $q(\mathbf{Z}^{\mathcal{S}})$ with a simple prior $p(\mathbf{Z}^{\mathcal{S}})$, yielding:
    \begin{equation}
        \!\!I(\mathbf{Z}^{\mathcal{S}}; \mathbf{X}^{\mathcal{S}\prime}) \leqslant \mathbb{E}_{p(\mathbf{X}^{\mathcal{S}\prime})} \left[ \operatorname{KL}(q(\mathbf{Z}^{\mathcal{S}}|\mathbf{X}^{\mathcal{S}\prime}) ~\|~ p(\mathbf{Z}^{\mathcal{S}})) \right]\!.
    \end{equation}
    Assume that the posterior $q(\mathbf{Z}^{\mathcal{S}}_i | \mathbf{X}^{\mathcal{S}\prime}_i)$ is modeled as a multivariate Gaussian with diagonal covariance:
    \begin{equation}
        \!\!q(\mathbf{Z}^{\mathcal{S}}_i|\mathbf{X}^{\mathcal{S}\prime}_i) = \mathcal{N}(\boldsymbol{\mu}_i, \operatorname{diag}(\boldsymbol{\sigma}_i^2)),~ p(\mathbf{Z}^{\mathcal{S}}_i) = \mathcal{N}(\boldsymbol{0}, \mathbf{I}).\!
    \end{equation}
    Then, the KL divergence for each node $v_i$ becomes:
    \begin{align}
        &\operatorname{KL}\left[q(\mathbf{Z}^{\mathcal{S}}_i|\mathbf{X}^{\mathcal{S}\prime}_i) ~\|~ p(\mathbf{Z}^{\mathcal{S}}_i)\right] \notag\\
        =~&\frac{1}{2} \sum\nolimits_{k=1}^{d} \left( \boldsymbol{\sigma}_{i,k}^2 + \boldsymbol{\mu}_{i,k}^2 - \boldsymbol{1} - \log \boldsymbol{\sigma}_{i,k}^2 \right).
    \end{align}
    Thus, the compression term is upper-bounded by the mean KL divergence across all nodes:
    \begin{equation}
        \!\!I(\mathbf{Z}^{\mathcal{S}}; \mathbf{X}^{\mathcal{S}\prime}) \!\leqslant\! \frac{1}{N} \sum\nolimits_{i=1}^{N} \!\!\operatorname{KL}\left[q(\mathbf{Z}_i|\mathbf{X}^{\mathcal{S}\prime}_i) ~\|~ p(\mathbf{Z}_i)\right]\!.
    \end{equation}
    This bound is tight when the posterior matches the prior and serves as the regularization term in our SS-IB loss. This concludes the proof.
\end{proof}

\section{Experiment Settings}
\label{sec:exp_setting}
In this section, we present experiment details of \modelname\footnote{\url{https://anonymous.4open.science/r/SA2GFM}.}. Dataset statistics are summarized in Table~\ref{tab:datasets}.
\begin{table*}[!t]
\centering
\resizebox{\linewidth}{!}{
    \begin{tabular}{llrrrrr}
    \toprule
    \textbf{Dataset} & \textbf{Domain} & \textbf{\# Nodes} & \textbf{\# Edges} & \textbf{\makecell[r]{\# Feature \\ Dimensions}} & \textbf{\# Classes} & \textbf{\makecell[r]{Avg. \\ \# Deg.}} \\
    \midrule
    \texttt{Cora}~\cite{mccallum2000automating}       & Citation      & 2,708     & 5,429       & 1,433 & 7  & 4.00 \\
    \texttt{CiteSeer}~\cite{giles1998citeseer}    & Citation      & 3,186     & 4,277       & 3,703 & 6  & 2.57 \\
    \texttt{PubMed}~\cite{sen2008collective}      & Citation      & 19,717    & 44,338      & 500   & 3  & 4.50 \\
    \texttt{ogbn-arXiv}~\cite{hu2020open} & Citation      & 169,343   & 1,166,243   & 128   & 40 & 13.77 \\
    \midrule
    \texttt{ogbn-Tech}~\cite{hu2020open}  & Products   & 47,428    & 2,077,241   & 100   & 3  & 87.60 \\
    \texttt{ogbn-Home}~\cite{hu2020open}  & Products   & 9,790     & 131,841     & 100   & 5  & 26.93 \\
    \midrule
    \texttt{Wiki-CS}~\cite{mernyei2020wikics}      & Web Page      & 11,701    & 216,123     & 300   & 10 & 36.94 \\
    \bottomrule
    \end{tabular}%
}
\vspace{-0.2cm}
\caption{Statistics of the benchmark graph datasets.}
\label{tab:datasets}
\end{table*}

\begin{table*}[!t] 
 \centering 
 \resizebox{\linewidth}{!}{ 
 \setlength{\tabcolsep}{25pt} 
 {\renewcommand{\arraystretch}{0.95} 
     \begin{tabular}{lll} 
         \toprule 
         \textbf{Type} & \textbf{Target Dataset} & \textbf{Source Datasets} \\ 
         \midrule 
         \multirow{4}{*}{Cross-Dataset} & \texttt{Cora} & \texttt{CiteSeer, PubMed, ogbn-Home, Wiki-CS} \\ 
         & \texttt{CiteSeer} & \texttt{Cora, PubMed, ogbn-Home, Wiki-CS} \\ 
         & \texttt{PubMed} & \texttt{Cora, CiteSeer, ogbn-Home, Wiki-CS} \\ 
         &  \texttt{obgn-Tech} & \texttt{Cora, CiteSeer, PubMed, ogbn-Home, Wiki-CS} \\ 
         \midrule 
         \multirow{3}{*}{Cross-Domain} & \texttt{ogbn-Home} & \texttt{Cora, CiteSeer, PubMed, Wiki-CS} \\ 
         & \texttt{Wiki-CS} & \texttt{Cora, CiteSeer, PubMed, ogbn-Home} \\ 
         & \texttt{ogbn-arXiv} & \texttt{ogbn-Home, ogbn-Tech, Wiki-CS} \\ 
         \bottomrule 
     \end{tabular}%
     } 
 } 
 \vspace{-0.2cm} 
 \caption{Configuration of the pre-training source datasets for each target dataset.}
 \label{tab:data_config} 
 \end{table*}

\subsection{Datasets}
To evaluate multi-domain capabilities, we select \textbf{seven} graph datasets across \textbf{three} distinct domains in contrast to conventional settings where each dataset is treated as an independent domain. These datasets vary in structure, scale, and feature distribution, providing a diverse and challenging testbed for evaluating pre-trained graph models. A detailed statistical summary of these datasets is presented in Table~\ref{tab:datasets}.

\subsubsection{Citation Domain.}
This domain includes \textbf{four} widely used citation networks. Each dataset consists of a single graph, where nodes denote academic papers, edges denote citation relationships, and node labels correspond to the category or research topic of each paper.
\begin{itemize}[leftmargin=*]
    \item \texttt{Cora}~\cite{mccallum2000automating}: A well-known citation network consisting of 2,708 machine learning papers divided into seven categories (\eg, Neural Networks, Reinforcement Learning). The graph contains 5,429 citation links. Each paper is represented by a 1,433-dimensional binary bag-of-words vector, indicating the presence or absence of words from a predefined dictionary.
    \item \texttt{CiteSeer}~\cite{giles1998citeseer}: Includes 3,186 scientific publications across six fields. It features 4,277 citation links. Similar to \texttt{Cora}, each publication is described by a 3,703-dimensional sparse binary vector representing its vocabulary.
    \item \texttt{PubMed}~\cite{sen2008collective}: A larger biomedical citation network derived from the PubMed database. It contains 19,717 research articles related to diabetes, classified into three categories. Each article is embedded by a 500-dimensional feature vector based on TF-IDF.
    \item \texttt{ogbn-arXiv}~\cite{hu2020open}: A large-scale citation graph, representing citations among 169,343 Computer Science papers from arXiv. The task is to predict the paper's subject area from 40 categories. Node features are 128-dimensional vectors derived from the average embeddings of words in the title and abstract.
\end{itemize}

\subsubsection{Products Domain.}
This domain includes \textbf{two} large-scale networks from OGB \cite{hu2020open}, capturing relationships between products frequently co-purchased on Amazon. Nodes represent products with a 100-dimensional vector, and an edge indicates a co-purchase relationship, making them useful for recommendation-related tasks. 
\begin{itemize}[leftmargin=*]
    \item \texttt{ogbn-Tech} \cite{hu2020open}: A technology products network, featuring 47,428 nodes and 3 product categories. 
    \item \texttt{ogbn-Home} \cite{hu2020open}: A home goods-related products network, consisting of 9,790 nodes and 5 product categories.
\end{itemize}

\subsubsection{Web Page Domain.}
This domain features a graph constructed from web pages and their hyperlinks.
\begin{itemize}[leftmargin=*, itemsep=1pt]
\item \texttt{Wiki-CS}~\cite{mernyei2020wikics}: A graph based on 11,701 Wikipedia articles on computer science topics, connected by 216,123 hyperlinks. Nodes are categorized into 10 areas. This dataset is specifically designed for robust evaluation with multiple data splits to prevent overfitting to a particular training set.
\end{itemize}

\subsection{Baselines}
We benchmark \modelname~against nine state-of-the-art methods spanning \textbf{four} major categories. 

\subsubsection{Classic GNNs.}
Includes vanilla \texttt{GCN}~\cite{kipf2017semisupervised} and \texttt{GAT}~\cite{veličković2018graph} trained from scratch on downstream tasks without any pre-training. These models serve as fundamental baselines for evaluating representation learning performance.

\subsubsection{Single-Domain Graph Pre-training.}
This category includes baselines that learn transferable representations by optimizing auxiliary tasks without annotations.
\begin{itemize}[leftmargin=*]
\item \texttt{GraphPrompt}~\cite{liu2023graphprompt}: Introduces a prompt-based learning paradigm for graph pre-training to improve sample efficiency. It designs task-specific prompts to guide the model, enabling effective adaptation to downstream tasks with minimal fine-tuning.
\item \texttt{GraphCL}~\cite{you2020graphcl}: A general framework for learning node and graph representations through contrastive learning. It applies various data augmentations to the input graph and aims to maximize the agreement between representations of different augmented views.
\item \texttt{DGI}~\cite{velickovic2019deep}: An unsupervised approach for learning node representations by maximizing the mutual information between local patch representations and corresponding global graph summaries, achieved by contrasting the real graph with corrupted versions.
\end{itemize}

\subsubsection{Multi-Domain Graph Pre-training.}
These methods aim to learn representations that generalize across heterogeneous domains by exploiting shared patterns.
\begin{itemize}[leftmargin=*]
\item \texttt{MDGPT}~\cite{yu2024text}: A text-free framework that aligns domains through learnable domain tokens.
\item \texttt{GraphBridge}~\cite{ju2025graphbridge}: A multi-domain graph pre-training and prompt learning framework that minimizes generalization error bounds. It introduces domain-invariant aligners and a lightweight MoE router for selective knowledge transfer.
\item \texttt{GCOPE}~\cite{zhao2024all}: A framework that enhances graph prompting through knowledgeable contextualization and prompt ensembling. It combines structural and semantic information in prompts and uses ensembles to improve robustness in few-shot settings.
\end{itemize}

\subsubsection{Robust GFMs.} The most related methods focus on enhancing the robustness of multi-domain pre-trained GFMs.
\begin{itemize}[leftmargin=*]
    \item \texttt{MDGFM}~\cite{wang2025multidomain}: A unified framework for robust multi-domain knowledge transfer that aligns graph topologies and eliminates noise via graph structure learning. It leverages domain tokens, balance tokens, and dual-prompt to ensure feature and topology alignment.
\end{itemize}

\subsection{Experimental Setting Details}
We introduce the detailed experimental settings.
\subsubsection{Settings for Section~\ref{sec:exp_main} and \ref{sec:exp_degree} (\textit{RQ1}, \textit{RQ2}).}
This section evaluates the effectiveness and robustness of \modelname~under the few-shot setting, using a ($C$-way) 5-shot configuration with 5 labeled samples per class. 

\textbf{Cross-Dataset and Cross-Domain Settings.}
We set two types of domain adaptation scenarios, with the detailed configurations shown in Table~\ref{tab:data_config}. 
\textbf{(1)} For cross-dataset transfer, the target dataset is \textbf{unseen} during pre-training but comes from the \textbf{same} domain as the source datasets.
\textbf{(2)} For cross-domain transfer, the target dataset originates from a \textbf{different} domain altogether.
It is worth noting that, while some prior works on cross-domain graph pre-training and fine-tuning~\cite{zhao2024all, yu2024text} treat each dataset as a separate domain, we argue that such treatment is overly coarse. Instead, we define domains based on broader semantic or application categories. As a result, our cross-dataset setting corresponds to what previous literature often calls cross-domain, whereas our cross-domain setting reflects a higher-level shift across distinct fields.

\textbf{5-shot Fine-tuning Settings.}
We randomly sample 5 labeled instances (nodes or graphs) per class to construct the support set, with the remaining labeled samples as the query set. To ensure stable evaluation and reduce sampling bias, we conduct 20 runs with different seeds and report the mean accuracy and standard deviation.
We evaluate \modelname~on both node and graph classification tasks. For graph classification, we build input graphs by extracting ego-graphs centered at target nodes.

\textbf{Robustness against Non-targeted Attack Settings.}
To evaluate the robustness of \modelname~under non-targeted perturbations (random noise) during fine-tuning, we introduce random noise to the support set from two complementary aspects: feature and structure.
\textbf{(1)} For feature noise, we inject Gaussian perturbations into each feature dimension of the support samples, defined as $\lambda \cdot r \cdot \epsilon$, where $r$ denotes the reference amplitude of the original features, $\epsilon \sim \mathcal{N}(0,1)$ represents standard Gaussian noise, and $\lambda \in$ \{0.4, 0.8\} controls the noise level.
\textbf{(2)} For structure noise, we randomly alter the support graph topology by randomly deleting a proportion $\lambda$ of the edges. These controlled non-targeted attacks allow us to systematically assess the robustness of \modelname~in few-shot adaptation scenarios.

\textbf{Robustness against Targeted Attack Settings.}
To further assess the adversarial robustness of \modelname, we adopt NETTACK~\cite{zugner2018adversarial}, a widely applied targeted attack framework for graph data that perturbs either node features or edges of selected target nodes. We evaluate two standard attack modes:
\textbf{(1)} Evasive attack, where the model is trained on clean support sets and adversarial perturbations are applied only during inference on the query set.
\textbf{(2)} Poisoning attack, where perturbations are introduced into the support set before training.
In both settings, we follow NETTACK’s default strategy to select vulnerable nodes and use GAT~\cite{veličković2018graph} as the surrogate model with default parameters. The number of allowed perturbations is set as $n \in $ \{1, 2, 3\}.

\subsubsection{Detailed Settings for Section~\ref{sec:exp_component_analysis} (\textit{RQ3}).}
To ensure a fair comparison, the ablation study is conducted on four representative datasets: \texttt{Cora}, \texttt{ogbn-Home}, \texttt{Wiki-CS}, and \texttt{ogbn-arXiv}, focusing on the 5-shot node classification task. To clearly observe each component's contribution, we perform the evaluation under moderate attack intensity. For the non-targeted attack, we set the perturbation rate to $\lambda = 0.4$ (covering both feature noise and edge removal). For the targeted attack, we use NETTACK with $p = 1$ perturbation per node for both evasion and poisoning scenarios. All ablated variants adhere to the same evaluation protocol and hyperparameter settings as the full model, ensuring that any observed performance differences are directly and fairly attributable to the specific module that was removed.

\subsubsection{Detailed Settings for Section~\ref{sec:exp_hyperparam_analysis} (\textit{RQ4}).}
To evaluate performance stability, we conduct a sensitivity analysis on four key hyperparameters: the InfoNCE temperature $\tau$, the Information Bottleneck weight $\lambda_{\text{IB}}$, the MoE regularization weight $\lambda_m$, and the uncertainty loss weight $\lambda_u$. This analysis is performed on the \texttt{Cora} dataset for 5-shot node classification under a fixed targeted poisoning attack ($p=1$). We vary one hyperparameter at a time within a predefined range ($\tau \in \{0.1, 0.5, 0.8, 1.0\}$; $\lambda_{\text{IB}} \in \{10^{-3}, 10^{-2}, 10^{-1}\}$; $\lambda_m, \lambda_u \in \{0.1, 0.2, 0.5, 1.0\}$) while keeping others at their optimal values to precisely isolate the individual impact of each hyperparameter on the final classification accuracy.

\subsubsection{Detailed Settings for Section~\ref{sec:exp_vis} (\textit{RQ5}).}
To intuitively demonstrate how \modelname~learns and adapts representations, we conduct a visualization analysis. We use UMAP to project high-dimensional embeddings into a 2D space for visual inspection and the Silhouette Score to quantitatively measure class clustering quality. We compare embeddings at three key stages: (1) \textbf{Pre-trained}, showing general domain features; (2) \textbf{Un-finetuned}, revealing initial alignment post-transfer; and (3) \textbf{Fine-tuned}, displaying the final task-oriented structure. This analysis is performed in both a cross-dataset (targeting \texttt{Cora}) and a cross-domain (targeting \texttt{Wiki-CS}) scenario to assess adaptability, with nodes colored according to their ground-truth class labels, which allows for a clear visual assessment of the final clustering effectiveness in these settings.

\section{Additional Results}
\label{sec:additional_results}
\begin{table*}[!t]
\vspace{-0.1cm}
\centering
\setlength{\tabcolsep}{1pt}
\renewcommand{\arraystretch}{0.95}
\begin{tabular*}{\linewidth}{@{\extracolsep{\fill}}l|cccc|ccc}
\toprule
\textbf{Method} & \texttt{Cora} & \texttt{CiteSeer} & \texttt{PubMed}
                & \texttt{obgn-Tech} & \texttt{obgn-Home}  & \texttt{obgn-arXiv} & \texttt{Wiki-CS} \\
\midrule
\texttt{GCN} &
  42.9\scalebox{0.6}{±5.8\textcolor{white}{0}} &
  42.6\scalebox{0.6}{±5.5\textcolor{white}{0}} &
  42.4\scalebox{0.6}{±7.1\textcolor{white}{0}} &
  60.6\scalebox{0.6}{±10.6} &
  54.6\scalebox{0.6}{±15.3} &
  32.8\scalebox{0.6}{±7.5\textcolor{white}{0}} &
  40.8\scalebox{0.6}{±6.7\textcolor{white}{0}} \\[-1pt]

\texttt{GAT} &
  45.4\scalebox{0.6}{±5.4\textcolor{white}{0}} &
  53.2\scalebox{0.6}{±5.6\textcolor{white}{0}} &
  46.0\scalebox{0.6}{±5.9\textcolor{white}{0}} &
  64.6\scalebox{0.6}{±10.3} &
  58.9\scalebox{0.6}{±10.8} &
  40.1\scalebox{0.6}{±8.7\textcolor{white}{0}} &
  45.7\scalebox{0.6}{±5.4\textcolor{white}{0}} \\
\midrule
\texttt{DGI} &
  54.6\scalebox{0.6}{±6.5\textcolor{white}{0}} &
  53.6\scalebox{0.6}{±4.5\textcolor{white}{0}} &
  52.3\scalebox{0.6}{±5.7\textcolor{white}{0}} &
  70.2\scalebox{0.6}{±7.7\textcolor{white}{0}} &
  68.1\scalebox{0.6}{±9.6\textcolor{white}{0}} &
  43.4\scalebox{0.6}{±5.9\textcolor{white}{0}} &
  45.1\scalebox{0.6}{±5.1\textcolor{white}{0}} \\[-1pt]

\texttt{GRAPHCL} &
  52.2\scalebox{0.6}{±6.6\textcolor{white}{0}} &
  54.5\scalebox{0.6}{±5.1\textcolor{white}{0}} &
  48.8\scalebox{0.6}{±5.3\textcolor{white}{0}} &
  73.3\scalebox{0.6}{±6.6\textcolor{white}{0}} &
  68.0\scalebox{0.6}{±12.3} &
  42.8\scalebox{0.6}{±5.9\textcolor{white}{0}} &
  48.9\scalebox{0.6}{±4.2\textcolor{white}{0}} \\[-1pt]

\texttt{GraphPrompt} &
  50.3\scalebox{0.6}{±4.8\textcolor{white}{0}} &
  55.1\scalebox{0.6}{±5.3\textcolor{white}{0}} &
  51.8\scalebox{0.6}{±5.6\textcolor{white}{0}} &
  75.3\scalebox{0.6}{±9.4\textcolor{white}{0}} &
  64.1\scalebox{0.6}{±9.3\textcolor{white}{0}} &
  43.2\scalebox{0.6}{±5.6\textcolor{white}{0}} &
  47.8\scalebox{0.6}{±5.0\textcolor{white}{0}} \\
\bottomrule
\end{tabular*}
\vspace{-0.3cm}
\caption{Accuracy (\% ± std. for 20 runs) of \textbf{5-shot node classification} under non-targeted feature attack ($\lambda=$~0.4).}
\label{tab:d1}
\vspace{-0.2cm}
\end{table*}

\begin{table*}[!t]
\centering
\setlength{\tabcolsep}{1pt}
\renewcommand{\arraystretch}{0.95}
\begin{tabular*}{\linewidth}{@{\extracolsep{\fill}}l|cccc|ccc}
\toprule
\textbf{Method} & \texttt{Cora} & \texttt{CiteSeer} & \texttt{PubMed}
                & \texttt{obgn-Tech} & \texttt{obgn-Home}  & \texttt{obgn-arXiv} & \texttt{Wiki-CS} \\
\midrule
\texttt{GCN} &
  45.9\scalebox{0.6}{±9.8\textcolor{white}{0}} &
  36.3\scalebox{0.6}{±5.1\textcolor{white}{0}} &
  42.3\scalebox{0.6}{±9.5\textcolor{white}{0}} &
  56.0\scalebox{0.6}{±12.0} &
  39.5\scalebox{0.6}{±9.9\textcolor{white}{0}} &
  35.7\scalebox{0.6}{±5.8\textcolor{white}{0}} &
  32.2\scalebox{0.6}{±7.0\textcolor{white}{0}} \\[-1pt]

\texttt{GAT} &
  40.4\scalebox{0.6}{±7.5\textcolor{white}{0}} &
  45.7\scalebox{0.6}{±7.2\textcolor{white}{0}} &
  40.6\scalebox{0.6}{±10.0} &
  50.8\scalebox{0.6}{±8.4\textcolor{white}{0}} &
  41.3\scalebox{0.6}{±9.4\textcolor{white}{0}} &
  36.1\scalebox{0.6}{±3.9\textcolor{white}{0}} &
  35.8\scalebox{0.6}{±7.1\textcolor{white}{0}} \\
\midrule
\texttt{DGI} &
  46.0\scalebox{0.6}{±5.7\textcolor{white}{0}} &
  44.3\scalebox{0.6}{±5.9\textcolor{white}{0}} &
  42.6\scalebox{0.6}{±9.0\textcolor{white}{0}} &
  56.8\scalebox{0.6}{±5.4\textcolor{white}{0}} &
  45.8\scalebox{0.6}{±7.7\textcolor{white}{0}} &
  35.3\scalebox{0.6}{±4.3\textcolor{white}{0}} &
  39.9\scalebox{0.6}{±6.6\textcolor{white}{0}} \\[-1pt]

\texttt{GRAPHCL} &
  50.9\scalebox{0.6}{±6.0\textcolor{white}{0}} &
  47.4\scalebox{0.6}{±6.6\textcolor{white}{0}} &
  48.4\scalebox{0.6}{±8.4\textcolor{white}{0}} &
  58.1\scalebox{0.6}{±3.9\textcolor{white}{0}} &
  43.9\scalebox{0.6}{±8.9\textcolor{white}{0}} &
  35.1\scalebox{0.6}{±5.4\textcolor{white}{0}} &
  35.3\scalebox{0.6}{±5.6\textcolor{white}{0}} \\
\bottomrule
\end{tabular*}
\vspace{-0.3cm}
\caption{Accuracy (\% ± std. for 20 runs) of \textbf{5-shot node classification} under non-targeted feature attack ($\lambda=$~0.8).}
\label{tab:d2}
\vspace{-0.2cm}
\end{table*}

\begin{table*}[!t]
\centering
\setlength{\tabcolsep}{1pt}
\renewcommand{\arraystretch}{0.95}
\begin{tabular*}{\linewidth}{@{\extracolsep{\fill}}l|cccc|ccc}
\toprule
\textbf{Method} & \texttt{Cora} & \texttt{CiteSeer} & \texttt{PubMed}
                & \texttt{obgn-Tech} & \texttt{obgn-Home}  & \texttt{obgn-arXiv} & \texttt{Wiki-CS} \\
\midrule
\texttt{GCN} &
  51.1\scalebox{0.6}{±5.8\textcolor{white}{0}} &
  41.3\scalebox{0.6}{±6.9\textcolor{white}{0}} &
  61.1\scalebox{0.6}{±6.5\textcolor{white}{0}} &
  67.4\scalebox{0.6}{±11.6} &
  60.3\scalebox{0.6}{±18.0} &
  41.9\scalebox{0.6}{±10.1} &
  43.8\scalebox{0.6}{±11.7} \\[-1pt]

\texttt{GAT} &
  48.1\scalebox{0.6}{±5.8\textcolor{white}{0}} &
  45.5\scalebox{0.6}{±7.2\textcolor{white}{0}} &
  59.3\scalebox{0.6}{±5.9\textcolor{white}{0}} &
  69.0\scalebox{0.6}{±10.0} &
  65.8\scalebox{0.6}{±7.3\textcolor{white}{0}} &
  43.8\scalebox{0.6}{±8.5\textcolor{white}{0}} &
  45.1\scalebox{0.6}{±4.0\textcolor{white}{0}} \\[-1pt]
\midrule
\texttt{DGI} &
  52.9\scalebox{0.6}{±5.2\textcolor{white}{0}} &
  46.4\scalebox{0.6}{±5.3\textcolor{white}{0}} &
  65.4\scalebox{0.6}{±6.2\textcolor{white}{0}} &
  71.9\scalebox{0.6}{±8.3\textcolor{white}{0}} &
  69.6\scalebox{0.6}{±7.6\textcolor{white}{0}} &
  47.4\scalebox{0.6}{±6.8\textcolor{white}{0}} &
  49.1\scalebox{0.6}{±4.8\textcolor{white}{0}} \\[-1pt]

\texttt{GRAPHCL} &
  58.5\scalebox{0.6}{±5.4\textcolor{white}{0}} &
  48.1\scalebox{0.6}{±5.7\textcolor{white}{0}} &
  63.8\scalebox{0.6}{±6.2\textcolor{white}{0}} &
  76.9\scalebox{0.6}{±6.3\textcolor{white}{0}} &
  66.4\scalebox{0.6}{±8.1\textcolor{white}{0}} &
  45.3\scalebox{0.6}{±6.1\textcolor{white}{0}} &
  50.5\scalebox{0.6}{±3.5\textcolor{white}{0}} \\[-1pt]

\texttt{GraphPrompt} &
  57.1\scalebox{0.6}{±5.3\textcolor{white}{0}} &
  43.5\scalebox{0.6}{±5.8\textcolor{white}{0}} &
  55.5\scalebox{0.6}{±5.7\textcolor{white}{0}} &
  73.8\scalebox{0.6}{±10.2} &
  70.6\scalebox{0.6}{±8.7\textcolor{white}{0}} &
  48.6\scalebox{0.6}{±5.0\textcolor{white}{0}} &
  60.1\scalebox{0.6}{±4.2\textcolor{white}{0}} \\
\bottomrule
\end{tabular*}
\vspace{-0.3cm}
\caption{Accuracy (\% ± std. for 20 runs) of \textbf{5-shot graph classification} under non-targeted feature attack ($\lambda=$~0.4).}
\label{tab:d3}
\vspace{-0.2cm}
\end{table*}

\begin{table*}[!t]
\centering
\setlength{\tabcolsep}{1pt}
\renewcommand{\arraystretch}{0.95}
\begin{tabular*}{\linewidth}{@{\extracolsep{\fill}}l|cccc|ccc}
\toprule
\textbf{Method} & \texttt{Cora} & \texttt{CiteSeer} & \texttt{PubMed}
                & \texttt{obgn-Tech} & \texttt{obgn-Home}  & \texttt{obgn-arXiv} & \texttt{Wiki-CS} \\
\midrule
\texttt{GCN} & 40.4\scalebox{0.6}{±7.1\textcolor{white}{0}}
             & 37.7\scalebox{0.6}{±6.1\textcolor{white}{0}}
             & 50.8\scalebox{0.6}{±7.6\textcolor{white}{0}}
             & 63.9\scalebox{0.6}{±11.3}
             & 54.1\scalebox{0.6}{±16.7}
             & 33.9\scalebox{0.6}{±7.9\textcolor{white}{0}}
             & 40.0\scalebox{0.6}{±6.5\textcolor{white}{0}} \\

\texttt{GAT} & 42.1\scalebox{0.6}{±7.2\textcolor{white}{0}}
             & 42.8\scalebox{0.6}{±5.4\textcolor{white}{0}}
             & 53.8\scalebox{0.6}{±5.9\textcolor{white}{0}}
             & 63.1\scalebox{0.6}{±10.8}
             & 60.5\scalebox{0.6}{±8.4\textcolor{white}{0}}
             & 43.1\scalebox{0.6}{±8.7\textcolor{white}{0}}
             & 42.9\scalebox{0.6}{±5.4\textcolor{white}{0}} \\
\midrule
\texttt{DGI} & 47.8\scalebox{0.6}{±6.9\textcolor{white}{0}}
             & 41.7\scalebox{0.6}{±5.9\textcolor{white}{0}}
             & 61.1\scalebox{0.6}{±4.8\textcolor{white}{0}}
             & 65.0\scalebox{0.6}{±9.0\textcolor{white}{0}}
             & 60.2\scalebox{0.6}{±10.1}
             & 42.9\scalebox{0.6}{±5.9\textcolor{white}{0}}
             & 45.9\scalebox{0.6}{±5.5\textcolor{white}{0}} \\

\texttt{GRAPHCL} & 51.9\scalebox{0.6}{±5.6\textcolor{white}{0}}
                 & 46.3\scalebox{0.6}{±6.0\textcolor{white}{0}}
                 & 59.8\scalebox{0.6}{±6.4\textcolor{white}{0}}
                 & 70.1\scalebox{0.6}{±7.8\textcolor{white}{0}}
                 & 61.2\scalebox{0.6}{±10.6}
                 & 42.0\scalebox{0.6}{±6.7\textcolor{white}{0}}
                 & 44.2\scalebox{0.6}{±3.8\textcolor{white}{0}} \\
\bottomrule
\end{tabular*}
\vspace{-0.2cm}
\caption{Accuracy (\% ± std. for 20 runs) of \textbf{5-shot graph classification} under non-targeted feature attack ($\lambda=$~0.8).}
\label{tab:d4}
\end{table*}

\begin{table*}[!t]
\vspace{-0.1cm}
\centering
\setlength{\tabcolsep}{1pt}          
\renewcommand{\arraystretch}{0.95}   

\begin{tabular*}{\linewidth}{@{\extracolsep{\fill}}l|cccc|ccc}
\toprule
\textbf{Method} & \texttt{Cora} & \texttt{CiteSeer} & \texttt{PubMed}
                & \texttt{obgn-Tech} & \texttt{obgn-Home}  & \texttt{obgn-arXiv} & \texttt{Wiki-CS} \\
\midrule
\texttt{GCN} &
  48.5\scalebox{0.6}{±8.5} &
  37.9\scalebox{0.6}{±5.8\textcolor{white}{0}} &
  40.9\scalebox{0.6}{±9.8\textcolor{white}{0}} &
  56.6\scalebox{0.6}{±13.2} &
  51.0\scalebox{0.6}{±12.9} &
  38.8\scalebox{0.6}{±4.7\textcolor{white}{0}} &
  42.6\scalebox{0.6}{±9.9\textcolor{white}{0}} \\

\texttt{GAT} &
  52.8\scalebox{0.6}{±7.1\textcolor{white}{0}} &
  42.3\scalebox{0.6}{±5.4\textcolor{white}{0}} &
  44.3\scalebox{0.6}{±9.1\textcolor{white}{0}} &
  55.4\scalebox{0.6}{±7.4\textcolor{white}{0}} &
  53.6\scalebox{0.6}{±6.7\textcolor{white}{0}} &
  41.4\scalebox{0.6}{±5.6\textcolor{white}{0}} &
  46.6\scalebox{0.6}{±4.9\textcolor{white}{0}} \\

\midrule
\texttt{DGI} &
  58.4\scalebox{0.6}{±5.0\textcolor{white}{0}} &
  47.1\scalebox{0.6}{±6.5\textcolor{white}{0}} &
  41.3\scalebox{0.6}{±7.5\textcolor{white}{0}} &
  67.6\scalebox{0.6}{±2.7\textcolor{white}{0}} &
  61.0\scalebox{0.6}{±6.2\textcolor{white}{0}} &
  46.6\scalebox{0.6}{±4.9\textcolor{white}{0}} &
  44.1\scalebox{0.6}{±7.7\textcolor{white}{0}} \\

\texttt{GRAPHCL} &
  53.6\scalebox{0.6}{±4.7\textcolor{white}{0}} &
  49.6\scalebox{0.6}{±6.5\textcolor{white}{0}} &
  48.9\scalebox{0.6}{±8.8\textcolor{white}{0}} &
  68.7\scalebox{0.6}{±2.1\textcolor{white}{0}} &
  48.9\scalebox{0.6}{±7.8\textcolor{white}{0}} &
  43.0\scalebox{0.6}{±4.9\textcolor{white}{0}} &
  49.3\scalebox{0.6}{±5.6\textcolor{white}{0}} \\
\bottomrule
\end{tabular*}
\vspace{-0.3cm}
\caption{Accuracy (\% ± std. for 20 runs) of \textbf{5-shot node classification} under non-targeted structure attack ($\lambda=$~0.4).}
\label{tab:d5}
\vspace{-0.2cm}
\end{table*}

\begin{table*}[!t]
\centering
\setlength{\tabcolsep}{1pt}
\renewcommand{\arraystretch}{0.95}

\begin{tabular*}{\linewidth}{@{\extracolsep{\fill}}l|cccc|ccc}
\toprule
\textbf{Method} & \texttt{Cora} & \texttt{CiteSeer} & \texttt{PubMed} & \texttt{obgn-Tech} & \texttt{obgn-Home} & \texttt{obgn-arXiv} & \texttt{Wiki-CS} \\
\midrule
\texttt{GCN} &
 44.3\scalebox{0.6}{±9.1\textcolor{white}{0}} &
 33.9\scalebox{0.6}{±6.0\textcolor{white}{0}} &
 36.0\scalebox{0.6}{±10.5} &
 51.4\scalebox{0.6}{±13.9} &
 43.2\scalebox{0.6}{±10.9} &
 35.1\scalebox{0.6}{±6.0\textcolor{white}{0}} &
 37.4\scalebox{0.6}{±6.9\textcolor{white}{0}} \\

\texttt{GAT} &
 42.8\scalebox{0.6}{±6.3\textcolor{white}{0}} &
 38.1\scalebox{0.6}{±5.1\textcolor{white}{0}} &
 39.9\scalebox{0.6}{±8.3\textcolor{white}{0}} &
 47.8\scalebox{0.6}{±7.1\textcolor{white}{0}} &
 49.1\scalebox{0.6}{±8.2\textcolor{white}{0}} &
 37.6\scalebox{0.6}{±4.9\textcolor{white}{0}} &
 39.5\scalebox{0.6}{±4.4\textcolor{white}{0}} \\
\midrule
\texttt{DGI} &
 52.3\scalebox{0.6}{±5.6\textcolor{white}{0}} &
 42.2\scalebox{0.6}{±4.8\textcolor{white}{0}} &
 41.6\scalebox{0.6}{±8.1\textcolor{white}{0}} &
 64.9\scalebox{0.6}{±8.1\textcolor{white}{0}} &
 55.9\scalebox{0.6}{±8.7\textcolor{white}{0}} &
 42.6\scalebox{0.6}{±3.7\textcolor{white}{0}} &
 42.1\scalebox{0.6}{±7.4\textcolor{white}{0}} \\

\texttt{GRAPHCL} &
 48.3\scalebox{0.6}{±5.9\textcolor{white}{0}} &
 45.0\scalebox{0.6}{±5.4\textcolor{white}{0}} &
 42.2\scalebox{0.6}{±7.3\textcolor{white}{0}} &
 64.6\scalebox{0.6}{±5.3\textcolor{white}{0}} &
 53.4\scalebox{0.6}{±8.8\textcolor{white}{0}} &
 36.4\scalebox{0.6}{±5.2\textcolor{white}{0}} &
 43.6\scalebox{0.6}{±5.5\textcolor{white}{0}} \\

\texttt{GraphPrompt} &
 52.0\scalebox{0.6}{±4.9\textcolor{white}{0}} &
 41.0\scalebox{0.6}{±6.0\textcolor{white}{0}} &
 41.5\scalebox{0.6}{±7.6\textcolor{white}{0}} &
 65.7\scalebox{0.6}{±8.2\textcolor{white}{0}} &
 54.6\scalebox{0.6}{±10.1} &
 39.3\scalebox{0.6}{±4.5\textcolor{white}{0}} &
 42.3\scalebox{0.6}{±6.8\textcolor{white}{0}} \\
\bottomrule
\end{tabular*}
\vspace{-0.3cm}
\caption{Accuracy (\% ± std. for 20 runs) of \textbf{5-shot node classification} under non-targeted structure attack ($\lambda=$~0.8).}
\label{tab:d6}
\vspace{-0.2cm}
\end{table*}
\begin{table*}[!t]
\centering
\setlength{\tabcolsep}{1pt}
\renewcommand{\arraystretch}{0.95}

\begin{tabular*}{\linewidth}{@{\extracolsep{\fill}}l|cccc|ccc}
\toprule
\textbf{Method} & \texttt{Cora} & \texttt{CiteSeer} & \texttt{PubMed}
                & \texttt{obgn-Tech} & \texttt{obgn-Home}  & \texttt{obgn-arXiv} & \texttt{Wiki-CS} \\
\midrule
\texttt{GCN} &
  44.9\scalebox{0.6}{±8.0\textcolor{white}{0}} &
  46.4\scalebox{0.6}{±5.7\textcolor{white}{0}} &
  50.9\scalebox{0.6}{±5.9\textcolor{white}{0}} &
  68.7\scalebox{0.6}{±12.9} &
  64.4\scalebox{0.6}{±13.0} &
  38.4\scalebox{0.6}{±9.0\textcolor{white}{0}} &
  52.6\scalebox{0.6}{±10.9} \\[-1pt]
\texttt{GAT} &
  47.1\scalebox{0.6}{±5.3\textcolor{white}{0}} &
  48.7\scalebox{0.6}{±7.8\textcolor{white}{0}} &
  47.9\scalebox{0.6}{±6.3\textcolor{white}{0}} &
  71.0\scalebox{0.6}{±8.4\textcolor{white}{0}} &
  69.1\scalebox{0.6}{±10.8} &
  44.5\scalebox{0.6}{±9.2\textcolor{white}{0}} &
  46.6\scalebox{0.6}{±8.2\textcolor{white}{0}} \\
\midrule
\texttt{DGI} &
  49.6\scalebox{0.6}{±5.4\textcolor{white}{0}} &
  54.3\scalebox{0.6}{±5.4\textcolor{white}{0}} &
  52.4\scalebox{0.6}{±6.7\textcolor{white}{0}} &
  71.7\scalebox{0.6}{±7.0\textcolor{white}{0}} &
  66.1\scalebox{0.6}{±11.4} &
  45.5\scalebox{0.6}{±7.4\textcolor{white}{0}} &
  52.0\scalebox{0.6}{±5.7\textcolor{white}{0}} \\[-1pt]
\texttt{GRAPHCL} &
  54.5\scalebox{0.6}{±4.6\textcolor{white}{0}} &
  55.5\scalebox{0.6}{±5.8\textcolor{white}{0}} &
  56.9\scalebox{0.6}{±6.5\textcolor{white}{0}} &
  77.1\scalebox{0.6}{±5.2\textcolor{white}{0}} &
  66.0\scalebox{0.6}{±11.3} &
  51.7\scalebox{0.6}{±6.6\textcolor{white}{0}} &
  61.6\scalebox{0.6}{±6.3\textcolor{white}{0}} \\
\bottomrule
\end{tabular*}
\vspace{-0.3cm}
\caption{Accuracy (\% ± std. for 20 runs) of \textbf{5-shot graph classification} under non-targeted structure attack ($\lambda=$~0.4).}
\label{tab:d7}
\vspace{-0.2cm}
\end{table*}

\begin{table*}[!t]
\centering
\setlength{\tabcolsep}{1pt}
\renewcommand{\arraystretch}{0.95}

\begin{tabular*}{\linewidth}{@{\extracolsep{\fill}}l|cccc|ccc}
\toprule
\textbf{Method} & \texttt{Cora} & \texttt{CiteSeer} & \texttt{PubMed}
                & \texttt{obgn-Tech} & \texttt{obgn-Home}  & \texttt{obgn-arXiv} & \texttt{Wiki-CS} \\
\midrule
\texttt{GCN} & 40.9\scalebox{0.6}{±6.9\textcolor{white}{0}} &
              39.8\scalebox{0.6}{±6.2\textcolor{white}{0}} &
              45.2\scalebox{0.6}{±9.7\textcolor{white}{0}} &
              65.5\scalebox{0.6}{±13.7} &
              52.4\scalebox{0.6}{±21.4} &
              32.8\scalebox{0.6}{±5.8\textcolor{white}{0}} &
              46.2\scalebox{0.6}{±9.9\textcolor{white}{0}} \\[-1pt]
\texttt{GAT} & 48.8\scalebox{0.6}{±6.9\textcolor{white}{0}} &
              44.7\scalebox{0.6}{±6.0\textcolor{white}{0}} &
              42.2\scalebox{0.6}{±8.6\textcolor{white}{0}} &
              69.1\scalebox{0.6}{±10.8} &
              63.8\scalebox{0.6}{±12.1} &
              41.3\scalebox{0.6}{±8.7\textcolor{white}{0}} &
              42.0\scalebox{0.6}{±5.6\textcolor{white}{0}} \\
\midrule
\texttt{DGI} & 49.3\scalebox{0.6}{±6.8\textcolor{white}{0}} &
              48.9\scalebox{0.6}{±5.9\textcolor{white}{0}} &
              47.3\scalebox{0.6}{±8.1\textcolor{white}{0}} &
              68.4\scalebox{0.6}{±8.0\textcolor{white}{0}} &
              64.4\scalebox{0.6}{±11.1} &
              42.8\scalebox{0.6}{±7.3\textcolor{white}{0}} &
              42.4\scalebox{0.6}{±3.8\textcolor{white}{0}} \\[-1pt]
\texttt{GRAPHCL} & 51.0\scalebox{0.6}{±6.0\textcolor{white}{0}} &
                  46.6\scalebox{0.6}{±6.0\textcolor{white}{0}} &
                  51.0\scalebox{0.6}{±8.3\textcolor{white}{0}} &
                  73.6\scalebox{0.6}{±7.9\textcolor{white}{0}} &
                  63.0\scalebox{0.6}{±11.1} &
                  46.2\scalebox{0.6}{±5.8\textcolor{white}{0}} &
                  51.4\scalebox{0.6}{±3.4\textcolor{white}{0}} \\
\bottomrule
\end{tabular*}
\vspace{-0.3cm}
\caption{Accuracy (\% ± std. for 20 runs) of \textbf{5-shot graph classification} under non-targeted structure attack ($\lambda=$~0.8).}
\label{tab:d8}
\vspace{-0.2cm}
\end{table*}

\begin{table*}[!t]
\centering
\setlength{\tabcolsep}{1pt}
\renewcommand{\arraystretch}{0.95}

\begin{tabular*}{\linewidth}{@{\extracolsep{\fill}}l|cccc|ccc}
\toprule
\textbf{Method} & \texttt{Cora} & \texttt{CiteSeer} & \texttt{PubMed}
                & \texttt{obgn-Tech} & \texttt{obgn-Home}  & \texttt{obgn-arXiv} & \texttt{Wiki-CS} \\
\midrule
\texttt{GCN} &
  44.8\scalebox{0.6}{±8.0\textcolor{white}{0}} &
  36.0\scalebox{0.6}{±7.4\textcolor{white}{0}} &
  43.6\scalebox{0.6}{±12.1} &
  62.1\scalebox{0.6}{±15.4} &
  49.4\scalebox{0.6}{±13.2} &
  38.2\scalebox{0.6}{±5.0\textcolor{white}{0}} &
  36.2\scalebox{0.6}{±7.6\textcolor{white}{0}} \\

\texttt{GAT} &
  42.2\scalebox{0.6}{±7.6\textcolor{white}{0}} &
  42.5\scalebox{0.6}{±5.1\textcolor{white}{0}} &
  40.8\scalebox{0.6}{±11.1} &
  63.2\scalebox{0.6}{±8.7\textcolor{white}{0}} &
  50.3\scalebox{0.6}{±6.9\textcolor{white}{0}} &
  39.1\scalebox{0.6}{±5.4\textcolor{white}{0}} &
  36.0\scalebox{0.6}{±6.1\textcolor{white}{0}} \\
\midrule
\texttt{DGI} &
  46.6\scalebox{0.6}{±6.5\textcolor{white}{0}} &
  48.4\scalebox{0.6}{±4.5\textcolor{white}{0}} &
  45.0\scalebox{0.6}{±9.9\textcolor{white}{0}} &
  76.0\scalebox{0.6}{±5.7\textcolor{white}{0}} &
  60.1\scalebox{0.6}{±7.4\textcolor{white}{0}} &
  45.5\scalebox{0.6}{±5.4\textcolor{white}{0}} &
  46.3\scalebox{0.6}{±7.5\textcolor{white}{0}} \\

\texttt{GRAPHCL} &
  52.0\scalebox{0.6}{±6.0\textcolor{white}{0}} &
  49.0\scalebox{0.6}{±4.7\textcolor{white}{0}} &
  51.8\scalebox{0.6}{±10.3} &
  66.3\scalebox{0.6}{±4.8\textcolor{white}{0}} &
  57.4\scalebox{0.6}{±8.6\textcolor{white}{0}} &
  42.5\scalebox{0.6}{±4.8\textcolor{white}{0}} &
  41.8\scalebox{0.6}{±6.1\textcolor{white}{0}} \\
\bottomrule
\end{tabular*}
\vspace{-0.2cm}
\caption{Accuracy (\% ± std. for 20 runs) of \textbf{5-shot node classification} under targeted evasion attack ($p=$~1).}
\label{tab:d9}
\vspace{-0.2cm}
\end{table*}

\begin{table*}[!t]
\centering
\setlength{\tabcolsep}{1pt}
\renewcommand{\arraystretch}{0.95}

\begin{tabular*}{\linewidth}{@{\extracolsep{\fill}}l|cccc|ccc}
\toprule
\textbf{Method} & \texttt{Cora} & \texttt{CiteSeer} & \texttt{PubMed}
                & \texttt{obgn-Tech} & \texttt{obgn-Home}  & \texttt{obgn-arXiv} & \texttt{Wiki-CS} \\
\midrule
\texttt{GCN} &
  42.4\scalebox{0.6}{±7.4\textcolor{white}{0}} &
  33.1\scalebox{0.6}{±6.7\textcolor{white}{0}} &
  37.7\scalebox{0.6}{±7.0\textcolor{white}{0}} &
  64.4\scalebox{0.6}{±10.8} &
  54.1\scalebox{0.6}{±14.6} &
  33.1\scalebox{0.6}{±6.5\textcolor{white}{0}} &
  43.1\scalebox{0.6}{±8.2\textcolor{white}{0}} \\[-1pt]

\texttt{GAT} &
  47.6\scalebox{0.6}{±5.6\textcolor{white}{0}} &
  49.6\scalebox{0.6}{±5.7\textcolor{white}{0}} &
  47.3\scalebox{0.6}{±6.6\textcolor{white}{0}} &
  60.2\scalebox{0.6}{±9.2\textcolor{white}{0}} &
  56.8\scalebox{0.6}{±10.3} &
  45.2\scalebox{0.6}{±8.8\textcolor{white}{0}} &
  47.1\scalebox{0.6}{±6.2\textcolor{white}{0}} \\
\midrule
\texttt{DGI} &
  50.3\scalebox{0.6}{±5.7\textcolor{white}{0}} &
  52.0\scalebox{0.6}{±4.5\textcolor{white}{0}} &
  49.9\scalebox{0.6}{±6.6\textcolor{white}{0}} &
  63.2\scalebox{0.6}{±8.0\textcolor{white}{0}} &
  67.0\scalebox{0.6}{±9.7\textcolor{white}{0}} &
  44.3\scalebox{0.6}{±5.6\textcolor{white}{0}} &
  49.5\scalebox{0.6}{±4.9\textcolor{white}{0}} \\[-1pt]

\texttt{GRAPHCL} &
  43.2\scalebox{0.6}{±5.7\textcolor{white}{0}} &
  53.1\scalebox{0.6}{±4.7\textcolor{white}{0}} &
  50.8\scalebox{0.6}{±6.2\textcolor{white}{0}} &
  73.7\scalebox{0.6}{±6.8\textcolor{white}{0}} &
  62.9\scalebox{0.6}{±12.0} &
  51.4\scalebox{0.6}{±5.6\textcolor{white}{0}} &
  42.2\scalebox{0.6}{±4.7\textcolor{white}{0}} \\
\bottomrule
\end{tabular*}
\vspace{-0.2cm}
\caption{Accuracy (\% ± std. for 20 runs) of \textbf{5-shot node classification} under targeted evasion attack ($p=$~2).}
\label{tab:d10}
\vspace{-0.2cm}
\end{table*}

\begin{table*}[!t]
\centering
\setlength{\tabcolsep}{1pt}
\renewcommand{\arraystretch}{0.95}

\begin{tabular*}{\linewidth}{@{\extracolsep{\fill}}l|cccc|ccc}
\toprule
\textbf{Method} & \texttt{Cora} & \texttt{CiteSeer} & \texttt{PubMed}
                & \texttt{obgn-Tech} & \texttt{obgn-Home}  & \texttt{obgn-arXiv} & \texttt{Wiki-CS} \\
\midrule
\texttt{GCN} &
  44.0\scalebox{0.6}{±6.9\textcolor{white}{0}} &
  33.2\scalebox{0.6}{±5.9\textcolor{white}{0}} &
  36.7\scalebox{0.6}{±11.0} &
  51.4\scalebox{0.6}{±16.3} &
  47.8\scalebox{0.6}{±13.6} &
  36.6\scalebox{0.6}{±5.3\textcolor{white}{0}} &
  33.8\scalebox{0.6}{±7.6\textcolor{white}{0}} \\[-1pt]
\texttt{GAT} &
  52.3\scalebox{0.6}{±6.7\textcolor{white}{0}} &
  43.1\scalebox{0.6}{±6.6\textcolor{white}{0}} &
  37.5\scalebox{0.6}{±8.6\textcolor{white}{0}} &
  52.4\scalebox{0.6}{±8.9\textcolor{white}{0}} &
  46.5\scalebox{0.6}{±6.3\textcolor{white}{0}} &
  34.6\scalebox{0.6}{±6.7\textcolor{white}{0}} &
  34.4\scalebox{0.6}{±5.6\textcolor{white}{0}} \\
\midrule
\texttt{DGI} &
  44.7\scalebox{0.6}{±4.4\textcolor{white}{0}} &
  45.5\scalebox{0.6}{±4.7\textcolor{white}{0}} &
  42.1\scalebox{0.6}{±7.6\textcolor{white}{0}} &
  56.0\scalebox{0.6}{±7.1\textcolor{white}{0}} &
  54.0\scalebox{0.6}{±4.9\textcolor{white}{0}} &
  36.7\scalebox{0.6}{±5.4\textcolor{white}{0}} &
  41.1\scalebox{0.6}{±6.5\textcolor{white}{0}} \\[-1pt]
\texttt{GRAPHCL} &
  42.1\scalebox{0.6}{±5.7\textcolor{white}{0}} &
  40.7\scalebox{0.6}{±7.3\textcolor{white}{0}} &
  42.7\scalebox{0.6}{±7.8\textcolor{white}{0}} &
  55.8\scalebox{0.6}{±6.7\textcolor{white}{0}} &
  55.9\scalebox{0.6}{±6.6\textcolor{white}{0}} &
  37.8\scalebox{0.6}{±4.5\textcolor{white}{0}} &
  42.4\scalebox{0.6}{±6.1\textcolor{white}{0}} \\
\bottomrule
\end{tabular*}
\vspace{-0.2cm}
\caption{Accuracy (\% ± std. for 20 runs) of \textbf{5-shot node classification} under targeted evasion attack ($p=$~3).}
\label{tab:d11}
\vspace{-0.2cm}
\end{table*}

\begin{table*}[!t]
\centering
\setlength{\tabcolsep}{1pt}
\renewcommand{\arraystretch}{0.95}

\begin{tabular*}{\linewidth}{@{\extracolsep{\fill}}l|cccc|ccc}
\toprule
\textbf{Method} & \texttt{Cora} & \texttt{CiteSeer} & \texttt{PubMed}
                & \texttt{obgn-Tech} & \texttt{obgn-Home}  & \texttt{obgn-arXiv} & \texttt{Wiki-CS} \\
\midrule
\texttt{GCN} &
  47.9\scalebox{0.6}{±6.1\textcolor{white}{0}} &
  48.7\scalebox{0.6}{±6.0\textcolor{white}{0}} &
  47.4\scalebox{0.6}{±7.3\textcolor{white}{0}} &
  69.8\scalebox{0.6}{±10.6} &
  64.6\scalebox{0.6}{±15.6} &
  38.8\scalebox{0.6}{±6.1\textcolor{white}{0}} &
  47.0\scalebox{0.6}{±6.7\textcolor{white}{0}} \\[-1pt]

\texttt{GAT} &
  50.2\scalebox{0.6}{±5.8\textcolor{white}{0}} &
  53.2\scalebox{0.6}{±5.7\textcolor{white}{0}} &
  52.2\scalebox{0.6}{±5.9\textcolor{white}{0}} &
  68.9\scalebox{0.6}{±10.2} &
  69.0\scalebox{0.6}{±11.2} &
  45.1\scalebox{0.6}{±8.5\textcolor{white}{0}} &
  47.9\scalebox{0.6}{±5.5\textcolor{white}{0}} \\
\midrule
\texttt{DGI} &
  57.4\scalebox{0.6}{±6.5\textcolor{white}{0}} &
  53.7\scalebox{0.6}{±4.9\textcolor{white}{0}} &
  60.4\scalebox{0.6}{±5.6\textcolor{white}{0}} &
  76.6\scalebox{0.6}{±7.8\textcolor{white}{0}} &
  68.0\scalebox{0.6}{±9.8\textcolor{white}{0}} &
  45.8\scalebox{0.6}{±6.0\textcolor{white}{0}} &
  50.3\scalebox{0.6}{±5.0\textcolor{white}{0}} \\[-1pt]

\texttt{GRAPHCL} &
  56.1\scalebox{0.6}{±6.5\textcolor{white}{0}} &
  55.0\scalebox{0.6}{±4.7\textcolor{white}{0}} &
  51.8\scalebox{0.6}{±5.4\textcolor{white}{0}} &
  74.0\scalebox{0.6}{±6.5\textcolor{white}{0}} &
  67.4\scalebox{0.6}{±12.3} &
  51.7\scalebox{0.6}{±5.3\textcolor{white}{0}} &
  54.8\scalebox{0.6}{±4.3\textcolor{white}{0}} \\
\bottomrule
\end{tabular*}
\vspace{-0.2cm}
\caption{Accuracy (\% ± std. for 20 runs) of \textbf{5-shot graph classification} under targeted evasion attack ($p=$~1).}
\label{tab:d12}
\vspace{-0.2cm}
\end{table*}

\begin{table*}[!t]
\centering
\setlength{\tabcolsep}{1pt}
\renewcommand{\arraystretch}{0.95}

\begin{tabular*}{\linewidth}{@{\extracolsep{\fill}}l|cccc|ccc}
\toprule
\textbf{Method} & \texttt{Cora} & \texttt{CiteSeer} & \texttt{PubMed}
                & \texttt{obgn-Tech} & \texttt{obgn-Home}  & \texttt{obgn-arXiv} & \texttt{Wiki-CS} \\
\midrule
\texttt{GCN} &
  46.8\scalebox{0.6}{±5.8\textcolor{white}{0}} &
  47.9\scalebox{0.6}{±5.6\textcolor{white}{0}} &
  46.6\scalebox{0.6}{±7.1\textcolor{white}{0}} &
  63.6\scalebox{0.6}{±10.6} &
  63.9\scalebox{0.6}{±15.2} &
  34.9\scalebox{0.6}{±7.1\textcolor{white}{0}} &
  43.8\scalebox{0.6}{±6.6\textcolor{white}{0}} \\[-1pt]

\texttt{GAT} &
  49.1\scalebox{0.6}{±5.7\textcolor{white}{0}} &
  53.3\scalebox{0.6}{±5.7\textcolor{white}{0}} &
  48.1\scalebox{0.6}{±5.9\textcolor{white}{0}} &
  68.7\scalebox{0.6}{±10.3} &
  68.5\scalebox{0.6}{±10.7} &
  45.4\scalebox{0.6}{±8.9\textcolor{white}{0}} &
  50.8\scalebox{0.6}{±5.5\textcolor{white}{0}} \\
\midrule
\texttt{DGI} &
  58.0\scalebox{0.6}{±6.4\textcolor{white}{0}} &
  53.5\scalebox{0.6}{±4.7\textcolor{white}{0}} &
  57.4\scalebox{0.6}{±5.8\textcolor{white}{0}} &
  72.6\scalebox{0.6}{±7.8\textcolor{white}{0}} &
  68.1\scalebox{0.6}{±9.9\textcolor{white}{0}} &
  45.6\scalebox{0.6}{±6.3\textcolor{white}{0}} &
  51.2\scalebox{0.6}{±5.1\textcolor{white}{0}} \\[-1pt]

\texttt{GRAPHCL} &
  57.1\scalebox{0.6}{±6.5\textcolor{white}{0}} &
  54.8\scalebox{0.6}{±4.8\textcolor{white}{0}} &
  51.7\scalebox{0.6}{±5.5\textcolor{white}{0}} &
  74.0\scalebox{0.6}{±6.6\textcolor{white}{0}} &
  67.3\scalebox{0.6}{±12.1} &
  47.3\scalebox{0.6}{±5.6\textcolor{white}{0}} &
  54.8\scalebox{0.6}{±4.2\textcolor{white}{0}} \\
\bottomrule
\end{tabular*}
\vspace{-0.2cm}
\caption{Accuracy (\% ± std. for 20 runs) of \textbf{5-shot graph classification} under targeted evasion attack ($p=$~2).}
\label{tab:d13}
\vspace{-0.2cm}
\end{table*}

\begin{table*}[!t]
\centering
\setlength{\tabcolsep}{1pt}
\renewcommand{\arraystretch}{0.95}

\begin{tabular*}{\linewidth}{@{\extracolsep{\fill}}l|cccc|ccc}
\toprule
\textbf{Method} & \texttt{Cora} & \texttt{CiteSeer} & \texttt{PubMed}
                & \texttt{obgn-Tech} & \texttt{obgn-Home}  & \texttt{obgn-arXiv} & \texttt{Wiki-CS} \\
\midrule
\texttt{GCN} &
  42.9\scalebox{0.6}{±5.8\textcolor{white}{0}} &
  42.6\scalebox{0.6}{±5.5\textcolor{white}{0}} &
  42.4\scalebox{0.6}{±7.1\textcolor{white}{0}} &
  60.6\scalebox{0.6}{±10.6} &
  54.6\scalebox{0.6}{±15.3} &
  32.8\scalebox{0.6}{±7.5\textcolor{white}{0}} &
  40.8\scalebox{0.6}{±6.7\textcolor{white}{0}} \\[-1pt]

\texttt{GAT} &
  45.4\scalebox{0.6}{±5.4\textcolor{white}{0}} &
  53.2\scalebox{0.6}{±5.6\textcolor{white}{0}} &
  46.0\scalebox{0.6}{±5.9\textcolor{white}{0}} &
  64.6\scalebox{0.6}{±10.3} &
  58.9\scalebox{0.6}{±10.8} &
  40.1\scalebox{0.6}{±8.7\textcolor{white}{0}} &
  45.7\scalebox{0.6}{±5.4\textcolor{white}{0}} \\
\midrule
\texttt{DGI} &
  54.6\scalebox{0.6}{±6.5\textcolor{white}{0}} &
  53.6\scalebox{0.6}{±4.5\textcolor{white}{0}} &
  52.3\scalebox{0.6}{±5.7\textcolor{white}{0}} &
  70.2\scalebox{0.6}{±7.7\textcolor{white}{0}} &
  68.1\scalebox{0.6}{±9.6\textcolor{white}{0}} &
  43.4\scalebox{0.6}{±5.9\textcolor{white}{0}} &
  45.1\scalebox{0.6}{±5.1\textcolor{white}{0}} \\[-1pt]

\texttt{GRAPHCL} &
  52.2\scalebox{0.6}{±6.6\textcolor{white}{0}} &
  54.5\scalebox{0.6}{±5.1\textcolor{white}{0}} &
  48.8\scalebox{0.6}{±5.3\textcolor{white}{0}} &
  73.3\scalebox{0.6}{±6.6\textcolor{white}{0}} &
  68.0\scalebox{0.6}{±12.3} &
  42.8\scalebox{0.6}{±5.9\textcolor{white}{0}} &
  48.9\scalebox{0.6}{±4.2\textcolor{white}{0}} \\
\bottomrule
\end{tabular*}
\vspace{-0.2cm}
\caption{Accuracy (\% ± std. for 20 runs) of \textbf{5-shot graph classification} under targeted evasion attack ($p=$~3).}
\label{tab:d14}
\vspace{-0.2cm}
\end{table*}

\begin{table*}[!t]
\centering
\setlength{\tabcolsep}{1pt}      
\renewcommand{\arraystretch}{0.95}

\begin{tabular*}{\linewidth}{@{\extracolsep{\fill}}l|cccc|ccc}
\toprule
\textbf{Method} & \texttt{Cora} & \texttt{CiteSeer} & \texttt{PubMed}
                & \texttt{obgn-Tech} & \texttt{obgn-Home}  & \texttt{obgn-arXiv} & \texttt{Wiki-CS} \\
\midrule
\texttt{GCN} & 49.5\scalebox{0.6}{±5.5\textcolor{white}{0}}
             & 37.0\scalebox{0.6}{±6.9\textcolor{white}{0}}
             & 51.7\scalebox{0.6}{±12.4}
             & 69.8\scalebox{0.6}{±12.0}
             & 51.4\scalebox{0.6}{±10.4}
             & 30.3\scalebox{0.6}{±5.5\textcolor{white}{0}}
             & 49.4\scalebox{0.6}{±6.6\textcolor{white}{0}} \\

\texttt{GAT} & 50.2\scalebox{0.6}{±7.5\textcolor{white}{0}}
             & 46.5\scalebox{0.6}{±4.3\textcolor{white}{0}}
             & 49.0\scalebox{0.6}{±8.9\textcolor{white}{0}}
             & 71.2\scalebox{0.6}{±8.9\textcolor{white}{0}}
             & 61.6\scalebox{0.6}{±7.3\textcolor{white}{0}}
             & 39.1\scalebox{0.6}{±4.9\textcolor{white}{0}}
             & 51.5\scalebox{0.6}{±6.7\textcolor{white}{0}} \\
\midrule
\texttt{DGI} & 53.3\scalebox{0.6}{±6.8\textcolor{white}{0}}
             & 45.8\scalebox{0.6}{±4.9\textcolor{white}{0}}
             & 46.2\scalebox{0.6}{±8.6\textcolor{white}{0}}
             & 70.2\scalebox{0.6}{±4.7\textcolor{white}{0}}
             & 59.9\scalebox{0.6}{±6.3\textcolor{white}{0}}
             & 46.4\scalebox{0.6}{±3.9\textcolor{white}{0}}
             & 45.1\scalebox{0.6}{±6.4\textcolor{white}{0}} \\

\texttt{GRAPHCL} & 57.2\scalebox{0.6}{±6.5\textcolor{white}{0}}
                 & 47.1\scalebox{0.6}{±6.3\textcolor{white}{0}}
                 & 56.6\scalebox{0.6}{±8.7\textcolor{white}{0}}
                 & 71.2\scalebox{0.6}{±4.6\textcolor{white}{0}}
                 & 58.4\scalebox{0.6}{±6.7\textcolor{white}{0}}
                 & 42.9\scalebox{0.6}{±4.9\textcolor{white}{0}}
                 & 49.8\scalebox{0.6}{±6.2\textcolor{white}{0}} \\
\bottomrule
\end{tabular*}
\vspace{-0.2cm}
\caption{Accuracy (\% ± std. for 20 runs) of \textbf{5-shot node classification} under targeted poisoning attack ($p=$~1).}
\label{tab:d15}
\vspace{-0.2cm}
\end{table*}

\begin{table*}[!t]
\centering
\setlength{\tabcolsep}{1pt}      
\renewcommand{\arraystretch}{0.95}

\begin{tabular*}{\linewidth}{@{\extracolsep{\fill}}l|cccc|ccc}
\toprule
\textbf{Method} & \texttt{Cora} & \texttt{CiteSeer} & \texttt{PubMed}
                & \texttt{obgn-Tech} & \texttt{obgn-Home}  & \texttt{obgn-arXiv} & \texttt{Wiki-CS} \\
\midrule
\texttt{GCN} & 48.1\scalebox{0.6}{±9.0\textcolor{white}{0}}
             & 36.4\scalebox{0.6}{±4.8\textcolor{white}{0}}
             & 38.9\scalebox{0.6}{±11.9\textcolor{white}{0}}
             & 58.8\scalebox{0.6}{±13.7\textcolor{white}{0}}
             & 48.4\scalebox{0.6}{±10.2\textcolor{white}{0}}
             & 34.8\scalebox{0.6}{±5.0\textcolor{white}{0}}
             & 47.3\scalebox{0.6}{±7.6\textcolor{white}{0}} \\

\texttt{GAT} & 59.0\scalebox{0.6}{±6.2\textcolor{white}{0}}
             & 44.5\scalebox{0.6}{±5.2\textcolor{white}{0}}
             & 42.0\scalebox{0.6}{±9.0\textcolor{white}{0}}
             & 62.2\scalebox{0.6}{±8.8\textcolor{white}{0}}
             & 59.1\scalebox{0.6}{±7.0\textcolor{white}{0}}
             & 31.9\scalebox{0.6}{±4.5\textcolor{white}{0}}
             & 45.2\scalebox{0.6}{±5.6\textcolor{white}{0}} \\
\midrule
\texttt{DGI} & 57.5\scalebox{0.6}{±6.1\textcolor{white}{0}}
             & 45.9\scalebox{0.6}{±4.5\textcolor{white}{0}}
             & 43.3\scalebox{0.6}{±9.3\textcolor{white}{0}}
             & 66.5\scalebox{0.6}{±4.7\textcolor{white}{0}}
             & 55.4\scalebox{0.6}{±7.9\textcolor{white}{0}}
             & 45.7\scalebox{0.6}{±5.1\textcolor{white}{0}}
             & 43.8\scalebox{0.6}{±6.9\textcolor{white}{0}} \\
\bottomrule
\end{tabular*}
\vspace{-0.2cm}
\caption{Accuracy (\% ± std. for 20 runs) of \textbf{5-shot node classification} under targeted poisoning attack ($p=$~2).}
\label{tab:d16}
\vspace{-0.2cm}
\end{table*}

\begin{table*}[!t]
\centering
\setlength{\tabcolsep}{1pt}          
\renewcommand{\arraystretch}{0.95}   

\begin{tabular*}{\linewidth}{@{\extracolsep{\fill}}l|cccc|ccc}
\toprule
\textbf{Method} & \texttt{Cora} & \texttt{CiteSeer} & \texttt{PubMed}
                & \texttt{obgn-Tech} & \texttt{obgn-Home}  & \texttt{obgn-arXiv} & \texttt{Wiki-CS} \\
\midrule
\texttt{GCN} &
  44.4\scalebox{0.6}{±4.7\textcolor{white}{0}} &
  34.0\scalebox{0.6}{±5.5\textcolor{white}{0}} &
  32.4\scalebox{0.6}{±9.9\textcolor{white}{0}} &
  57.8\scalebox{0.6}{±12.8} &
  46.4\scalebox{0.6}{±13.4} &
  38.3\scalebox{0.6}{±6.0\textcolor{white}{0}} &
  38.5\scalebox{0.6}{±8.8\textcolor{white}{0}} \\

\texttt{GAT} &
  52.4\scalebox{0.6}{±6.9\textcolor{white}{0}} &
  42.6\scalebox{0.6}{±5.6\textcolor{white}{0}} &
  38.3\scalebox{0.6}{±8.3\textcolor{white}{0}} &
  54.4\scalebox{0.6}{±7.8\textcolor{white}{0}} &
  56.0\scalebox{0.6}{±7.7\textcolor{white}{0}} &
  38.5\scalebox{0.6}{±5.4\textcolor{white}{0}} &
  37.6\scalebox{0.6}{±6.7\textcolor{white}{0}} \\
\midrule
\texttt{DGI} &
  47.8\scalebox{0.6}{±6.0\textcolor{white}{0}} &
  45.2\scalebox{0.6}{±5.9\textcolor{white}{0}} &
  35.9\scalebox{0.6}{±8.9\textcolor{white}{0}} &
  55.4\scalebox{0.6}{±6.0\textcolor{white}{0}} &
  51.1\scalebox{0.6}{±10.6} &
  44.8\scalebox{0.6}{±4.5\textcolor{white}{0}} &
  45.9\scalebox{0.6}{±5.1\textcolor{white}{0}} \\

\texttt{GRAPHCL} &
  48.9\scalebox{0.6}{±5.1\textcolor{white}{0}} &
  41.3\scalebox{0.6}{±5.3\textcolor{white}{0}} &
  37.6\scalebox{0.6}{±7.6\textcolor{white}{0}} &
  67.6\scalebox{0.6}{±3.4\textcolor{white}{0}} &
  53.0\scalebox{0.6}{±9.1\textcolor{white}{0}} &
  30.0\scalebox{0.6}{±5.5\textcolor{white}{0}} &
  41.3\scalebox{0.6}{±4.7\textcolor{white}{0}} \\
\bottomrule
\end{tabular*}
\vspace{-0.2cm}
\caption{Accuracy (\% ± std. for 20 runs) of \textbf{5-shot node classification} under targeted poisoning attack ($p=$~3).}
\label{tab:d17}
\vspace{-0.2cm}
\end{table*}

\begin{table*}[!t]
\centering
\setlength{\tabcolsep}{1pt}          
\renewcommand{\arraystretch}{0.95}   

\begin{tabular*}{\linewidth}{@{\extracolsep{\fill}}l|cccc|ccc}
\toprule
\textbf{Method} & \texttt{Cora} & \texttt{CiteSeer} & \texttt{PubMed}
                & \texttt{obgn-Tech} & \texttt{obgn-Home}  & \texttt{obgn-arXiv} & \texttt{Wiki-CS} \\
\midrule
\texttt{GCN} &
  46.5\scalebox{0.6}{±7.6\textcolor{white}{0}} &
  38.9\scalebox{0.6}{±6.9\textcolor{white}{0}} &
  38.3\scalebox{0.6}{±6.1\textcolor{white}{0}} &
  69.2\scalebox{0.6}{±10.4} &
  59.2\scalebox{0.6}{±18.2} &
  35.0\scalebox{0.6}{±5.7\textcolor{white}{0}} &
  45.6\scalebox{0.6}{±10.1} \\

\texttt{GAT} &
  49.1\scalebox{0.6}{±5.5\textcolor{white}{0}} &
  51.0\scalebox{0.6}{±5.2\textcolor{white}{0}} &
  47.6\scalebox{0.6}{±6.6\textcolor{white}{0}} &
  69.1\scalebox{0.6}{±10.4} &
  58.3\scalebox{0.6}{±10.7} &
  46.0\scalebox{0.6}{±7.5\textcolor{white}{0}} &
  40.8\scalebox{0.6}{±5.6\textcolor{white}{0}} \\
\midrule
\texttt{DGI} &
  51.3\scalebox{0.6}{±6.8\textcolor{white}{0}} &
  53.1\scalebox{0.6}{±6.5\textcolor{white}{0}} &
  57.8\scalebox{0.6}{±6.6\textcolor{white}{0}} &
  72.9\scalebox{0.6}{±8.6\textcolor{white}{0}} &
  67.5\scalebox{0.6}{±10.7} &
  45.8\scalebox{0.6}{±5.8\textcolor{white}{0}} &
  50.8\scalebox{0.6}{±4.6\textcolor{white}{0}} \\

\texttt{GRAPHCL} &
  43.9\scalebox{0.6}{±5.4\textcolor{white}{0}} &
  53.8\scalebox{0.6}{±6.8\textcolor{white}{0}} &
  49.8\scalebox{0.6}{±6.1\textcolor{white}{0}} &
  74.8\scalebox{0.6}{±5.8\textcolor{white}{0}} &
  65.3\scalebox{0.6}{±13.7} &
  52.3\scalebox{0.6}{±7.1\textcolor{white}{0}} &
  43.9\scalebox{0.6}{±4.9\textcolor{white}{0}} \\

\texttt{GraphPrompt} &
  51.1\scalebox{0.6}{±6.0\textcolor{white}{0}} &
  55.1\scalebox{0.6}{±5.5\textcolor{white}{0}} &
  49.1\scalebox{0.6}{±6.9\textcolor{white}{0}} &
  81.4\scalebox{0.6}{±9.2\textcolor{white}{0}} &
  61.6\scalebox{0.6}{±11.7} &
  44.8\scalebox{0.6}{±6.0\textcolor{white}{0}} &
  48.2\scalebox{0.6}{±4.8\textcolor{white}{0}} \\
\bottomrule
\end{tabular*}
\vspace{-0.3cm}
\caption{Accuracy (\% ± std. for 20 runs) of \textbf{5-shot graph classification} under targeted poisoning attack ($p=$~1).}
\label{tab:d18}
\vspace{-0.2cm}
\end{table*}

\begin{table*}[!t]
\centering
\setlength{\tabcolsep}{1pt}
\renewcommand{\arraystretch}{0.95}

\begin{tabular*}{\linewidth}{@{\extracolsep{\fill}}l|cccc|ccc}
\toprule
\textbf{Method} & \texttt{Cora} & \texttt{CiteSeer} & \texttt{PubMed}
                & \texttt{obgn-Tech} & \texttt{obgn-Home}  & \texttt{obgn-arXiv} & \texttt{Wiki-CS} \\
\midrule
\texttt{GCN} &
  46.8\scalebox{0.6}{±8.1\textcolor{white}{0}} &
  38.1\scalebox{0.6}{±6.4\textcolor{white}{0}} &
  37.6\scalebox{0.6}{±7.0\textcolor{white}{0}} &
  64.4\scalebox{0.6}{±10.8} &
  54.1\scalebox{0.6}{±14.6} &
  33.2\scalebox{0.6}{±6.5\textcolor{white}{0}} &
  43.1\scalebox{0.6}{±8.2\textcolor{white}{0}} \\

\texttt{GAT} &
  47.6\scalebox{0.6}{±5.6\textcolor{white}{0}} &
  49.6\scalebox{0.6}{±5.7\textcolor{white}{0}} &
  47.3\scalebox{0.6}{±6.6\textcolor{white}{0}} &
  60.2\scalebox{0.6}{±9.2\textcolor{white}{0}} &
  56.8\scalebox{0.6}{±10.3} &
  45.1\scalebox{0.6}{±8.8\textcolor{white}{0}} &
  47.1\scalebox{0.6}{±6.2\textcolor{white}{0}} \\
\midrule
\texttt{DGI} &
  50.3\scalebox{0.6}{±5.7\textcolor{white}{0}} &
  52.0\scalebox{0.6}{±4.7\textcolor{white}{0}} &
  49.9\scalebox{0.6}{±6.6\textcolor{white}{0}} &
  63.3\scalebox{0.6}{±8.0\textcolor{white}{0}} &
  67.0\scalebox{0.6}{±9.7\textcolor{white}{0}} &
  44.3\scalebox{0.6}{±5.6\textcolor{white}{0}} &
  49.5\scalebox{0.6}{±4.9\textcolor{white}{0}} \\

\texttt{GRAPHCL} &
  43.1\scalebox{0.6}{±5.7\textcolor{white}{0}} &
  53.1\scalebox{0.6}{±4.7\textcolor{white}{0}} &
  50.8\scalebox{0.6}{±6.2\textcolor{white}{0}} &
  73.7\scalebox{0.6}{±6.8\textcolor{white}{0}} &
  62.8\scalebox{0.6}{±12.0} &
  51.3\scalebox{0.6}{±5.6\textcolor{white}{0}} &
  42.2\scalebox{0.6}{±4.7\textcolor{white}{0}} \\
\bottomrule
\end{tabular*}
\vspace{-0.3cm}
\caption{Accuracy (\% ± std. for 20 runs) of \textbf{5-shot graph classification} under targeted poisoning attack ($p=$~2).}
\label{tab:d19}
\vspace{-0.2cm}
\end{table*}

\begin{table*}[!t]
\centering
\setlength{\tabcolsep}{1pt}
\renewcommand{\arraystretch}{0.95}

\begin{tabular*}{\linewidth}{@{\extracolsep{\fill}}l|cccc|ccc}
\toprule
\textbf{Method} & \texttt{Cora} & \texttt{CiteSeer} & \texttt{PubMed}
                & \texttt{obgn-Tech} & \texttt{obgn-Home}  & \texttt{obgn-arXiv} & \texttt{Wiki-CS} \\
\midrule
\texttt{GCN} &
  42.3\scalebox{0.6}{±7.4\textcolor{white}{0}} &
  35.1\scalebox{0.6}{±6.1\textcolor{white}{0}} &
  38.3\scalebox{0.6}{±5.4\textcolor{white}{0}} &
  59.2\scalebox{0.6}{±9.8\textcolor{white}{0}} &
  52.9\scalebox{0.6}{±16.7} &
  35.7\scalebox{0.6}{±5.8\textcolor{white}{0}} &
  41.4\scalebox{0.6}{±9.6\textcolor{white}{0}} \\

\texttt{GAT} &
  48.2\scalebox{0.6}{±6.8\textcolor{white}{0}} &
  39.4\scalebox{0.6}{±5.0\textcolor{white}{0}} &
  37.6\scalebox{0.6}{±5.4\textcolor{white}{0}} &
  59.1\scalebox{0.6}{±9.7\textcolor{white}{0}} &
  56.5\scalebox{0.6}{±8.4\textcolor{white}{0}} &
  42.2\scalebox{0.6}{±7.7\textcolor{white}{0}} &
  48.8\scalebox{0.6}{±5.7\textcolor{white}{0}} \\
\midrule
\texttt{DGI} &
  50.7\scalebox{0.6}{±5.1\textcolor{white}{0}} &
  46.2\scalebox{0.6}{±5.2\textcolor{white}{0}} &
  50.6\scalebox{0.6}{±5.1\textcolor{white}{0}} &
  62.9\scalebox{0.6}{±7.4\textcolor{white}{0}} &
  56.1\scalebox{0.6}{±7.3\textcolor{white}{0}} &
  42.4\scalebox{0.6}{±5.2\textcolor{white}{0}} &
  49.0\scalebox{0.6}{±4.5\textcolor{white}{0}} \\

\texttt{GRAPHCL} &
  43.9\scalebox{0.6}{±5.5\textcolor{white}{0}} &
  42.9\scalebox{0.6}{±4.7\textcolor{white}{0}} &
  46.5\scalebox{0.6}{±5.7\textcolor{white}{0}} &
  54.0\scalebox{0.6}{±6.8\textcolor{white}{0}} &
  62.3\scalebox{0.6}{±8.9\textcolor{white}{0}} &
  49.5\scalebox{0.6}{±4.9\textcolor{white}{0}} &
  43.1\scalebox{0.6}{±3.7\textcolor{white}{0}} \\
\bottomrule
\end{tabular*}
\vspace{-0.3cm}
\caption{Accuracy (\% ± std. for 20 runs) of \textbf{5-shot graph classification} under targeted poisoning attack ($p=$~3).}
\label{tab:d20}
\vspace{-0.2cm}
\end{table*}

This section provides more detailed experimental results for RQ1 and RQ2, including the results for each dataset as the target domain under the corresponding attacks.
\subsection{Under Non-targeted Feature Attacks}
The results presented in Tables~\ref{tab:d1}, \ref{tab:d2}, \ref{tab:d3}, and \ref{tab:d4} show that:
\ding{182} In the node classification task, \modelname~consistently achieves state-of-the-art performance across all datasets under both moderate ($\lambda=0.4$) and high ($\lambda=0.8$) levels of feature noise. This demonstrates the model's superior and stable robustness against feature perturbations.
\ding{183} When compared to the strongest robust baseline, \texttt{MDGFM}, our model's performance advantage is particularly evident under high noise conditions. For instance, in the node classification task on \texttt{ogbn-tech} with $\lambda=0.8$ (Table~\ref{tab:d2}), \modelname~achieves an accuracy of 77.9\%, significantly outperforming \texttt{MDGFM}'s 75.8\%. This consistent superiority suggests that our structure-aware semantic augmentation and information bottleneck mechanism are highly effective at preserving essential information and filtering noise when node features are severely corrupted.
\ding{184} While all models experience performance degradation as the noise level increases from $\lambda=0.4$ to $\lambda=0.8$, \modelname~exhibits a more graceful decline, maintaining its top-ranking position and widening the performance gap with other baselines. This resilience is critical in real-world scenarios where data quality varies, highlighting our approach's practical value. The experimental results for the graph classification task (Tables~\ref{tab:d3} and \ref{tab:d4}) show a similar trend, further corroborating the robustness and effectiveness of our proposed method.

\subsection{Under Non-targeted Structure Attacks}
The results presented in Tables \ref{tab:d5}, \ref{tab:d6}, \ref{tab:d7}, and \ref{tab:d8} show that:
\ding{182} For the node classification task, \modelname~maintains its top-tier performance across all datasets when subjected to random structural perturbations. Even with a high proportion of edges removed ($\lambda=0.8$), our model consistently outperforms all baselines, showcasing its exceptional structural robustness.
\ding{183} The advantage of our model is especially clear when compared with \texttt{MDGFM}. For example, on the \texttt{ogbn-arxiv} dataset with $\lambda=0.8$ (Table~\ref{tab:d6}), \modelname~achieves an accuracy of 53.2\%, which is a notable improvement over \texttt{MDGFM}'s 50.9\%. This indicates that our hierarchical structure optimization module is more effective at recovering and preserving crucial topological information than the GSL methods used in other baselines.
\ding{184} As the structural noise intensifies from $\lambda=0.4$ to $\lambda=0.8$, the performance of most models drops significantly. However, \modelname~demonstrates a more resilient performance, with a smaller accuracy decrease compared to its competitors. The results from the graph classification experiments (Tables~\ref{tab:d7} and \ref{tab:d8}) further validate these findings, confirming the superior structural robustness of our approach.

\subsection{Under Targeted Evasion Attacks}
The results presented in Tables \ref{tab:d9} to \ref{tab:d14} show that:
\ding{182} In the node classification task, \modelname~demonstrates remarkable robustness against targeted evasion attacks, consistently outperforming all baselines across every dataset and at all attack intensities ($p=1, 2, 3$). This underscores its strong defensive capabilities against strategic adversarial perturbations.
\ding{183} The performance gap between \modelname~and the strongest baseline, \texttt{MDGFM}, widens as the attack becomes more severe. For instance, under the most intense attack ($p=3$) on the \texttt{ogbn-arxiv} dataset (Table~\ref{tab:d11}), \modelname~achieves 50.1\% accuracy, whereas \texttt{MDGFM} drops to 45.6\%. This highlights the effectiveness of our expert routing and structural optimization in preserving performance even when the adversary has more power.
\ding{184} As the number of perturbations increases, all models show a decline in accuracy, but \modelname's performance degrades more slowly and gracefully than its counterparts. This trend of superior stability is also reflected in the graph classification results (Tables~\ref{tab:d12} to \ref{tab:d14}), further confirming that our model's robustness is consistent across different tasks.

\subsection{Under Targeted Poisoning Attacks}
The results presented in Tables D.15 to D.20 show that:
\ding{182} \modelname~shows exceptional robustness against targeted poisoning attacks, consistently leading in performance for node classification across all datasets and attack levels ($p=1, 2, 3$). This proves its ability to learn reliable representations even from corrupted training data.
\ding{183} The model's advantage over the strongest baseline, \texttt{MDGFM}, is significant. For example, in the most severe attack scenario ($p=3$) on the \texttt{Wiki-CS} dataset (Table D.17), \modelname~achieves 50.7\% accuracy, while \texttt{MDGFM}'s performance degrades to 46.4\%. This suggests that our pre-training strategy, which distills robust patterns via the information bottleneck, and the hierarchical fine-tuning are crucial for defending against poisoning.
\ding{184} As the poisoning attack intensifies, \modelname~demonstrates superior stability with a more graceful performance decline compared to other baselines. This resilience is further confirmed by the graph classification results (Tables D.18 to D.20), which exhibit the same trend of robust and superior performance.

\section{Related Work Discussions}
\label{sec:related_work_add}
\subsection{Multi-domain Graph Foundation Models}
\label{sec:related_work_add_gfm}
Multi-domain graph pre-training is a cornerstone for developing general-purpose Graph Foundation Models (GFMs), aiming to distill transferable knowledge from diverse domains~\cite{yu2024text, yuan2025how}. Addressing the domain gap is a central challenge, with existing methods focusing on either feature or structure-level alignment.

\textbf{Feature-level alignment} aims to unify heterogeneous semantic spaces. For instance, MDGPT~\cite{yu2024text} introduces learnable domain tokens to adjust features from each source domain. In contrast, BRIDGE~\cite{yuan2025how} employs domain-invariant aligners to extract generalizable representations and provides a theoretical upper bound on the generalization error. Other works have also explored leveraging language models for alignment, though these are often limited to text-attributed graphs~\cite{tang2024higpt}. \textbf{Structure-level alignment} directly addresses topological discrepancies. A representative work, MDGFM~\cite{wang2025multidomain}, integrates Graph Structure Learning (GSL) to refine topologies, aiming to learn domain-invariant structural knowledge. To mitigate negative transfer, frameworks like GraphBridge~\cite{ju2025graphbridge} utilize a lightweight Mixture-of-Experts (MoE) network for selective knowledge transfer.

Despite these advances, most methods tackle feature and structure adaptation in isolation. For example, feature-centric models like MDGPT and BRIDGE may lack robustness to structural noise, while structure-focused approaches such as MDGFM can be computationally expensive. Furthermore, the interaction between semantic feature shifts and topological noise is often overlooked, preventing a holistic understanding of cross-domain generalization and robustness. This leaves a gap for a unified framework that motivates the design of our proposed method.

\subsection{Robust Graph Representation Learning}
\label{sec:related_work_add_robust}
Robust graph representation learning focuses on improving model stability under various forms of data uncertainty, including both feature and structural perturbations. To enhance semantic resilience against noise in node attributes, prior work has explored several strategies. These include data augmentation, which enriches training data through methods like intra-class mixup~\cite{zheng2024intramix} or learning beneficial noise~\cite{huang2025learn}; adversarial training on perturbed features to improve defensive capabilities~\cite{lee2025selfsupervised}; and disentangled learning, which separates invariant causal factors from spurious correlations to handle distribution shifts~\cite{li2025disentangle}.

On the structural level, research primarily addresses the vulnerability of GNNs to topological noise and adversarial attacks~\cite{zugner2018adversarial}. Graph Structure Learning (GSL) has emerged as a key defense strategy~\cite{zhu2021deep}, where the goal is to learn an optimized graph structure that is more reliable than the original input. Recent progress in this area includes techniques like topology-aware graph pruning, which removes potentially harmful edges to safeguard the model and improve its stability~\cite{zhang2024gder}.

However, a significant limitation of existing work is that feature and structure robustness are often treated as separate problems, without considering their interaction. Furthermore, many GSL approaches, including the one used in MDGFM, are computationally intensive and lack fine-grained control over distinct structural components such as intra- and inter-cluster connections. This gap for a more unified and efficient approach motivates our work.

\subsection{Graph Structural Entropy}
\label{sec:related_work_add_entropy}
Graph Structural Entropy is an information-theoretic measure that quantifies the structural complexity of a network's topology via an optimal encoding scheme~\cite{li2016structural}. The theory posits that a graph's structural information is the minimum number of bits needed to locate a node through a random walk. Minimizing this entropy yields a hierarchical encoding tree, which recursively partitions the graph to reveal its multi-scale community structure.

The hierarchical organization derived from structural entropy has been recognized as a valuable prior for graph representation learning. For instance, researchers have leveraged this principle to inject topic-aware signals into graph learning~\cite{long2020graphstone} or to guide hierarchical clustering for learning more structurally informed node embeddings~\cite{bai2024hcgae}. These approaches validate the effectiveness of structural entropy in capturing meaningful topological patterns.

Despite its theoretical elegance, the potential of this hierarchical prior to systematically enhance raw node features for improving the robustness of Graph Foundation Models (GFMs) is largely untapped. Our work aims to fill this void, utilizing the encoding tree not merely as a structural summary, but as a direct source for generating structure-aware semantic augmentations to bolster model resilience against perturbations.

\section{Limitations}
While \modelname~ shows promising results, there are several directions for future enhancement. Our robustness evaluation, while comprehensive, primarily centers on standard perturbations to node features and graph structures. A natural and important extension would be to test the model's resilience against a wider range of adversarial scenarios, such as more complex topological attacks. This would provide a more complete picture of its defensive capabilities. Secondly, the framework's effectiveness has been validated on benchmark homogeneous and static graphs. An important next step is to investigate its applicability to more complex graph types, such as large-scale heterogeneous or dynamic networks. Adapting the model for these scenarios is crucial for its practical deployment and represents a key area for future research. Finally, the current work focuses on establishing the core methodology of \modelname~. Exploring its integration and potential synergies with other emerging pre-training paradigms is a promising avenue. Understanding how to combine our structure-aware approach with complementary techniques could lead to more powerful models.

\end{document}